\documentclass{applemlr}
\pdfoutput=1

\usepackage{xspace}
\usepackage[table,xcdraw,dvipsnames]{xcolor}

\usepackage{hyperref}
\usepackage{url}
\usepackage{amsthm}

\usepackage{fontawesome5}
\usepackage[smartEllipses]{markdown}

\usepackage[most]{tcolorbox}
\usepackage{wrapfig}
\usepackage{multirow}
\usepackage{graphicx,float}
\usepackage{threeparttable}
\usepackage[ruled,vlined]{algorithm2e}
\usepackage{colortbl}
\usepackage{color}
\usepackage{tabularx}
\usepackage{float}
\usepackage{booktabs}
\usepackage{arydshln}
\usepackage{enumitem}
\usepackage{caption}
\usepackage{graphicx}
\usepackage{makecell}
\usepackage{float} 
\usepackage{mathtools}
\usepackage{bbding}
\usepackage{tabularx}
\usepackage{amssymb,mathrsfs,amsmath}
\usepackage{pifont}
\usepackage{bbm}
\usepackage{hhline}
\usepackage{boldline}
\usepackage{lmodern}

\usepackage{listings}

\def\dataset{PAP-12K}
\def\method{PAP}
\renewcommand{\cite}[2][]{\citep[#1]{#2}}

\definecolor{newgray}{gray}{0.4}

\usepackage{cleveref}
\SetKwInOut{Input}{Input}\SetKwInOut{Output}{Output}
\usepackage{twemojis}

\hypersetup{
    colorlinks=true,
    linkcolor=magenta,
    citecolor=cyan,
    filecolor=magenta,      
    urlcolor=magenta,
    }

\title{Panoramic Affordance Prediction}

\author[1\dagger]{Zixin Zhang}
\author[1\dagger]{Chenfei Liao}
\author[1]{Hongfei Zhang}
\author[1]{Harold H. Chen}
\author[1]{Kanghao Chen}
\author[3]{Zichen Wen}
\author[1]{Litao Guo}
\author[4]{Bin Ren}
\author[1]{Xu Zheng}
\author[6]{Yinchuan Li}
\author[1,2]{Xuming Hu}
\author[5]{Nicu Sebe}
\author[1,2\ddagger]{Ying-Cong Chen}
\affiliation[1]{HKUST(GZ)}
\affiliation[2]{HKUST}
\affiliation[3]{SJTU}
\affiliation[4]{MBZUAI}
\affiliation[5]{UniTrento}
\affiliation[6]{Knowin}
\contribution[\dagger]{Equal contribution}
\contribution[\ddagger]{Corresponding author}

\abstract{Affordance prediction serves as a critical bridge between perception and action in embodied AI. However, existing research is confined to pinhole camera models, which suffer from narrow Fields of View (FoV) and fragmented observations, often missing critical holistic environmental context. In this paper, we present the first exploration into \textbf{Panoramic Affordance Prediction}, utilizing 360-degree imagery to capture global spatial relationships and holistic scene understanding. To facilitate this novel task, we first introduce \textbf{\dataset}, a large-scale benchmark dataset containing over 1,000 ultra-high-resolution (12k, 11904$\times$5952) panoramic images with over 12k carefully annotated QA pairs and affordance masks. Furthermore, we propose \textbf{\method}, a training-free, coarse-to-fine pipeline inspired by the human foveal visual system to tackle the ultra-high resolution and severe distortion inherent in panoramic images. \method~employs recursive visual routing via grid prompting to progressively locate targets, applies an adaptive gaze mechanism to rectify local geometric distortions, and utilizes a cascaded grounding pipeline to extract precise instance-level masks. Experimental results on \dataset~reveal that existing affordance prediction methods designed for standard perspective images suffer severe performance degradation and fail due to the unique challenges of panoramic vision. In contrast, \method~framework effectively overcomes these obstacles, significantly outperforming state-of-the-art baselines and highlighting the immense potential of panoramic perception for robust embodied intelligence.}

\metadata[{\faClock\ Date}]{\today}
\metadata[{\faGlobe\ Project}]{\url{https://zixinzhang02.github.io/Panoramic-Affordance-Prediction/}}
\metadata[{\faGithub\ Code \& Dataset}]{\url{https://github.com/EnVision-Research/PAP}}

\usepackage[most]{tcolorbox}

\newcommand{\promptsection}[1]{\vspace{1ex}\noindent\textbf{\large #1}} 
\newcommand{\promptsubsection}[1]{\vspace{1ex}\noindent\textbf{\normal #1}}
\newcommand{\promptsubsubsection}[1]{\vspace{1ex}\noindent\textbf{\small #1}}

\newtcolorbox{renderedpromptbox}[1][]{
  enhanced,
  breakable,
  colback=gray!5,
  colframe=gray!50,
  fonttitle=\bfseries,
  coltitle=black,
  attach boxed title to top left={yshift=-2mm, xshift=5mm},
  boxed title style={colback=white, colframe=gray!50},
  title=#1,
  fontupper=\small,
  before upper={
    \let\section\promptsection
    \let\subsection\promptsubsection
    \let\subsubsection\promptsubsubsection
    \setlist[itemize]{nosep, leftmargin=*}
    \setlist[enumerate]{nosep, leftmargin=*}
  }
}

\providecommand{\tightlist}{%
  \setlength{\itemsep}{0pt}\setlength{\parskip}{0pt}}
\usepackage{titletoc}
\providecommand{\authcount}[1]{}
\providecommand{\tocauthor}[1]{}

\begin{document}
\maketitle

\begin{figure*}[!t]
    \centering
    \includegraphics[width=\linewidth]{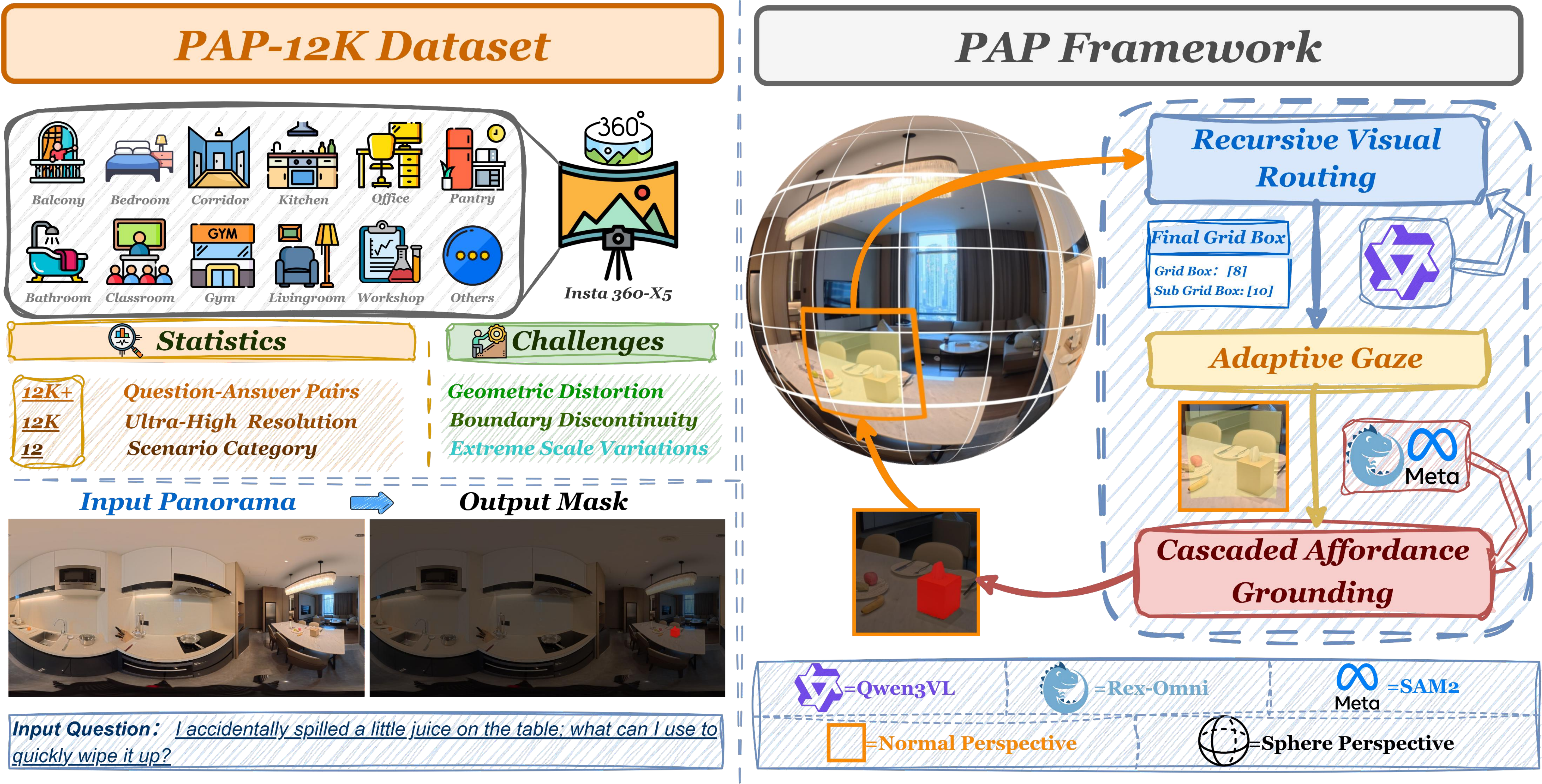}
    \caption{\textbf{Overview of our work.} \textbf{Left:} We introduce \dataset, the first large-scale benchmark dedicated to panoramic affordance prediction, featuring ultra-high resolution (12K) imagery, rich reasoning-based QA pairs, and explicitly capturing unique panoramic challenges (geometric distortion, boundary discontinuity, and extreme scale variations). \textbf{Right:} We propose the \method~framework, which mimics human foveal vision to tackle these challenges. It employs \textit{Recursive Visual Routing} for efficient coarse localization, an \textit{Adaptive Gaze} mechanism to rectify spatial distortions, and \textit{Cascaded Affordance Grounding} for precise instance-level mask extraction.}
    \label{fig:teaser}
\end{figure*}

\section{Introduction}
\label{sec:intro}

Affordance~\cite{gibson1977theory} defines the action possibilities that an environment offers to an agent. As a critical intermediate representation, it serves as the bridge connecting visual perception with actionable capabilities in embodied AI. By explicitly mapping the functionality of the environment—identifying \textit{where} and \textit{how} an agent can interact—affordance prediction provides invaluable semantic information for downstream robotic applications, ranging from task planning and tool usage to complex object manipulation.

To interact seamlessly within complex 3D environments, embodied agents require not just local action cues, but global situational awareness. However, existing research in affordance prediction~\cite{zhang2025a4,affordance-r1,wu2025ragnet} is predominantly confined to the ``Pinhole Camera Model''. This reliance inherently suffers from a restricted Field-of-View (FoV), or ``tunnel vision'', which fragments the environment into incomplete observations. In practical scenarios, this limitation forces robots to frequently reorient themselves to gather sufficient environmental information, thereby increasing time costs and computational memory burdens. Furthermore, the lack of a holistic view means agents often miss critical environmental cues or potential interaction targets located in their periphery or rear, leading to inefficient task planning and suboptimal decision-making.
 
Panoramic cameras offer a natural solution to this bottleneck. By capturing a $360^\circ$ FoV in a single shot, they preserve global spatial relationships and enable holistic scene understanding. Currently, panoramic vision is widely used to help agents ``see'' the global environment, such as in spatial navigation~\cite{huang2022360vo, chen2024360orb} and scene understanding~\cite{360scene, zheng2026panorama}. However, its potential to help agents ``act'' through affordance reasoning is still largely unexplored. Moving affordance prediction into a panoramic view is a critical step forward for embodied AI. 

Bridging this gap, we make the \textbf{\textit{first exploration}} into the task of \textbf{\textit{Panoramic Affordance Prediction}}. First, to address the absence of suitable benchmarks in this emerging domain, we introduce \textbf{\dataset}, a high-resolution, large-scale benchmark dataset designed for panoramic affordance prediction. As shown in Fig.~\ref{fig:teaser}, \dataset~ comprises over 1,000 images with ultra-high 12K resolution (11904$\times$5952), captured across hundreds of scenes that span 12 distinct categories, such as daily life, work, and entertainment. \dataset~are also carefully annotated with over 12k image-question-answer pairs along with corresponding affordance masks, covering diverse objects and tasks. Moreover, we have carefully designed the dataset to incorporate the unique challenges of  $360^\circ$ ERP imagery, including geometric distortion, extreme scale variations, and boundary discontinuity.
Extensive experiments on the existing state-of-the-art (SoTA) affordance prediction methods inherently designed for standard perspective images reveal that they suffer severe performance degradation and largely fail when directly applied to panoramic environments. These pipelines struggle to effectively handle the unique challenges posed by  $360^\circ$ imagery, particularly the ultra-high resolution and severe spatial distortions.

To tackle these challenges, we propose \textbf{\method}, a training-free, coarse-to-fine pipeline inspired by the human foveal visual system. 
\method mimics how humans visually scan a broad scene before focusing on a specific target. Specifically, the framework operates in three logical stages: (i) To handle extreme scale variations and resolution burdens, we introduce \textit{Recursive Visual Routing via Grid Prompting}. This progressively guides Vision-Language Models (VLMs) to efficiently locate the general area of target tools. (ii) Once the target region is localized, an \textit{Adaptive Gaze} mechanism steps in. It projects this specific spherical region onto a tailored perspective plane, effectively eliminating the geometric distortions and boundary discontinuities inherent to ERP images. (iii) Finally, with a distortion-free local patch secured, a \textit{Cascaded Affordance Grounding} module deploys robust 2D visual foundation models to extract precise, instance-level affordance masks. On our proposed \dataset, \method~overcomes the unique challenges of  $360^\circ$ imagery without the need for specialized panoramic fine-tuning, and achieves SoTA performance, underscoring the immense potential of omnidirectional perception for advancing embodied AI.
In summary, our main contributions are threefold:
\begin{itemize}[leftmargin=1.6em]
    \item[\ding{111}] \textbf{New Task Formulation:} To the best of our knowledge, we make the first exploration into the task of \textbf{Panoramic Affordance Prediction}. 
    
    \item[\ding{111}]\textbf{Pioneering Benchmark:} We introduce \textbf{\dataset}, a large-scale, ultra-high-resolution (12K) benchmark dataset explicitly designed for panoramic affordance prediction. It provides rich annotations while encapsulating the unique challenges of $360^\circ$ ERP imagery.
    
    \item[\ding{111}] \textbf{Novel Framework \& SoTA Performance:} We propose \textbf{\method}, a training-free, fovea-inspired pipeline. Our method effectively overcomes challenges of panoramic images, achieving state-of-the-art performance.
\end{itemize}
\section{Related Work}

\noindent\textbf{Affordance Prediction:}
The concept of affordance~\cite{gibson1977theory} defines how embodied agents interact with objects in dynamic physical environments. Predicting affordance is pivotal for downstream embodied tasks, bridging the gap between perception and manipulation by informing task planning~\cite{mo2021where2act,wu2022vat}, object interaction~\cite{hoi-1, affordance-diffusion}, and tool usage~\cite{zhang2025phystoolbench,huang2024manipvqa, umd}. Consequently, it has been widely integrated into numerous robotic systems~\cite{rt-affordance,afford2act,a0,vla-r1}.
Early learning approaches~\cite{luo2022learning,3doi,li2023locate,jang2024intra} typically formulated the problem as a regression task to predict masks or heatmaps; however, they were often limited to task-specific inferences, lacking robust reasoning capabilities and generalization. With the advent of Vision-Language Models (VLMs)~\cite{gpt-4o,gemini,yang2025qwen3}, recent methods~\cite{affordancellm,wu2025ragnet,affordance-r1} have leveraged fine-tuning strategies to interpret complex task instructions and generate corresponding affordance maps. Most recently, A4-Agent~\cite{zhang2025a4} proposed an agentic framework that decouples reasoning and grounding, achieving strong performance in both areas.

Despite these advancements, existing affordance research relies predominantly on single-frame imagery from pinhole cameras. This restricted ``tunnel vision'' fragments the environment, forcing robots to perform task planning without a holistic understanding of the scene. Consequently, critical spatial context and potential interaction targets outside the current view are often missed, limiting the applicability of current methodologies in complex real-world scenarios.

\vspace{2mm}
\noindent\textbf{Panoramic Vision:}
Panoramic images offer a 360$^{\circ}$ field of view, fundamentally distinguishing them from the limited perspective of pinhole cameras. By capturing the entire environment in a single shot, panoramic sensors preserve complete spatial context and global geometric relationships. Despite inherent processing challenges like distortion and scale variation~\cite{lin2025one}, the advantage of ``seeing everything at once'' has made panoramic vision a crucial enabler for scene understanding~\cite{lin2025one,zhou2025dense360}, virtual reality~\cite{linhqgs}, autonomous driving~\cite{qi2019amodal}, and embodied AI~\cite{wan2025rapid}.
In Embodied Intelligence, this omnidirectional capability effectively eliminates blind spots. For autonomous navigation and SLAM, it allows agents to maintain stable tracking of landmarks and build consistent maps without frequent reorientation~\cite{chen2024360orb,wang2025panogen++}. Beyond navigation, 360$^{\circ}$ visual inputs significantly enhance an agent's ability in manipulation tasks~\cite{kerr2025eye} and facilitate precise localization~\cite{huang2024360loc}. Furthermore, integrating panoramic vision with multimodal large language models has opened new avenues for spatial reasoning and question answering~\cite{dongfang2025multimodal,zhang2025towards}, achieving a level of environmental completeness that pinhole-based systems cannot match.

Despite the promising applications of panoramic vision in embodied intelligence, its potential for affordance reasoning remains largely unexplored. To bridge this gap, we dive into the field of panoramic affordance prediction. By leveraging the holistic 360$^{\circ}$ context, we aim to empower embodied agents with a comprehensive understanding of action possibilities across the entire environment, addressing the critical limitations of narrow-field observations.
\section{\dataset}
\label{sec:dataset}

\subsection{Dataset Overview}
\label{subsec:dataset_overview}

\begin{table}[t]
    \centering
    \caption{Comparison of our proposed \dataset~with existing affordance and panoramic datasets. Some Dataset has various resolutions, we adopt the largest resolution for comparison.}
    \vspace{-1em}
    \label{tab:dataset_comparison}
    \renewcommand\arraystretch{1.2}
    \resizebox{\textwidth}{!}{
    \begin{tabular}{l| >{\centering\arraybackslash}m{4.5cm} >{\centering\arraybackslash}m{1.8cm} >{\centering\arraybackslash}m{1.25cm} >{\centering\arraybackslash}m{2.5cm} cc}
    \hlineB{2.5}
    \rowcolor{CadetBlue!20} 
    \textbf{Dataset} & \textbf{Affordance Type} & \textbf{\# Images} & \textbf{\# QAs} & \textbf{Resolution} & \textbf{FoV} & \textbf{Data Source} \\
    \hlineB{1.5}
    \multicolumn{7}{c}{\textit{Affordance Datasets}} \\
    \hline
    \rowcolor{gray!10}
    UMD~\cite{umd} & Action Category & 30k & - & $640 \times 480$ & $\sim 50^\circ$ & Real-world Capture \\
    ReasonAff~\cite{affordance-r1} & Instruction \& Reasoning & 3k & 3k & $2000 \times 1500$ & $\sim 30^\circ$ & Web Crawled \\
    \rowcolor{gray!10}
    HANDAL~\cite{guo2023handal} & Action Category & 200k & - & $1920 \times 1440$ & $\sim 50^\circ$ & Real-world Capture \\
    3DOI~\cite{3doi} & Action Category & 10k & - & $1920 \times 1080$ & $\sim 70^\circ$ & Real-world Capture \\
    \rowcolor{gray!10}
    RAGNet~\cite{wu2025ragnet} & Instruction \& Reasoning & 273k & 273k & $1920 \times 1440$ & $\sim 50^\circ$ & Web Crawled \\
    \hline
    \multicolumn{7}{c}{\textit{Panoramic Datasets}} \\
    \hline
    \rowcolor{gray!10}
    SUN360~\cite{sun360} & - & 67.6k & - & $9104 \times 4552$ & $360^\circ$ & Multi-view Stitching \\
    Matterport3D~\cite{chang2018matterport3d} & - & 10.8k & - & $2048 \times 1024$ & $360^\circ$ & Multi-view Stitching \\
    \rowcolor{gray!10}
    Stanford2D3D~\cite{armeni2017joint} & - & 1.4k & - & $4096 \times 2048$ & $360^\circ$ & Multi-view Stitching \\
    ReplicaPano~\cite{replicapano} & - & 2.7k & - & $1024 \times 512$ & $360^\circ$ & Synthetic Rendering \\
    \rowcolor{gray!10}
    DeepPanoContext~\cite{zhang2021deeppanocontext} & - & 1.5k & - & $1024 \times 512$ & $360^\circ$ & Synthetic Rendering \\
    Thinking-in-360~\cite{yu2025thinking} & - & 3k & - & $7680 \times 3840$ & $360^\circ$ & Web Crawled \\
    \rowcolor{gray!10}
    Realsee3D~\cite{Li2025realsee3d_data} & - & 10k & - & $1600 \times 800$ & $360^\circ$ & Multi-view Stitching \\
    \hline
    \multicolumn{7}{c}{\textit{Panoramic + Affordance Datasets}} \\
    \hline
    \rowcolor{gray!10}
    \textbf{\dataset~(Ours)} & \textbf{Instruction \& Reasoning} & \textbf{1k} & \textbf{13k} & \boldmath{$11904 \times 5952$} & \boldmath{$360^\circ$} & \textbf{Native 360$^\circ$ Camera} \\
    \hlineB{2.5}
    \end{tabular}
    }
    \vspace{-5mm}
\end{table}

\begin{figure}[t]
    \centering
    \includegraphics[width=1\linewidth]{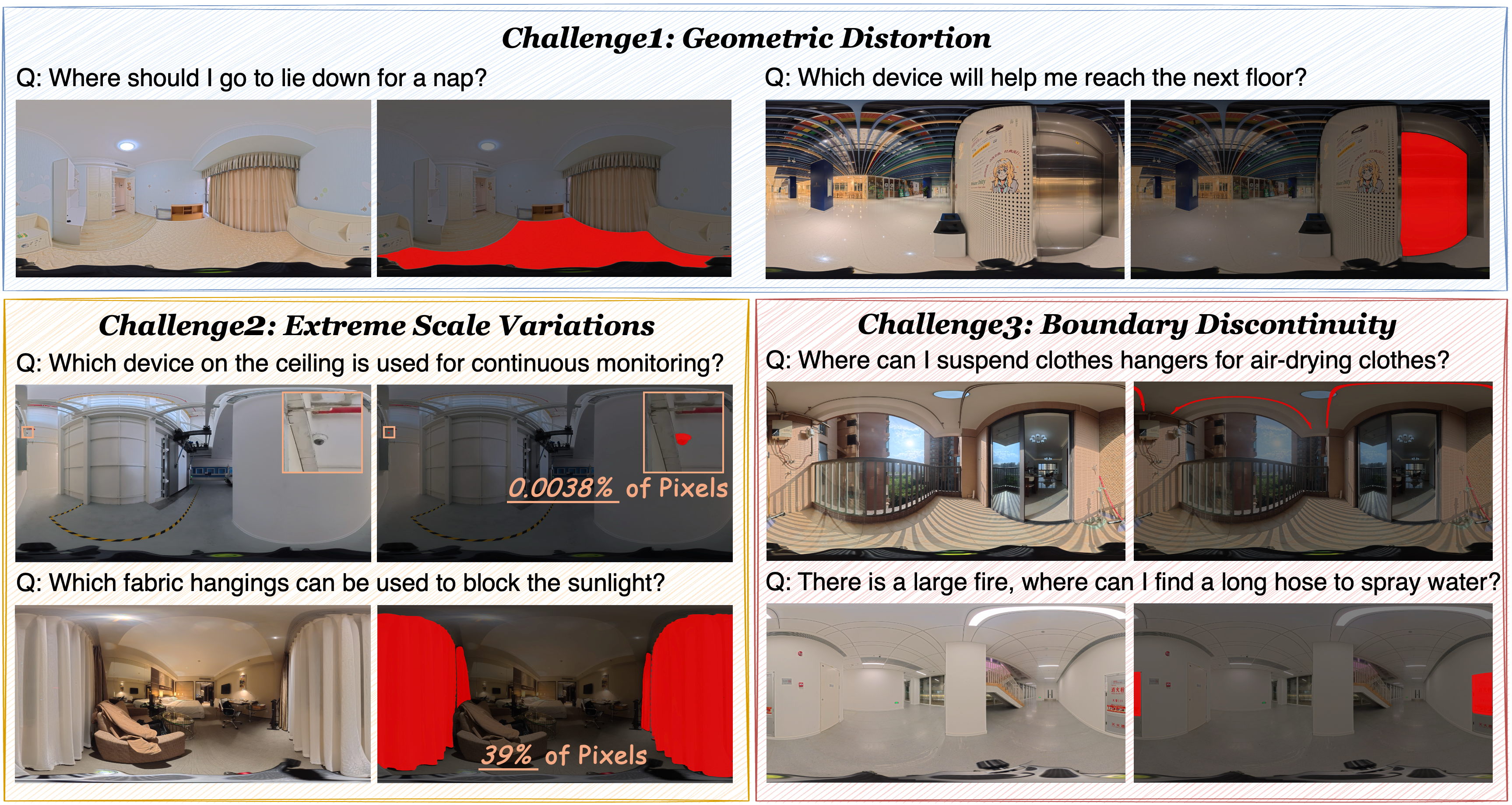}
    \vspace{-6mm}
    \caption{\dataset~specifically features three challenges inherent to $360^{\circ}$ panoramic imagery and ERP: (1) Geometric Distortion (e.g., the bed and the elevator); (2) Extreme Scale Variations (e.g., the extremely small security camera and the extremely large curtain); (3) Boundary Discontinuity (e.g., the drying rod and the fire hose).}
    \label{fig:challenge}
\end{figure}

We introduce \dataset, the first and large-scale benchmark dataset dedicated to panoramic affordance prediction. 
\dataset~has three key features:

\noindent\textbf{\scalebox{1.15}{\ding{172}}~Ultra-High Resolution:} To support fine-grained affordance analysis, our images are captured at \textbf{an ultra-high resolution of} \boldmath{$11904\times5952$}. As shown in Table~\ref{tab:dataset_comparison}, this resolution not only far exceeds that of existing affordance datasets, but also surpasses standard 360$^\circ$ panoramic datasets. Such a collection of high-quality panoramic images itself provides a valuable resource for the research community. Furthermore, unlike most panoramic datasets relying on multi-view stitching or synthetic rendering from RGB images, \dataset~is natively captured using professional 360$^\circ$ cameras in real-world environments. This authentic capturing process makes the data much closer to downstream applications, while allowing for the detection of small objects and subtle affordance cues that are often lost in lower-quality imagery.

\unboldmath

\begin{figure}[t]
    \centering
    \includegraphics[width=\linewidth]{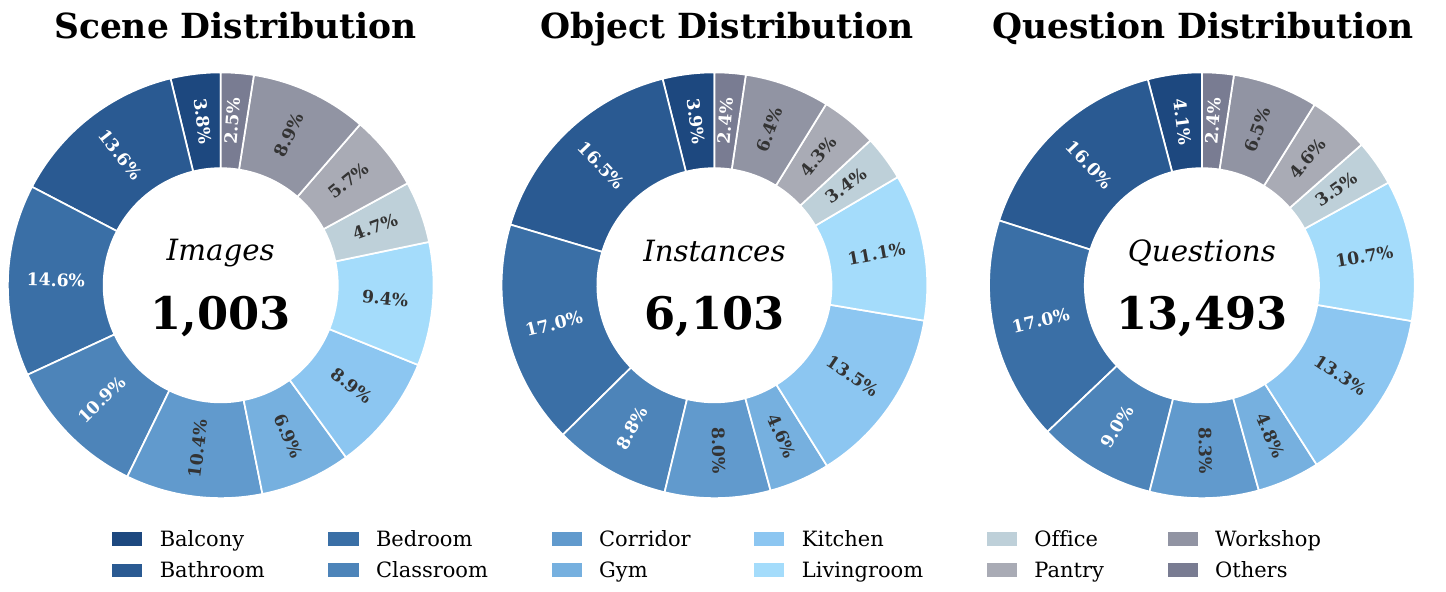}
    \vspace{-3mm}
    \caption{ \textbf{Statistics of the \dataset.} (Left) Scene distribution with 1,003 high-resolution panoramic images; (Middle) Object distribution featuring 6,103 annotated object instances; (Right) Question distribution comprising 13,493 affordance questions.}
    \label{fig:dataset_stats}
    \vspace{-1.5em}
\end{figure}

\begin{wrapfigure}{r}{0.45\textwidth}
    \vspace{-3.8mm}
    \centering
   \includegraphics[width=0.45\textwidth]{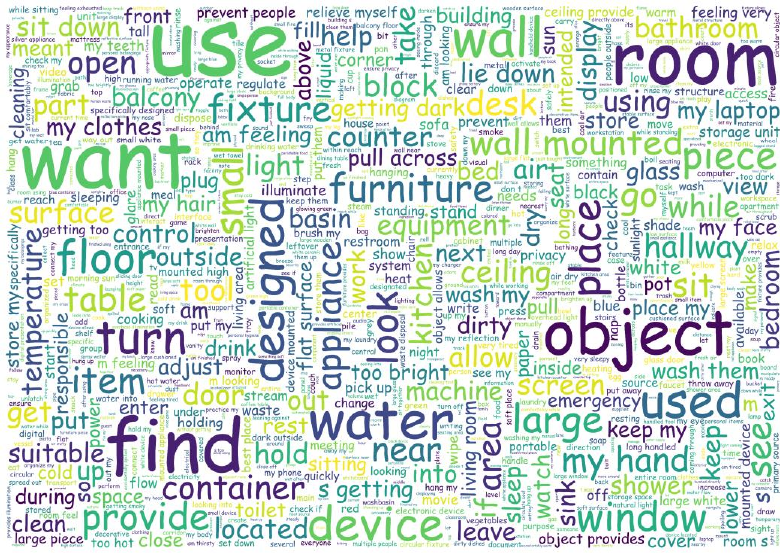}
   \vspace{-6mm}
   \caption{\centering Word Cloud of Questions in \dataset. }
\vspace{-2em}
\end{wrapfigure}
\noindent\textbf{\scalebox{1.15}{\ding{173}}~Rich QA Pairs and Masks:} We provide over 13,000 carefully annotated question-answer pairs. Aligning with the latest advancements in affordance datasets (e.g., ReasonAff~\cite{affordance-r1} and RAGNet~\cite{wu2025ragnet}), our QA pairs are specifically designed to require complex reasoning rather than simple perception. Each pair is associated with a precise pixel-level segmentation mask, grounding the affordance in the visual domain. This moves beyond simple classification to require explicit logical deduction and precise localization.

\noindent\textbf{\scalebox{1.15}{\ding{174}}~Panoramic-Specific Challenges:} A key design philosophy of \dataset~ is to explicitly incorporate challenges inherent to 360$^{\circ}$ panoramic imagery and Equirectangular Projection (ERP). As shown in Fig.~\ref{fig:challenge}, \dataset~includes: \textit{1) Geometric Distortion:} Objects in ERP images suffer from severe stretching, particularly near the poles. \dataset~ includes varied affordance targets with high distortion to evaluate model robustness. 
\textit{2) Extreme Scale Variations:} Because panoramas capture an unconstrained 360$^{\circ}$ environment, the visual scale of interactive objects fluctuates drastically. While some objects appear prominent, others occupy an exceptionally small proportion of the total pixels. \dataset~carefully curates this massive range of scales—with a particular focus on these minute, sub-scale targets—to test a model's ability to localize fine-grained details within a massive global context.
\textit{3) Boundary Discontinuity:} In the ERP format, continuous objects or surfaces are often split at the left and right image boundaries. So we have also included some cases where affordance regions wrap around these boundaries, requiring models to reason about continuity rather than treating the image as a static 2D plane. 

\subsection{Dataset Statistics}

As illustrated in Fig.~\ref{fig:dataset_stats}, we present the comprehensive statistics of the \dataset~dataset, categorizing the data across 12 diverse indoor scene types. In total, the dataset comprises 1,003 ultra-high-resolution panoramic images, 6,103 annotated object instances, and 13,493 affordance-related questions. The distribution demonstrates a rich variety of everyday environments. Notably, primary living spaces such as Bedrooms, Bathrooms, Kitchens, and Livingrooms constitute a significant portion of the data across all three hierarchical perspectives (images, instances, and questions). Such a data distribution is intentionally designed to align with the core operational domains of future home assistants and general-purpose robots, bridging the gap between static dataset evaluation and practical downstream robotic applications. This well-distributed and diverse collection therefore provides a robust foundation for training models capable of complex, real-world panoramic affordance reasoning in human-centric environments.

\subsection{Dataset Construction}

\noindent\textbf{Panoramic Image Collection.}
All panoramic images were captured using the Insta360-X5, a state-of-the-art professional high-resolution panoramic camera that can capture images at 12K resolution. To ensure high fidelity and simulate realistic robot viewpoints, we employed a professional tripod for stabilization and utilized delayed shooting to eliminate human obstruction from the field of view. Notably, we randomly adjusted the tripod height across different data collection sessions. This variation shifts the camera's equatorial plane, resulting in varying degrees of geometric distortion for objects at different elevations, which poses a realistic challenge for visual perception. For each scene (e.g., the same bedroom), we captured 2 to 4 different images by varying the camera's angle and height, as well as randomly shuffling object arrangements. Consequently, through extensive staging across hundreds of scenes, we collected over 1,000 high-quality, object-rich panoramic images.

\vspace{2mm}
\noindent\textbf{Annotation Process.}
With the high-quality panoramic images collected, the next step involves generating question-answer pairs and corresponding segmentation masks. We decompose this annotation process into two distinct phases. \textit{1) Question Formulation:} The first phase aims to construct valid action/task instruction-object pairs based on the visual content. To streamline this process, we developed an automated agent powered by a Multimodal Large Language Model (MLLM) to generate candidate pairs. While the agent produces a diverse set of instructions, some generated pairs may contain inaccuracies. Therefore, we conducted rigorous multi-round manual verification to ensure high quality and eliminate ambiguity. \textit{2) Mask Segmentation:} The second phase focuses on obtaining precise segmentation masks for the identified objects. 
This process was done manually with a specifically designed annotation tool.
\textit{3) Final Verification:} The final step involves a comprehensive review of the generated question-answer pairs and segmentation masks to ensure high quality and eliminate ambiguity. 
\section{\method~}
\label{sec:method}
\begin{figure}[t]
    \centering
    \includegraphics[width=1\linewidth]{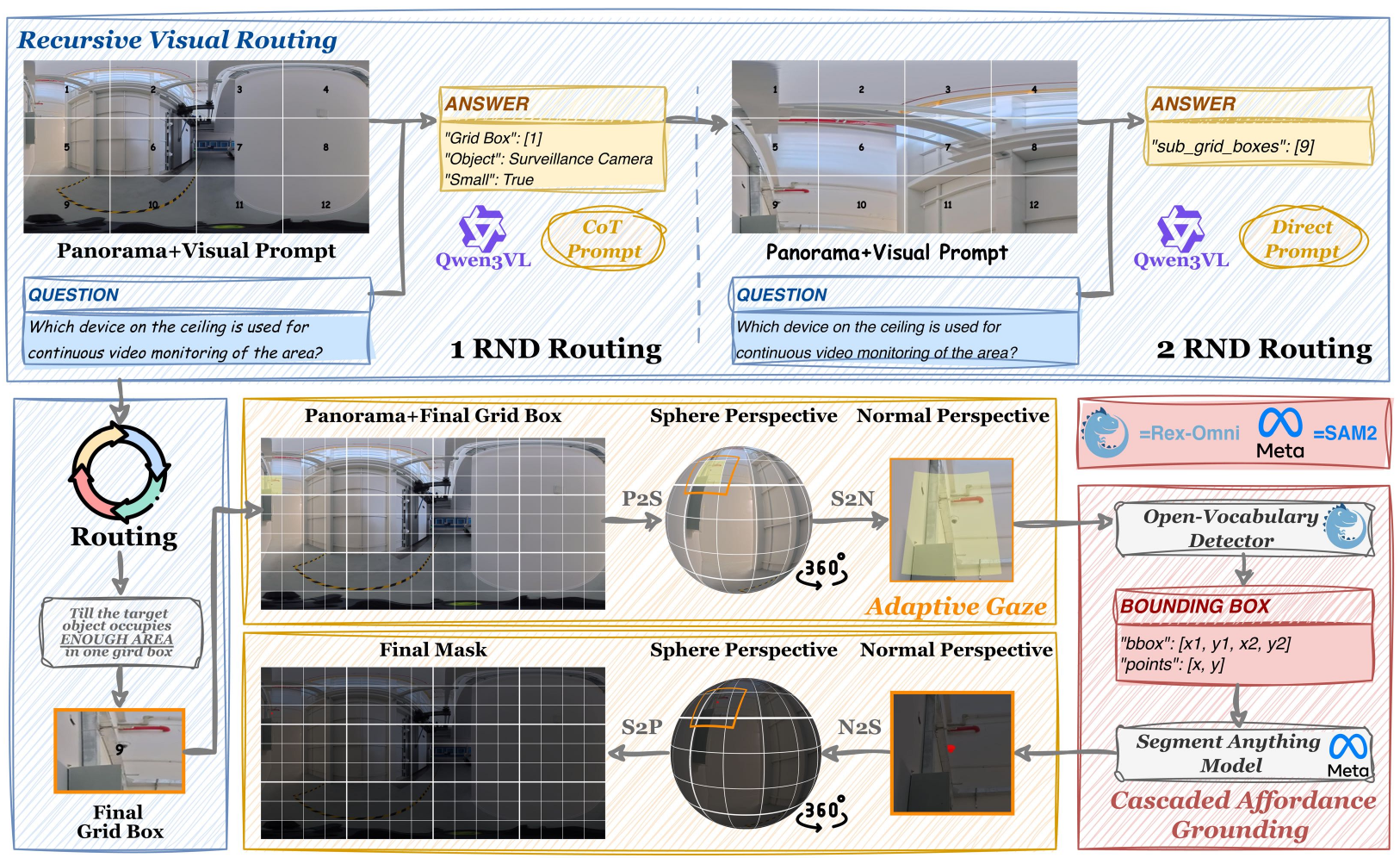}
    \vspace{-5mm}
    \caption{\centering \textbf{Illustration of the \method~framework.}}
    \vspace{-5mm}
    \label{fig:method}
\end{figure}
\subsection{Overview}
\label{subsec:overview}
Panoramic images offer rich global context but present two critical challenges for affordance prediction: the ultra-high resolution hinders VLMs from precisely capturing fine-grained targets, and Equirectangular Projection (ERP) distortions create a severe domain gap that degrades foundation model performance. 
To tackle these, we propose \textbf{\method}, a training-free, coarse-to-fine pipeline inspired by the human foveal visual system. Instead of processing 360$^{\circ}$ scenes with uniform acuity, humans use peripheral vision to locate regions of interest, direct their gaze for a clear view, and perform detailed parsing. Mirroring this, \method~operates in three stages (Fig.~\ref{fig:method}): \textit{Recursive Visual Routing via Grid Prompting} (Sec.~\ref{subsec:recursive_prompting}) leverages VLMs for coarse spatial localization. Then, \textit{Adaptive Gaze} (Sec.~\ref{subsec:rectification}) eliminates geometric distortions to bridge the domain gap. Finally, a \textit{Cascaded Affordance Grounding} pipeline (Sec.~\ref{subsec:segmentation}) deploys an Open-Vocabulary Detector (OVD) and the Segment Anything Model (SAM) within the rectified patch for precise instance-level segmentation.

\subsection{Recursive Visual Routing via Grid Prompting}
\label{subsec:recursive_prompting}
While existing VLMs possess strong semantic reasoning capabilities for deducing implicit affordances (i.e., inferring \textit{what} tool is needed for a task), they inherently struggle with precise spatial grounding, especially in high-resolution panoramic images due to token length constraints and attention dilution. To compensate for this grounding deficiency and avoid forcing the VLM to regress explicit continuous coordinates, we introduce a \textbf{Visual Grid Prompting} mechanism. Given an input ERP image $\mathcal{I}_{ERP}$ and a task description $\mathcal{T}$, we overlay a $4 \times 3$ numerical grid (indexed 1 to 12) onto the image. The VLM is then tasked to output a concise textual object description $\mathcal{T}_{obj}$ that can be used to do the task $\mathcal{T}$, and select the specific grid index containing the target.This visual prompt effectively transforms the complex continuous localization task into a discrete, multi-modal multiple-choice question, successfully bridging the gap between semantic reasoning and spatial grounding.

However, a single-pass grid selection is often insufficient for tiny objects within a massive 360$^{\circ}$ view. To address the extreme scale variations of objects, we further propose \textbf{Recursive Visual Routing} that guides the VLM to dynamically ``zoom in'' based on the target's relative scale: \ding{182} Scale-Salient Targets: 
If the target spans multiple grids, the current resolution is sufficient, and the recursion terminates. \ding{183} \textit{Sub-Scale Targets:} If confined to a single grid, the object is too small. We crop this grid, overlay a new grid, and infer recursively.
Crucially, to ensure computational efficiency, input images are aggressively downsampled during this recursion. Initially, the global panorama is severely downsampled to extract broad contextual cues. After this first step, the system recursively zooms into smaller cropped grids and reduces the downsampling ratio. This guided localization efficiently filters out irrelevant 360° backgrounds, securing a reliable coarse bounding region for subsequent modules without the massive overhead of exhaustive high-resolution processing.

\subsection{Adaptive Gaze}
\label{subsec:rectification}
Once the local grid containing the target is coarsely isolated, directly giving precise segmentation masks is still challenging. This is due to the pronounced challenges inherent in the ERP format (as discussed in Sec.~\ref{subsec:dataset_overview}), which creates a significant domain gap between panoramic images and standard perspective images. 
To bridge this gap, we introduce the \textbf{Adaptive Gaze}, simulating the human action of turning one's head to focus directly on a target. This mechanism adaptively adjusts both its viewing direction and scope to match the routed grid. Specifically, we first direct the \textit{``gaze''} by aligning the camera's principal point (point of tangency) with the latitude and longitude $(\phi_c, \theta_c)$ of the grid's center. Then, we adjust the \textit{``focus''} by adaptively scaling the Field of View (FoV) based on the grid's location and dimensions to perfectly enclose the target region. By projecting this local spherical region onto a tangent plane, we yield a perspective image $\mathcal{I}_{persp}$ tailored specifically to the target.

This spherical-to-perspective projection acts as a \textbf{training-free domain adapter} that elegantly resolves key panoramic challenges: 1) It eliminates \textit{geometric distortion} by mapping the local curved surface to a flat tangent plane. 2) The adaptive FoV naturally mitigates \textit{extreme scale variations} by magnifying sub-scale targets to standard resolutions and maintaining fine-grained details. 3) It seamlessly handles \textit{boundary discontinuity} as the projection operates intrinsically on the continuous spherical manifold; severed targets on the 2D ERP images are reconstructed as whole objects in the perspective crop. By overcoming these inherent ERP limitations, we ensure seamless integration with powerful 2D visual foundation models without the need for panoramic-specific fine-tuning.

\begin{table}[t]
    \centering
    \caption{\centering Comparison on \dataset. \textbf{Best} results and \underline{Second-best} results are highlighted in bold and underline.}
    \label{tab:performance}
    \renewcommand{\arraystretch}{1.35} 
    \resizebox{\linewidth}{!}{
        \begin{tabular}{l| *{4}{>{\centering\arraybackslash}m{1.8cm}}| >{\centering\arraybackslash}m{2.5cm}}
            \hlineB{2.5}
            \rowcolor{CadetBlue!20} 
            \textbf{Method} &\textbf{gIoU$\uparrow$}  & \textbf{cIoU$\uparrow$} & $\mathbf{P_{50}\uparrow}$ & $\mathbf{P_{50-95}\uparrow}$ & \textbf{Inference Time} \\
            \hlineB{1.5}
            OV-Seg~\cite{ovseg}         & 29.48 & 17.85 & 32.00 & 18.80 & $\sim$8s \\
            \rowcolor{gray!10}
            LISA~\cite{lai2024lisa}           & 15.21 & 16.34 & 13.66 & 8.30 & $\sim$7s \\
            VisionReasoner~\cite{visionreasoner} & 49.33 & 44.64 & 51.06 & 38.06 & $\sim$12s\\
            \rowcolor{gray!10}
            AffordanceVLM~\cite{wu2025ragnet}  & 9.66 & 13.11 & 8.96 & 5.41 & $\sim$7.8s \\
            Affordance-R1~\cite{affordance-r1}  & 51.80 & \underline{50.32} & 55.47 & 40.70 & $\sim$10.4s \\
            \rowcolor{gray!10}
            A4-Agent~\cite{zhang2025a4}       & \underline{62.55} & 49.97 & \underline{67.09} & \underline{54.28} & $\sim$11.8s \\
            \hline
            \textbf{\method~(Ours)}        & \textbf{71.56} & \textbf{62.30} & \textbf{75.49} & \textbf{64.97} & $\sim$10s \\
            \hlineB{2.5}
        \end{tabular}
    }
\end{table}

\subsection{Cascaded Affordance Grounding}
\label{subsec:segmentation}
With the reasoning and coarse localization handled by the Recursive Visual Routing, and the geometric distortion eliminated by the Adaptive Gaze, the pipeline proceeds to pixel-level grounding. Following recent success in decoupling 2D affordance grounding into distinct reasoning and perception modules~\cite{zhang2025a4}, we deploy a cascaded pipeline on the high-resolution local perspective image $\mathcal{I}_{persp}$.

We first deploy an Open-Vocabulary Detector (OVD) to parse the deduced explicit object description $\mathcal{T}_{obj}$ (generated by the Recursive Visual Routing) and scan the refined perspective image $\mathcal{I}_{persp}$ to generate a bounding box $\mathcal{B}$ and key points $\mathcal{P}$. By utilizing the specific object name rather than the complex, implicit task description $\mathcal{T}$, and because the Recursive Visual Routing has already filtered out the vast majority of the 360$^{\circ}$ background, the OVD is relieved of the immense burden of semantic guessing and searching the entire complex scene. This results in significantly higher detection accuracy and fewer false positives.
Subsequently, this bounding box $\mathcal{B}$ and key points $\mathcal{P}$ serve as a dense spatial prompt for the SAM. Leveraging SAM's robust zero-shot segmentation capabilities, we extract a highly accurate, instance-level segmentation mask $\mathcal{M}_{persp}$ along the target's boundaries. 
Finally, to fulfill the end-to-end panoramic affordance prediction task, the predicted perspective mask $\mathcal{M}_{persp}$ is re-mapped back to the original ERP space via an inverse perspective-to-spherical projection transformation, yielding the final panoramic mask $\mathcal{M}_{ERP}$. This synergistic design naturally breaks down the formidable 360$^{\circ}$ grounding problem into manageable stages, playing precisely to the strengths of each foundation model.

\section{Experiments}
\label{sec:experiments}

\subsection{Experimental Settings}

\noindent\textbf{Baseline Methods:}
To the best of our knowledge, our method is the first framework dedicated to panoramic affordance prediction. Due to the lack of direct panoramic baselines, we validate the effectiveness of our approach by benchmarking it against state-of-the-art methods designed for perspective imagery, including A4-Agent~\cite{zhang2025a4}, Affordance-R1~\cite{affordance-r1}, and AffordanceVLM~\cite{wu2025ragnet}. Furthermore, following the evaluation protocols in \cite{zhang2025a4,affordance-r1}, we extend our comparison to include general Open Vocabulary Segmentation models, such as Vision Reasoner~\cite{visionreasoner},OVSeg~\cite{ovseg}, and LISA~\cite{lai2024lisa}.

\begin{figure}[t]
    \centering
    \includegraphics[width=1\textwidth]{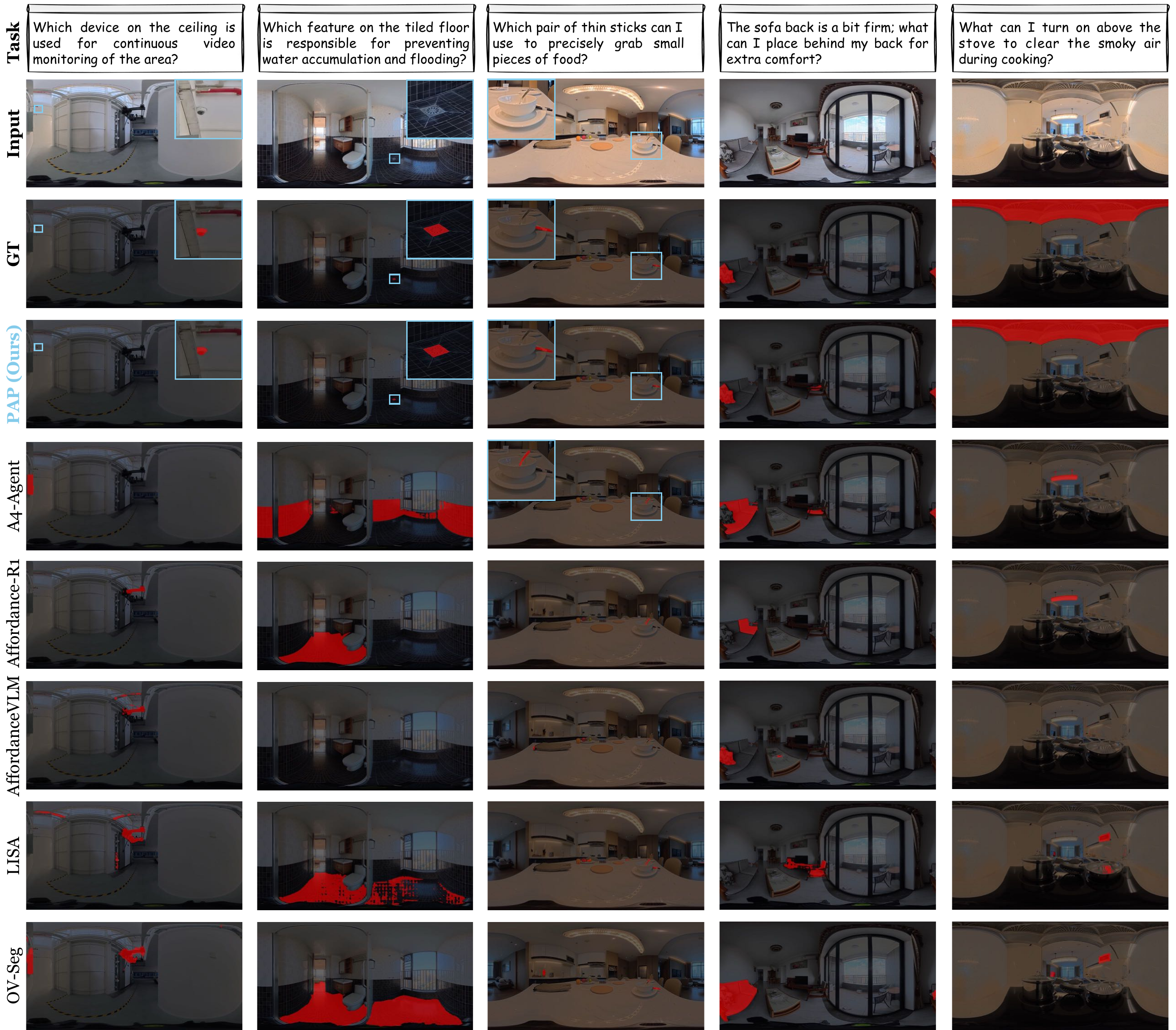}
    \vspace{-2em}
    \caption{Qualitative comparison on \dataset. We highlight the predicted affordance regions in red. The extremely small objects are enlarged for better visualization.}
    \label{fig:main_results}
\end{figure}

\noindent\textbf{Evaluation Metrics:}
Following the standard evaluation protocol in affordance prediction~\cite{affordance-r1,zhang2025a4}, we adopt four complementary metrics to comprehensively assess prediction quality, including:
\textit{1) gIoU (Generalized IoU):} The average Intersection-over-Union across all test samples, measuring the overall segmentation quality of the predicted affordance regions.
\textit{2) cIoU (Cumulative IoU):} The cumulative intersection over cumulative union across the entire dataset, providing a dataset-level quality measure that is less sensitive to the size of individual objects.
\textit{3) P@50 (Precision at IoU=0.5):} The percentage of predictions with an IoU score exceeding 0.5, evaluating the model's ability to generate high-quality predictions.
\textit{4) P@50:95:} The average precision calculated across a range of IoU thresholds from 0.5 to 0.95 with 0.05 increments, providing a stricter and more fine-grained assessment of segmentation accuracy.

\noindent\textbf{Implementation Details:}
For our \method~framework, we adopt Qwen3-VL-32B~\cite{bai2025qwen3} as the backbone of our VLM, and following~\cite{zhang2025a4}, we adopt Rex-Omni~\cite{rex-omni} as the OVD model and the SAM-2-Large~\cite{ravi2024sam} as the Segmentation model. During the Recursive Visual Routing stage, we apply the dynamic resolution adaptation strategy to balance computational cost and visual details. Specifically, in the first routing round, the full panorama is downsampled to $2000 \times 1000$ (a $\sim1/6$ scaling ratio) before being fed into the VLM. If the target is sub-scale and requires a second routing round, the cropped grid region is downsampled to $1500 \times 1000$ (a $\sim1/2$ scaling ratio). This design elegantly balances computational efficiency and visual performance, ensuring that as the routing progresses into smaller regions, the VLM receives images with progressively higher relative pixel fidelity without exceeding strict token constraints.

\begin{table}[t]
    \centering
    \caption{Comparison on different difficulty levels of \dataset. \textbf{Best} results are highlighted in bold. \underline{Second-best} results are highlighted in underline.}
    \vspace{-1em}
    \label{tab:performance:main}
    \renewcommand{\arraystretch}{1.2} 
    \resizebox{\linewidth}{!}{
    \begin{tabular}{l|*{4}{>{\centering\arraybackslash}m{1.4cm}}|*{4}{>{\centering\arraybackslash}m{1.4cm}}}
    \hlineB{2.5}
    \multirow{2}{*}{\textbf{Method}} & \multicolumn{4}{c}{\textbf{\dataset-Hard}} & \multicolumn{4}{c}{\textbf{\dataset-Normal}} \\
    \cmidrule(lr){2-5} \cmidrule(lr){6-9}
    &gIoU$\uparrow$  & cIoU$\uparrow$ & $P_{50}$$\uparrow$ & $P_{50-95}$$\uparrow$ &gIoU$\uparrow$  & cIoU$\uparrow$ & $P_{50}$$\uparrow$ & $P_{50-95}$$\uparrow$ \\
    \hlineB{1.5}
    \rowcolor{CadetBlue!10} 
    OV-Seg~\cite{ovseg} & 10.66 & 9.18 & 10.32 & 4.90 & 37.66 & 21.72 & 41.44 & 24.84 \\
    LISA~\cite{lai2024lisa} & 4.75 & 13.04 & 3.23 & 2.01 & 19.75 & 17.24 & 18.20 & 11.03 \\
    \rowcolor{CadetBlue!10} 
    VisionReasoner~\cite{visionreasoner} & 25.96 & 29.74 & 24.91 & 15.00 & 59.50 & 49.59 & 62.44 & 48.10 \\
    AffordanceVLM~\cite{wu2025ragnet} & 4.94 & 15.94 & 3.59 & 1.92 & 11.72 & 12.36 & 11.30 & 6.93 \\
    \rowcolor{CadetBlue!10} 
    Affordance-R1~\cite{affordance-r1} & 33.31 & \underline{40.84} & 36.26 & 21.39 & 59.85 & 52.97 & 63.82 & 49.11 \\
    A4-Agent~\cite{zhang2025a4} & \underline{42.75} & 36.42 & \underline{46.48} & \underline{30.38} & \underline{70.91} & \underline{54.94} & \underline{75.79} & \underline{64.36} \\
    \hline
    \rowcolor{CadetBlue!10} 
    \textbf{\method~(Ours)} & \textbf{60.35} & \textbf{52.59} & \textbf{63.82} & \textbf{52.17} & \textbf{76.40} & \textbf{65.20} &\textbf{ 80.52} & \textbf{70.49} \\
    \hlineB{2.5}
    \end{tabular}
    }
\end{table}

\subsection{Main Results}

\paragraph{\textbf{Overall Performance.}} Table~\ref{tab:performance} presents the overall quantitative comparison between our proposed \method~and state-of-the-art baselines. \method~ significantly outperforms all existing methods across all evaluation metrics by a large margin. Specifically, \method~ achieves a gIoU of 71.56\% and a cIoU of 62.30\%, surpassing the second-best method, A4-Agent, by absolute margins of 9.01\% and 12.33\%, respectively. In terms of precision, our method obtains 75.49\% on $P_{50}$ and 64.97\% on $P_{50-95}$, demonstrating its superior capability in precise affordance localization and segmentation compared to recent strong baselines. The substantial improvements indicate that our approach effectively captures complex panoramic affordance relationships. Fig.~\ref{fig:main_results} presents several representative cases. As illustrated, our method robustly accomplishes the task even in challenging scenarios—such as when objects are excessively large or small, split across boundaries, or severely distorted—whereas other methods struggle.

\paragraph{\textbf{Performance on Different Difficulty Levels.}} Furthermore, we quantitatively evaluate the performance of different models under varying difficulty levels. We partition \dataset~into \textit{Hard} and \textit{Normal} subsets based on the following criteria: 1) The object is exceptionally large or small, with the mask occupying more than 30\% or less than 0.1\% of the entire image. 2) The object is truncated across the left and right boundaries, resulting in the mask appearing on both sides of the image. This process identifies approximately 30\% of the cases as the \textit{Hard} subset, leaving the remainder as the \textit{Normal} subset. Table~\ref{tab:performance:main} illustrates the comparison across these difficulty levels. Notably, in the \textit{Hard} subset, our method exhibits an even more pronounced advantage, outperforming the second-best method (A4-Agent) by absolute margins of 17.60\% and 16.17\% in terms of gIoU and cIoU. In terms of precision, \method~leads A4-Agent by absolute margins of 17.34\% and 21.79\% on $P_{50}$ and $P_{50-95}$. This demonstrates that our method effectively addresses the inherent challenges of panoramic ERP images such as extreme scale variation and boundary discontinuity.

\subsection{Ablation Studies}
To validate the effectiveness of our approach, we conducted extensive ablation studies. For computational efficiency, we randomly sampled 10\% of the \dataset~for these experiments. We evaluate three key components: Prompt Style, Recursive Visual Routing, and Adaptive Gaze. In this section, we analyze the overall impact of integrating versus omitting each module. Detailed hyperparameter analyses and additional analytical experiments are deferred to the Appendix (Sec.~\ref{appendix:more_analytical_study}).

\paragraph{\textbf{Impact of Prompt Style.}}
We investigate the impact of different prompting strategies, specifically \textit{Chain-of-Thought (CoT)} and the proposed \textit{Visual Grid Prompting (VGP)}. As shown in Table~\ref{tab:albation:prompt}, incorporating CoT reasoning steadily improves the model's performance, demonstrating that step-by-step logical deduction aids in complex spatial prediction. More importantly, the explicit visual grid overlay proves to be an even more crucial component. When replacing the visual grid with a purely textual spatial description (e.g., verbally describing the image divisions to the VLM). Comparing Row 1 vs.\ Row 2, and Row 3 vs.\ Row 4, we observe a drastic performance drop across all metrics. This confirms that while VLMs possess strong semantic reasoning capabilities, they struggle to map abstract textual spatial divisions to complex visual features in ultra-high-resolution panoramas without explicit visual anchors. Our Visual Grid Prompting successfully bridges this gap by providing concrete reference points, effectively grounding the VLM's reasoning process into a manageable multi-modal discrete choice task.

\paragraph{\textbf{Impact of Recursive Visual Routing.}}
To evaluate the efficacy of our Recursive Visual Routing (RVR) module, we compare the full model against a baseline that relies solely on a single-step grid routing across different difficulty subsets. As shown in Table~\ref{tab:ablation:rvr}, integrating RVR yields consistent improvements on the entire dataset. More importantly, the performance gains are significantly more pronounced on the ``Hard'' subset, where cIoU and gIoU surge by 12.30\% and 5.03\%, respectively, compared to the modest increases of 4.50\% and 0.78\% on the ``Normal'' subset. This disparity demonstrates that due to the extreme object scale variations inherent in $360^\circ$ scenes, a single-step routing approach tends to produce excessively large localization regions and struggles to precisely locate challenging targets. Conversely, our RVR module overcomes this limitation by iteratively zooming in on the target, ensuring robust coarse localization even for the most difficult objects regardless of their scales.

\begin{table*}[t]
    \centering
    \begin{minipage}[t]{0.48\textwidth}
        \centering
        \renewcommand{\arraystretch}{1.2}
        \caption{\centering Ablation on Prompt Style.}
        \vspace{-1.5mm}
        \label{tab:albation:prompt}
        \resizebox{\linewidth}{!}{
        \begin{tabular}{*{2}{>{\centering\arraybackslash}m{1.0cm}}|*{4}{>{\centering\arraybackslash}m{1.6cm}}}
            \hlineB{2.5}
            \textbf{CoT} & \textbf{VGP} & \textbf{gIoU$\uparrow$}  & \textbf{cIoU$\uparrow$} & $\mathbf{P_{50}\uparrow}$ & $\mathbf{P_{50-95}\uparrow}$  \\ 
            \hlineB{1.5}
            \ding{55}   & \ding{55}   & 57.70 & 57.68 & 60.76 & 50.52 \\ 
            \ding{55}   & \ding{51} & 69.56 & 61.05 & 74.02 & 61.90 \\ 
            \ding{51} & \ding{55}   & 67.22 & 58.75 & 70.61 & 60.51 \\ 
            \rowcolor{CadetBlue!20} 
            \ding{51} & \ding{51} & 72.69 & 63.85 & 76.29 & 66.13 \\ 
            \hlineB{2.5}
        \end{tabular}
        }

        \vspace{1mm}

        \renewcommand{\arraystretch}{1.2}
        \caption{\centering Ablation on Adaptive Gaze.}
        \vspace{-3mm}
        \label{tab:ablation:ag}
        \resizebox{\linewidth}{!}{
        \begin{tabular}{c|*{4}{>{\centering\arraybackslash}m{1.6cm}}}
            \hlineB{2.5}
            \textbf{Method} & \textbf{gIoU$\uparrow$} & \textbf{cIoU$\uparrow$} & $\mathbf{P_{50}\uparrow}$ & $\mathbf{P_{50-95}\uparrow}$  \\
            \hlineB{1.5}
            w/o AG & 64.99 & 55.43 & 68.43 & 56.37 \\
            w/ AG   & 72.69 & 63.85 & 76.29 & 66.13 \\
            \rowcolor{CadetBlue!20} 
            $\Delta$  & 7.70  & 8.42  & 7.77  & 9.76  \\
            \hlineB{2.5}
        \end{tabular}
        }
    \end{minipage}
    \hfill
    \begin{minipage}[t]{0.48\textwidth}
        \centering
        \renewcommand{\arraystretch}{1.26}
        \caption{\centering Ablation on RVR.}
        \vspace{-1.5mm}
        \label{tab:ablation:rvr}
        \resizebox{\linewidth}{!}{
        \begin{tabular}{c|>{\centering\arraybackslash}m{1.5cm}| *{4}{>{\centering\arraybackslash}m{1.31cm}}}
            \hlineB{2.5}
            \textbf{Subset} & \textbf{Method} & \textbf{gIoU$\uparrow$} & \textbf{cIoU$\uparrow$} & $\mathbf{P_{50}\uparrow}$ & $\mathbf{P_{50-95}\uparrow}$  \\
            \hlineB{1.5}
            \multirow{3}{*}{All}  
            & w/o RVR & 70.56 & 57.05 & 74.43 & 63.23 \\
            & w/ RVR  & 72.69 & 63.85 & 76.29 & 66.13 \\
            \rowcolor{CadetBlue!20} 
            & $\Delta$  & 2.13  & 6.80  & 1.86  & 2.90  \\
            \hline
            \multirow{3}{*}{Hard} 
            & w/o RVR & 57.37 & 34.81 & 61.00 & 47.65 \\
            & w/ RVR  & 62.40 & 47.11 & 64.89 & 53.64 \\
            \rowcolor{CadetBlue!20} 
            & $\Delta$  & 5.03  & 12.30 & 3.89  & 5.99  \\
            \hline
            \multirow{3}{*}{Normal} 
            & w/o RVR & 76.29 & 63.55 & 80.26 & 69.99 \\
            & w/ RVR  & 77.07 & 68.05 & 81.15 & 71.45 \\
            \rowcolor{CadetBlue!20} 
            & $\Delta$  & 0.78  & 4.50  & 0.89  & 1.46  \\
            \hlineB{2.5}
        \end{tabular}
        }
    \end{minipage}
    \vspace{-3mm}
\end{table*}

\paragraph{\textbf{Impact of Adaptive Gaze.}}
A core insight of our pipeline is the training-free domain adaptation via Adaptive Gaze (AG). We ablate this module by directly applying the OVD and SAM on the cropped ERP regions without AG. The results show a significant decline in accuracy. This validates that the inherent spatial distortions of ERP formats introduce a severe domain shift that disrupts the pre-trained priors of 2D foundation models. Our AG successfully eliminates this distortion, seamlessly aligning the data domains.

\section{Conclusion}
\label{sec:conclusion}

In this paper, we present the first exploration into \textbf{Panoramic Affordance Prediction}, bridging the critical gap between holistic 360$^\circ$ scene understanding and actionable intelligence in embodied AI. We first introduce \dataset, a pioneering, large-scale benchmark featuring over 1,000 natively captured ultra-high-resolution (12K) panoramic images and more than 12K reasoning-based QA pairs with precise affordance masks. Furthermore, to address the profound challenges of panoramic vision, we propose \method, a training-free framework inspired by the human foveal visual system. 
Extensive experiments demonstrate that \method~significantly outperforms existing state-of-the-art methods, particularly in highly challenging scenarios. We hope our dataset and framework will serve as a foundational stepping stone, inspiring future research toward the combination of panoramic vision and embodied intelligence.

\bibliographystyle{plainnat}
\bibliography{main}

\clearpage
\appendix

\renewcommand{\theHsection}{supp.\thesection}
\renewcommand{\theHsubsection}{supp.\thesubsection}
\renewcommand{\theHsubsubsection}{supp.\thesubsubsection}

\begin{center}
    \Large\bfseries Appendix
\end{center}
\vspace*{1em}

\startcontents[supp]
\begingroup
    \setlength{\parskip}{0.8em} 
    \definecolor{eccvblue}{rgb}{0.12,0.49,0.85}
    \hypersetup{linkcolor=black}

    \printcontents[supp]{}{1}{\section*{Contents}\setcounter{tocdepth}{2}}
\endgroup
\vspace*{2em}

\clearpage
\section{More Analytical Study}
\label{appendix:more_analytical_study}
While the main text validates the efficacy of each proposed module via extensive ablation studies, this section provides further empirical analyses to comprehensively illustrate the findings from our experimental process. 
\subsection{Superparameter Analysis of Visual Prompt}

\begin{wrapfigure}{r}{0.5\textwidth}
    \centering
    \vspace{-1.5em}
    \captionof{table}{Performance comparison of different visual prompt types. \colorbox{CadetBlue!20}{Row 1} indicates the default setting used in our final pipeline.}
    \label{tab:ablation-grid-type}
    \renewcommand{\arraystretch}{1.5}
    \resizebox{\linewidth}{!}{
    \begin{tabular}{>{\centering\arraybackslash}m{0.6cm}|>{\centering\arraybackslash}m{3.5cm}|*{4}{>{\centering\arraybackslash}m{1.3cm}}}
        \hlineB{2.5}
        \textbf{\#} & \textbf{Visual Prompt Type} & \textbf{gIoU$\uparrow$} & \textbf{cIoU$\uparrow$} & $\mathbf{P_{50}\uparrow}$ & $\mathbf{P_{50-95}\uparrow}$ \\
        \hline
        \rowcolor{CadetBlue!20}
        1 & Line                 & \textbf{72.69} & \textbf{63.85} & \textbf{76.29} & \textbf{66.13} \\
        2 & Color ($\alpha=50$)  & 70.91 & 59.75 & 74.92 & 64.58 \\
        3 & Color ($\alpha=100$) & 69.96 & 57.98 & 74.09 & 63.76 \\
        4 & Color ($\alpha=150$) & 68.67 & 57.46 & 71.48 & 62.70 \\
        \hlineB{2.5}
    \end{tabular}
    }

    \vspace{1.5em}

    \includegraphics[width=\linewidth]{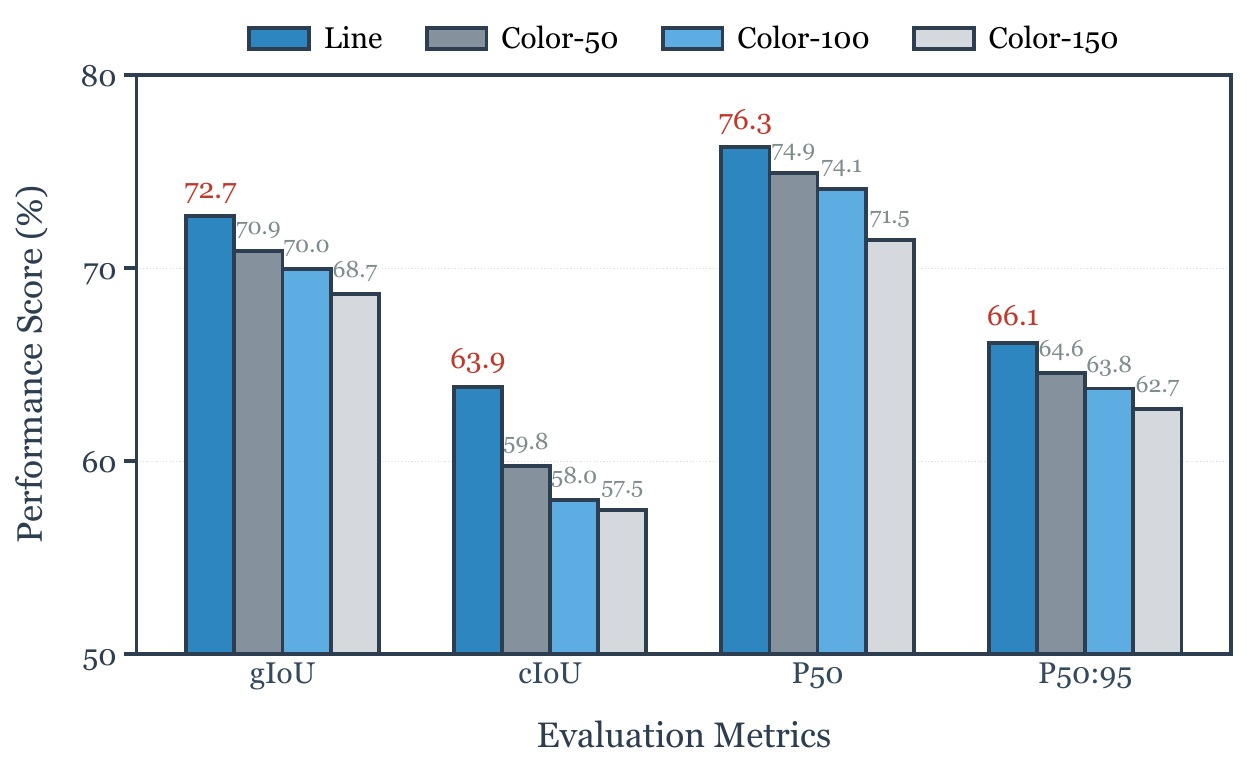}
    \captionof{figure}{Performance comparison of different visual prompting types. \textcolor{red}{Red} highlights the best results for each metric.}
    \vspace{-1em}
    \label{fig:ablation-grid-type}
\end{wrapfigure}

\begin{figure}[h]
    \centering
    \includegraphics[width=0.8\linewidth]{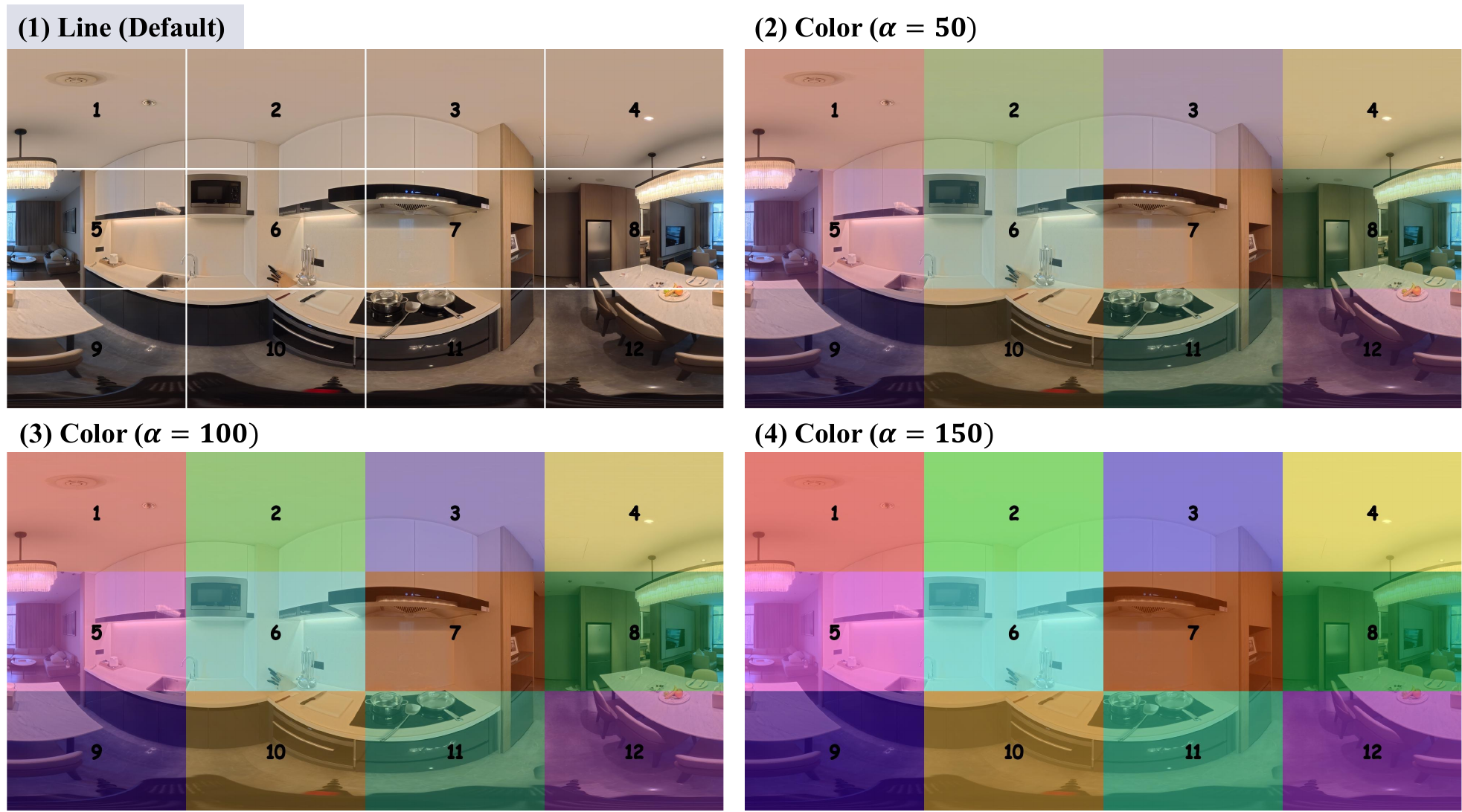}
    \caption{\centering Visualization of Different Visual Prompt in Table~\ref{tab:ablation-grid-type}.}
    \label{fig:visual-grid-type}
    \vspace{-1.5em}
\end{figure}

\paragraph{\textbf{Grid Prompting Style: Lines vs. Color Blocks}}
We explore different visual prompting styles to effectively demarcate the spatial cells for the VLM. 
The primary comparison is between utilizing a grid composed of distinct solid lines versus applying semi-transparent colored blocks (color-coding each grid cell differently). A visualization is shown in Fig.~\ref{fig:visual-grid-type}. 
As detailed in Table~\ref{tab:ablation-grid-type} and Fig.~\ref{fig:ablation-grid-type}, the line-based prompt achieves the optimal performance across all metrics, yielding a gIoU of 72.69 and a cIoU of 63.85. Conversely, applying color blocks significantly degrades performance. 
As the opacity level ($\alpha$) of the color blocks increases from 50 to 150, the gIoU continuously drops from 70.91 to 68.67, and the cIoU drops from 59.75 to 57.46. This steady performance decline occurs because the color block overlay inevitably destroys the original color distribution of the underlying panoramic image. Such severe color distortion introduces unnatural tinting that confuses the VLM, severely interfering with its perception of the inherent texture, material, and color attributes critical for precise semantic deduction. In contrast, drawing simple grid lines minimally perturbs the original visual features of the scene, allowing the VLM to maintain robust recognition capability while successfully parsing the spatial layout.

\begin{wrapfigure}{r}{0.5\textwidth}
    \centering
    \captionof{table}{Performance comparison of different line widths and font sizes. \colorbox{CadetBlue!20}{Row 1} indicates the default setting used in our final pipeline.}
    \label{tab:line_font_performance}
    \renewcommand{\arraystretch}{1.25}
    \resizebox{0.95\linewidth}{!}{
    \begin{tabular}{>{\centering\arraybackslash}m{0.6cm}|>{\centering\arraybackslash}m{1.3cm} >{\centering\arraybackslash}m{1.3cm}|*{4}{>{\centering\arraybackslash}m{1.1cm}}}
        \hlineB{2.5} 
        \textbf{\#} & \textbf{Line Width} & \textbf{Font Size} & \textbf{gIoU$\uparrow$} & \textbf{cIoU$\uparrow$} & $\mathbf{P_{50}\uparrow}$ & $\mathbf{P_{50-95}\uparrow}$ \\
        \hlineB{2.5} 
        \rowcolor{CadetBlue!20}
        1 & \textbf{5} & \textbf{50} & 72.69 & \textbf{63.85} & \textbf{76.29} & \textbf{66.13} \\
        \hline 
        2 & 1  & \textbf{50} & 71.93 & 60.35 & 75.47 & 65.43 \\
        3 & 10 & \textbf{50} & 72.13 & 62.39 & 75.99 & 65.45 \\
        4 & 15 & \textbf{50} & 71.00 & 59.74 & 74.85 & 64.25 \\
        5 & 50 & \textbf{50} & 67.49 & 57.46 & 71.14 & 60.06 \\
        \hline 
        6 & \textbf{5} & 10  &\textbf{72.70} & 61.89 & 74.73 & 64.95 \\
        7 & \textbf{5} & 25  & 71.44 & 62.99 & 75.06 & 64.93 \\
        8 & \textbf{5} & 100 & 71.46 & 62.65 & 74.68 & 65.01 \\
        9 & \textbf{5} & 200 & 71.27 & 58.44 & 74.04 & 64.92 \\
        \hlineB{2.5} 
    \end{tabular}
    }
\end{wrapfigure}

\begin{figure}[t]
    \centering
    \includegraphics[width=\linewidth]{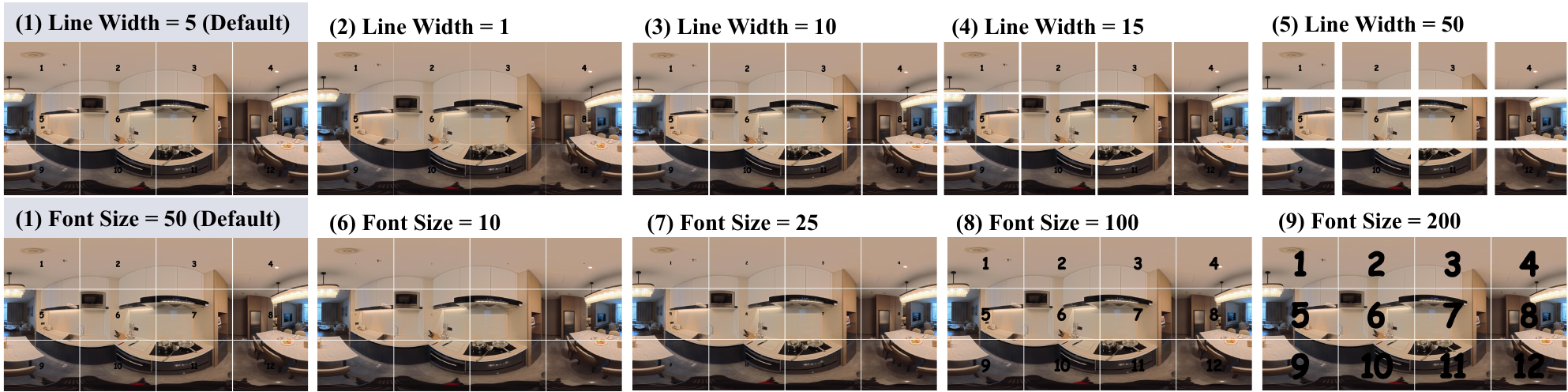}
    \caption{\centering Visualization of different line width and font size in Table~\ref{tab:line_font_performance}.}
    \label{fig:ablation-line-font}
\end{figure}

\begin{figure}[t]
    \centering
    \includegraphics[width=\linewidth]{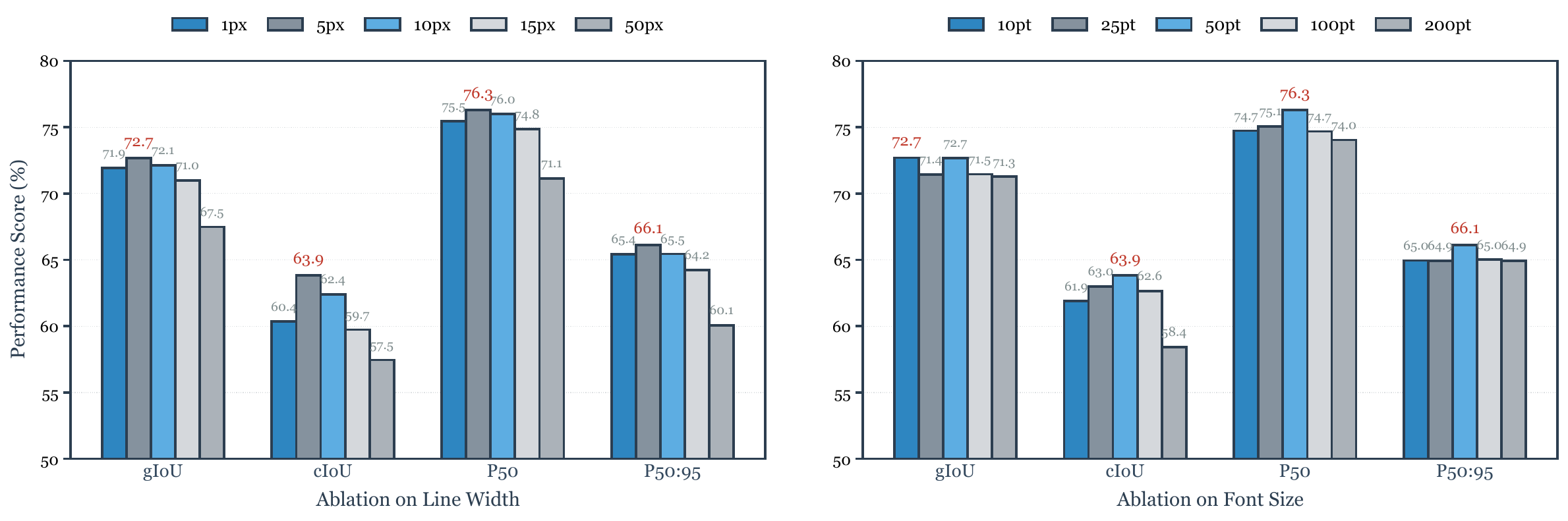}
    \vspace{-6mm}
    \caption{\centering Ablation study on line width and font size.  \textcolor{red}{Red} highlights the best results.}
    \label{fig:ablation-lf-type}
\end{figure}

\paragraph{\textbf{Grid Line Width}}
We evaluate the impact of grid line width by drawing the visual prompt on downsampled panoramic images of $2000 \times 1000$ resolution. As shown in Table~\ref{tab:line_font_performance} and Fig.~\ref{fig:ablation-lf-type}, our pipeline exhibits strong robustness to this parameter. Interestingly, even an extremely thin line of just 1 pixel (accounting for merely 0.1\% of the image height) is highly effective in helping the VLM improve its spatial perception, yielding a competitive gIoU of 71.93 and cIoU of 60.35. The performance remains stable and optimal within a moderate range, peaking at a line width of 5 pixels. However, when the line width becomes excessively wide, such as 50 pixels (spanning 5\% of the image height), the thick grid lines severely fragment the scene and occlude critical fine-grained visual information. This fragmentation causes a drastic performance drop, with gIoU falling to 67.49 and cIoU dropping to 57.46. Qualitative comparisons are provided in Fig.~\ref{fig:ablation-line-font}.

\paragraph{\textbf{Font Size of Grid Coordinates}}
Similarly, our pipeline demonstrates strong robustness to the font size of the numerical indices (1 to 12). As reported in Table~\ref{tab:line_font_performance}, varying the font size within a broad range (from 10 to 100) yields consistently stable results, with an intermediate font size of 50 (occupying about 5\% of the image height) achieving the optimal balance. However, akin to the line width degradation, excessively large text negatively impacts performance. For instance, when the font size is increased to 200 (spanning 20\% of the image height), the oversized numbers act as severe artificial occlusions. These massive digits mask important contextual cues and target objects in the underlying image, causing the cIoU to drop notably to 58.44. The visual impact of varying font sizes is also illustrated in Fig.~\ref{fig:ablation-line-font}.

\paragraph{\textbf{Analysis of Grid Resolution.}}
\begin{wrapfigure}{r}{0.5\textwidth}
    \vspace{-6mm}
    \centering
    \includegraphics[width=\linewidth]{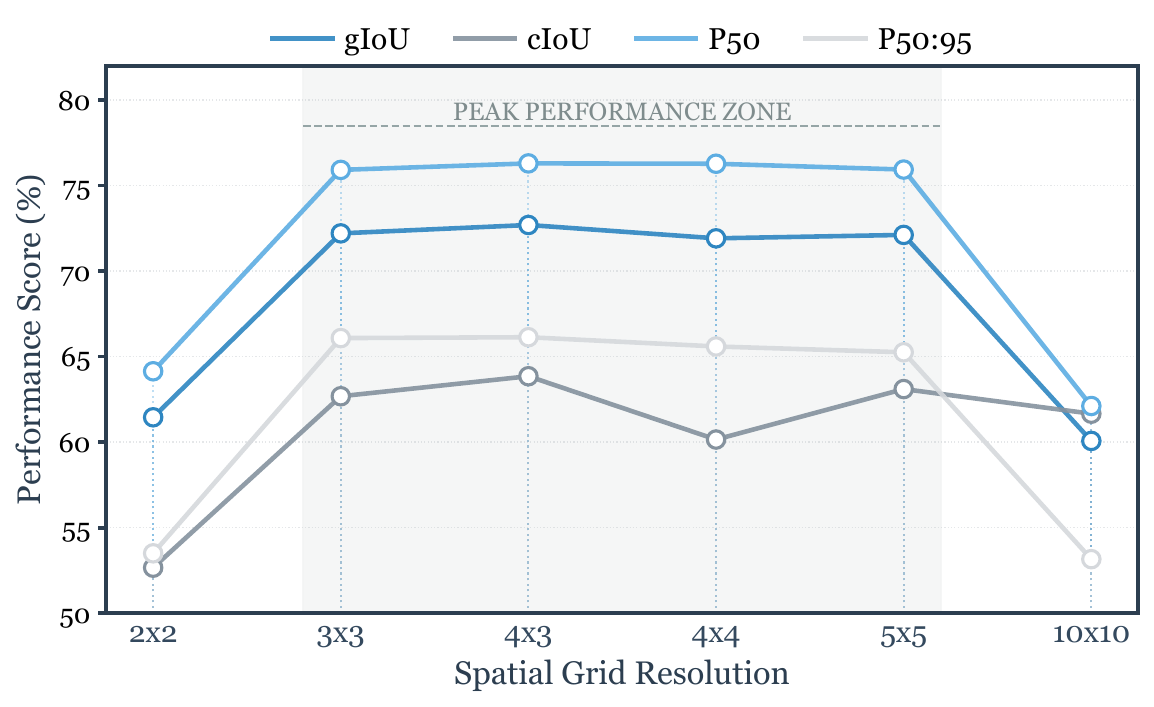}
    \vspace{-8mm}
    \captionof{figure}{Comparison between different grid resolutions.}
    \label{fig:ablation-grid}
    \vspace{-5mm}
\end{wrapfigure}
Furthermore, we analyze the impact of different visual grid resolutions. As shown in Table~\ref{tab:grid_resolution_performance} and Fig.~\ref{fig:ablation-grid}, the performance remains robust across a moderate range of resolutions, with $3 \times 3$, $4 \times 3$, $4 \times 4$, and $5 \times 5$ grids yielding comparable results. However, both overly coarse and excessively dense grids negatively impact performance. For instance, a $2 \times 2$ grid provides insufficient spatial constraints, offering limited assistance to the VLM. Conversely, a dense $10 \times 10$ grid containing 100 cells imposes a heavy burden on the VLM's reasoning capacity, leading to a noticeable performance drop. We adopt the $4 \times 3$ grid as our default setting since it achieves the best performance.

\begin{table}[t]
    \centering
    \begin{minipage}[t]{0.48\textwidth}
        \centering
        \caption{Performance comparison of different grid resolutions. \colorbox{CadetBlue!20}{Row 3} indicates the default setting used in our final pipeline.}
        \vspace{-2mm}
        \label{tab:grid_resolution_performance}
        \renewcommand{\arraystretch}{1.2}
        \resizebox{\linewidth}{!}{
        \begin{tabular}{>{\centering\arraybackslash}m{0.6cm}|>{\centering\arraybackslash}m{2.5cm}|*{4}{>{\centering\arraybackslash}m{1.5cm}}}
            \toprule
            \textbf{\#} & \textbf{Grid Resolution} & \textbf{gIoU$\uparrow$} & \textbf{cIoU$\uparrow$} & $\mathbf{P_{50}\uparrow}$ & $\mathbf{P_{50-95}\uparrow}$ \\
            \midrule
            1 & 2 $\times$ 2 & 61.45 & 52.66 & 64.14 & 53.50 \\
            2 & 3 $\times$ 3 & 72.20 & 62.68 & 75.91 & 66.08 \\
            \rowcolor{CadetBlue!20}
            3 & \textbf{4 $\times$ 3} & \textbf{72.69} & \textbf{63.85} & \textbf{76.29} & \textbf{66.13} \\
            4 & 4 $\times$ 4 & 71.91 & 60.15 & 76.27 & 65.59 \\
            5 & 5 $\times$ 5 & 72.11 & 63.10 & 75.93 & 65.25 \\
            6 & 10 $\times$ 10 & 60.07 & 61.66 & 62.11 & 53.16 \\
            \bottomrule
        \end{tabular}
        }
    \end{minipage}
    \hfill
    \begin{minipage}[t]{0.48\textwidth}
        \centering
        \caption{Performance comparison of \method~using different VLM models. \colorbox{CadetBlue!20}{Row 2} indicates the default setting used in our final pipeline.}
        \vspace{-2mm}
        \label{tab:vlm_model_comparison}
        \renewcommand{\arraystretch}{1.37}
        \resizebox{\linewidth}{!}{
        \begin{tabular}{>{\centering\arraybackslash}m{0.6cm}|>{\centering\arraybackslash}m{3cm}|*{4}{>{\centering\arraybackslash}m{1.1cm}}}
            \toprule
            \textbf{\#} & \textbf{VLM Model} & \textbf{gIoU$\uparrow$} & \textbf{cIoU$\uparrow$} & $\mathbf{P_{50}\uparrow}$ & $\mathbf{P_{50-95}\uparrow}$ \\
            \midrule
            1 & Gemini-3-Flash & 74.03 & 67.65 & 77.74 & 67.20 \\
            \rowcolor{CadetBlue!20}
            2 & Qwen-3-VL-32B & 72.69 & 63.85 & 76.29 & 66.13 \\
            3 & Qwen-3-VL-8B  & 68.93 & 55.57 & 72.43 & 62.34 \\
            4 & Qwen-3-VL-4B  & 66.28 & 52.79 & 70.15 & 59.94 \\
            5 & Qwen-2.5-VL-7B & 65.47 & 52.12 & 68.08 & 57.35 \\
            \bottomrule
        \end{tabular}
        }
    \end{minipage}
    \vspace{-4mm}
\end{table}

\subsection{Analysis of Different VLM backbones}

To evaluate the impact of the Vision-Language Model (VLM) backbone on our pipeline, we conduct experiments across models with varying capabilities and scales, as presented in Table~\ref{tab:vlm_model_comparison}. First, we observe a strong positive correlation between the overall performance metrics and the intrinsic capabilities of the VLM. Specifically, substituting our default setting, Qwen-3-VL-32B, with a more powerful, closed-source model (Gemini-3-Flash) yields substantial improvements across all metrics. Conversely, adopting smaller open-source models (such as Qwen-3-VL-8B/4B or Qwen-2.5-VL-7B) results in a corresponding decline in performance.

Furthermore, these results demonstrate the \textbf{\textit{high flexibility}} of our proposed pipeline. It can seamlessly integrate state-of-the-art closed-source models to pursue absolute performance, effectively serving as a robust offline data generation engine. Alternatively, for scenarios where computational efficiency is a priority, the pipeline can readily adopt more lightweight models to maintain a desirable balance between performance and inference cost.

\subsection{Analysis of Different Resolution}
\begin{wrapfigure}{r}{0.5\textwidth}
    \centering
    \vspace{-2mm}
    \captionof{table}{Performance comparison of different image resolutions. \colorbox{CadetBlue!20}{Row 2} indicates the default resolution used in our final pipeline.}
    \vspace{-2mm}
    \label{tab:image_resolution_performance}
    \renewcommand{\arraystretch}{1.4}
    \resizebox{\linewidth}{!}{
    \begin{tabular}{>{\centering\arraybackslash}m{0.6cm}|>{\centering\arraybackslash}m{2.5cm}|*{4}{>{\centering\arraybackslash}m{1.5cm}}}
        \hlineB{2.5}
        \textbf{\#} & \textbf{Resolution} & \textbf{gIoU$\uparrow$} & \textbf{cIoU$\uparrow$} & $\mathbf{P_{50}\uparrow}$ & $\mathbf{P_{50-95}\uparrow}$ \\
        \hline
        1 & 4000 $\times$ 2000 & \textbf{73.32} & 58.98 & \textbf{76.81} & \textbf{67.30} \\
        \rowcolor{CadetBlue!20}
        2 & 2000 $\times$ 1000 & 72.69 & \textbf{63.85} & 76.29 & 66.13 \\
        3 & 1000 $\times$ 500  & 66.46 & 56.45 & 70.78 & 59.18 \\
        \hlineB{2.5}
    \end{tabular}
    }
    \vspace{-2mm}
\end{wrapfigure}
To evaluate the impact of image resolution, we conducted experiments across three different settings: $4000 \times 2000$, $2000 \times 1000$, and $1000 \times 500$. As shown in Table~\ref{tab:image_resolution_performance}, the model exhibits relatively poor performance at $1000 \times 500$. This is likely due to the low resolution preventing the model from capturing sufficient fine-grained details. Conversely, increasing the resolution to $2000 \times 1000$ yields a significant performance boost. This aligns with our expectations, as $2000 \times 1000$ is visibly much sharper to the naked eye, providing adequate clarity for detail extraction. However, further scaling the resolution to $4000 \times 2000$ results in marginal gains; while metrics such as gIoU, $P_{50}$, and $P_{50-95}$ show slight improvements, cIoU actually declines. We attribute this to our Recursive Visual Routing mechanism, which effectively leverages a coarse-to-fine approach. By initially analyzing global context from a highly downsampled image and subsequently narrowing the field of view with reduced downsampling to capture finer details, the $2000 \times 1000$ resolution proves to be entirely sufficient for this paradigm. Furthermore, since the $4000 \times 2000$ resolution introduces four times the number of image tokens compared to $2000 \times 1000$, the inference latency is effectively doubled. Therefore, to achieve an optimal trade-off between performance and inference efficiency, we adopted $2000 \times 1000$ as our default configuration.

\section{Discussion}
\subsection{Acceptable but Unimpressive Latency}
As demonstrated in Table 2 in the main text, our \method~significantly outperforms other baselines while maintaining comparable inference latency(\underline{\textit{$\sim$10 seconds}}). However, for real-world robotic deployment, minimizing latency is always desirable. As shown in Table~\ref{tab:vlm_model_comparison}, we explored using more lightweight VLMs, such as Qwen3-VL-4B, as the backbone. Although this resulted in an approximate \underline{\textit{8\%}} performance drop, it still vastly exceeds the baseline methods and bounds the inference latency to \underline{\textit{2–3 seconds}}. We consider this latency acceptable for the high-level planning task of locating target objects within a massive and complex 360-degree environment based on intricate instructions. For future work, we plan to further explore latency reduction; for instance, we could leverage our proposed \method~for offline data generation to distill an smaller end-to-end model, thereby further minimizing inference time.
\subsection{End-to-End Model v.s. Decoupled Pipeline}
Following the success of prior work~\cite{zhang2025a4} that decouples the reasoning and grounding stages of affordance prediction, the proposed \method~adopts a similar paradigm. Compared to end-to-end solutions, the decoupled approach offers several key advantages: 1) greater flexibility in integrating various Vision-Language Models (VLMs); 2) the ability to fully exploit the specific strengths of individual models; 3) strong zero-shot generalization capabilities without the need for training on massive datasets; and 4) enhanced interpretability, as the intermediate outputs of each step are accessible, which facilitates debugging and optimization. As the first exploration of the panoramic affordance prediction task, our work embraces this decoupled strategy. The state-of-the-art performance of \method~on \dataset~demonstrates the effectiveness of this decoupled philosophy within the panoramic domain.

Nevertheless, we recognize the significant potential of end-to-end models for this task. While we have introduced several tailored designs to bridge the domain gap between panoramic and conventional camera imaging, a model capable of natively understanding Equirectangular Projection (ERP) images and efficiently inferring affordances would offer even greater value for downstream applications. Notably, our proposed framework intrinsically serves as a robust, zero-shot data generation engine. In future work, \method~can be utilized to further scale up the training data, paving the way for the highly efficient end-to-end models.
\begin{figure}[t]
    \centering
    \includegraphics[width=\linewidth]{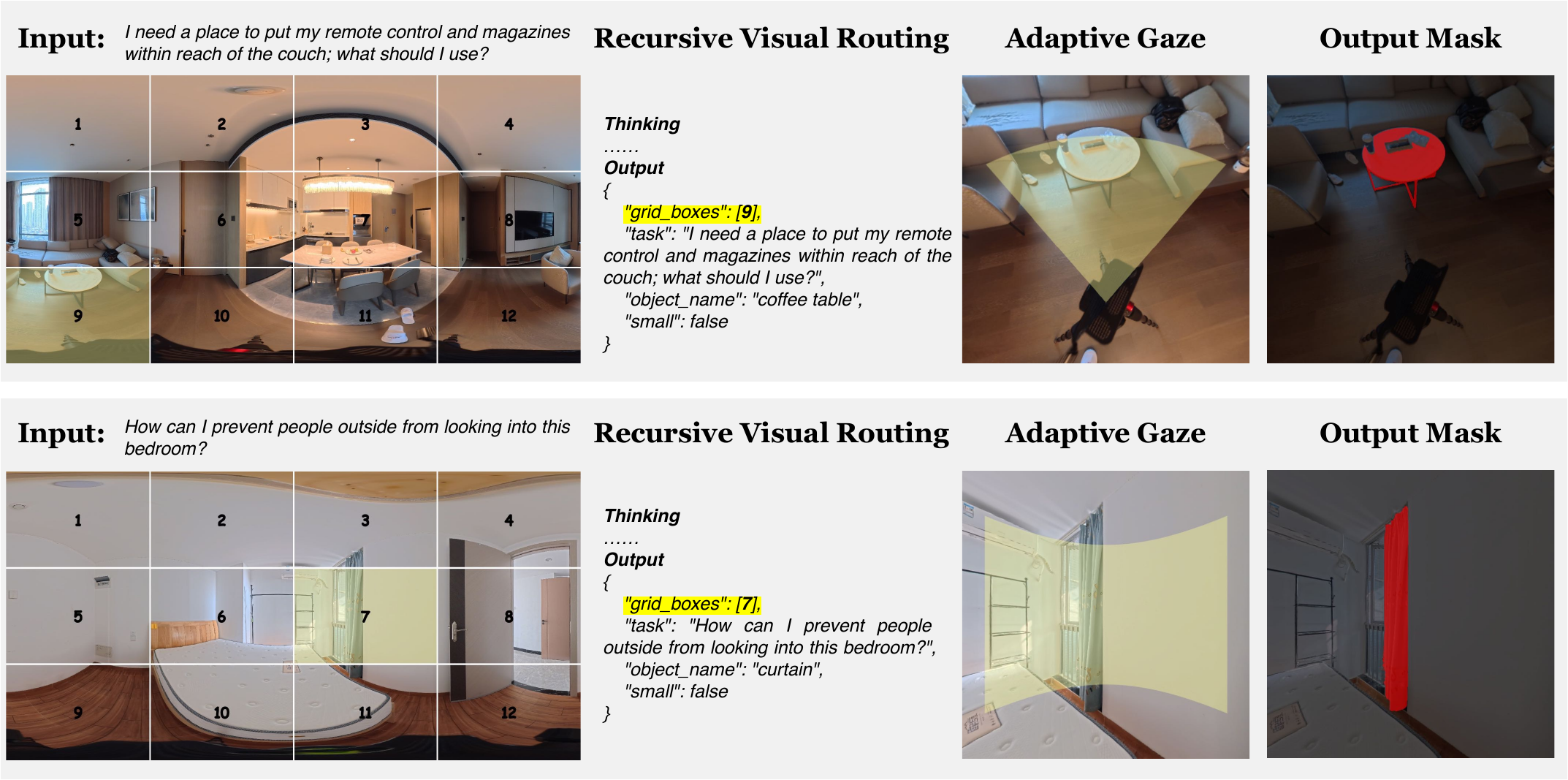}
    \caption{Visualization of the error cases of Recursive Visual Routing. We also visualize the coverage of the grid, where the yellow translucent areas show the areas covered by the grid in the ERP view and the FoV view. It can be seen that our Adaptive Gaze module effectively covers the entire grid containing the object, with a small margin of redundancy.}
    \label{fig:rvr-error}
\end{figure}
\subsection{Managing Cascading Errors}
While decoupled models offer numerous advantages over end-to-end models, a potential pitfall is the accumulation of errors. We accounted for this in our design, primarily by managing the errors generated by our Recursive Visual Routing module. We observed that the model outputs accurate grid indices in the vast majority of cases, but it is prone to minor errors in one specific scenario: when a small portion of an object happens to cross a grid boundary. In this situation, the model yields one of two possible outputs: 
\textit{1)} It accurately provides the indices of both grids the object spans. This occurs most of the time and poses no issue, allowing for direct progression to the next step.
\textit{2)} It outputs only one of the two grids. When this happens and the process advances to the adaptive gaze step, if the selected FoV (Field of View) only covers this single grid exactly, the object is very likely to be split in half, which compromises the subsequent grounding.

To effectively resolve this error, we proposed a solution: in the adaptive gaze module, we add a margin of redundancy (around 10 degrees) to the adaptive gaze's FoV. This way, when reverting to the normal view, the FoV covers a small area beyond the grid, ensuring that objects situated right at the boundary between two grids are not split. As shown in two cases in Fig.~\ref{fig:rvr-error}, the correct objects (coffee table in the top and curtain in the bottom) have an extremely small portion falling outside the grid. During the RVR process, the VLM fails to correctly output all the grids occupied by the object. However, our Adaptive Gaze module, by adding a small margin to the FoV, successfully resolves this error and allows the entire pipeline to produce the correct mask.

\section{More Implementation Details}

\subsection{System Prompt of Recursive Visual Routing}

\begin{renderedpromptbox}[System Prompt: Recursive Visual Routing]
    \scriptsize
    \label{prompt:rvr}
    \subsubsection{Environment Setting}\label{environment-setting}

Given a panoramic image of a scene, the task is to decide which object
to use and predict the part of the object that matches the provided
task. The task instruction is ``TASK''. To assist your spatial
reasoning, the image is overlaid with a \emph{4x3 grid marked with large
numbers 1 through 12}. The grid follows a standard reading order:
Top-Row (1, 2, 3, 4), Middle-Row (5, 6, 7, 8), and Bottom-Row (9, 10,
11, 12). These numbers serve as your spatial reference system.

\subsubsection{Your task}\label{your-task}

\begin{enumerate}
\def\labelenumi{\arabic{enumi}.}
\tightlist
\item
  Identify the target object in the 4x3 grid panoramic image according
  to the task instructions.
\item
  Accurately locate the grid boxes corresponding to the target object.
\item
  \emph{STRICTLY} follow the output format, especially the JSON format.
\end{enumerate}

\subsubsection{Follow these reasoning
steps:}\label{follow-these-reasoning-steps}

\begin{enumerate}
\def\labelenumi{\arabic{enumi}.}
\tightlist
\item
  \emph{Step 1 Identify the Target}

  \begin{itemize}
  \tightlist
  \item
    First identify the target to use from the scene that best matches
    the provided task.
  \item
    Then, identify the key components of this object in the image (e.g.,
    shape, features, possible points of interaction).
  \item
    Finally, analyze the object and the task instruction, provide the
    final answer ``object\_name'' or a more specific part of the object
    ``object\_part''.
  \item
    \emph{Rule}: If there are multiple similar objects in the image,
    please make sure your ``object\_name'' or ``object\_part'' is
    uniquely identifiable and clearly distinguishable from the other
    similar objects by adding additional descriptions.
  \end{itemize}
\item
  \emph{Step 2 Spatial Occupancy (Grid Mapping)}

  \begin{itemize}
  \tightlist
  \item
    Closely observe the \emph{4x3 grid numbering (1-12)}.
  \item
    List \emph{ALL} grid boxes that contain any part of the target
    object.
  \item
    \emph{Rule}: If the object's body or edges cross the edge line,
    include all grid boxes on both sides of the line.
  \end{itemize}
\item
  \emph{Step 3 Clarity Analysis (Small Object Refinement)}

  \begin{itemize}
  \tightlist
  \item
    When the target object occupies \emph{only ONE} grid box, judge
    whether the object appears \emph{small}.
  \item
    \emph{Rule}: An object is ``small'' compared to other normal
    objects, making it difficult to precisely segment for the task.
  \end{itemize}
\item
  \emph{Step 4 Output}

  \begin{itemize}
  \tightlist
  \item
    Output the result in a structured JSON format. \emph{STRICTLY}
    follow the output format.
  \end{itemize}
\end{enumerate}

\subsubsection{Output format:}\label{output-format}
\paragraph{Thinking}\label{thinking}
Thinking process 1. Identify the target object. 2. Identify the number
of grid boxes. 3. Identify whether the target object is small.
\paragraph{Output}\label{output}
\{ ``grid\_boxes'': {[}index1, index2, \ldots{]}, // e.g., {[}4, 5{]} or
{[}5{]} --- first-level 4×3 grid indices (1--12) ``task'': ``the task
instruction'', ``object\_name'': ``the name of the object'',
``object\_part'': ``the part of the object'', // if the whole object is
the target, leave it the same as ``object\_name'' ``small'': true/false
\}
 
\end{renderedpromptbox}

\subsection{Formulation of the Adaptive Gaze Module}

To extract a distortion-free rectilinear image (viewport) from an Equirectangular Projection (ERP) panorama, an inverse mapping approach is employed in our Adaptive Gaze Module. This approach guarantees a dense target image without holes. The procedure involves mapping each pixel of the target viewport to a 3D ray, applying rotational transformations based on the desired viewing direction, and finally mapping the ray back to the ERP image coordinates to sample the pixel values.

Let the target rectilinear image have a width of $W$, a height of $H$, and a horizontal Field of View denoted as $\text{FOV}$. The focal length $f$ of the virtual camera is first calculated as:
\begin{equation}
    f = \frac{W}{2 \tan(\text{FOV} / 2)}
\end{equation}

For each pixel $(x, y)$ in the target image, assuming the optical center is at $(c_x, c_y)$ (typically the center of the image), its local 3D ray coordinates $(X_c, Y_c, Z_c)$ are defined as:
\begin{align}
    X_c &= x - c_x \\
    Y_c &= y - c_y \\
    Z_c &= f
\end{align}

To simulate the camera orientation, these 3D rays are rotated according to the specified yaw ($\theta$) and pitch ($\varphi$) angles. Let the rotation matrices for pitch (around the X-axis) and yaw (around the Y-axis) be $R_x$ and $R_y$, respectively. The world coordinates $(X_w, Y_w, Z_w)$ of the ray are obtained by applying these rotations:
\begin{equation}
    \begin{bmatrix} X_w \\ Y_w \\ Z_w \end{bmatrix} = R_y R_x \begin{bmatrix} X_c \\ Y_c \\ Z_c \end{bmatrix}
\end{equation}

The rotated 3D direction vectors in the world coordinate system are then converted into spherical coordinates, representing longitude $\lambda$ and latitude $\phi$:
\begin{align}
    \lambda &= \arctan2(X_w, Z_w) \\
    \phi &= \arcsin\left(\frac{Y_w}{\sqrt{X_w^2 + Y_w^2 + Z_w^2}}\right)
\end{align}
where $\lambda \in [-\pi, \pi]$ and $\phi \in [-\pi/2, \pi/2]$. 

Finally, the spherical coordinates are normalized and mapped to the 2D pixel coordinates $(U, V)$ of the original ERP image, which has a width of $W_{\text{erp}}$ and a height of $H_{\text{erp}}$:
\begin{align}
    U &= \left(\frac{\lambda}{2\pi} + 0.5\right) \times W_{\text{erp}} \\
    V &= \left(\frac{\phi}{\pi} + 0.5\right) \times H_{\text{erp}}
\end{align}

Once the continuous coordinates $(U, V)$ are computed for all pixels in the target viewport, bilinear interpolation is applied to sample the corresponding color values from the discrete grid of the ERP image, thereby constructing the final rectilinear output.

\subsection{Details of the Implementation of Baseline Methods}
As we are the first to tackle the panoramic affordance prediction task, there are no existing baseline methods can directly process the \dataset. All of the baseline methods are designed for single perspective image. Therefore, we kindly introduce specific adaptations for them to ensure a fair and valid comparison. 
First, methods such as Affordance-R1, AffordanceVLM, VisionReasoner, LISA, and OV-Seg are trained exclusively on standard perspective image affordance datasets. In our preliminary experiments, directly feeding them full-resolution panoramic images resulted in near-complete failure, yielding performance metrics close to zero. To ensure meaningful evaluations, we resize the panoramic input images to match the respective training resolutions of each method before testing. On the other hand, A4-Agent is a training-free framework whose underlying Vision-Language Model (VLM) inherently possesses some capability to process ultra-high-resolution images. Thus, while it does not fail completely when given full-resolution inputs, its performance remains suboptimal and incurs prohibitive computational costs. Consequently, we resize the input images for A4-Agent to $2000 \times 1000$. This resolution precisely matches the input resolution used in the initial stage of our proposed pipeline, further ensuring the validity and fairness of the evaluation. All the experiments and the calculation of metrics in this paper are conducted on the same Server with A40 GPU. For the baseline methods, we use the official code and hyperparameters provided by the authors, except for the resize operation mentioned above.

\section{More Details about the Data Annotation Process}

Here, we provide further details regarding the data annotation process for Affordance VQA.
Furthermore, to maximize our contribution to the research community, we will \textbf{\textit{fully open-source}} not only the \dataset~and \method~themselves, but also our complete data annotation pipeline. This encompasses the Agent employed in Phase 1 and the customized WebUI annotation tool developed for Phase 2.
\paragraph{\textbf{Phase 1: Affordance Question Formulation.}}
First, regarding question generation, to facilitate scalable production, we adopted a pipeline consisting of batch generation by an AI agent, followed by manual human refinement and filtering. For this agent, we utilized the system prompt illustrated in Fig.~\ref{prompt:qa_generation}. Our design process was guided by the following core principles:

\begin{enumerate}
    \item Due to the distortion present at the poles of ERP images, we implemented a specific processing step: we simultaneously feed the original image along with the converted cubemap projections---totaling seven images---into the VLM to generate the questions.
    \item Instead of visual identification (``Find the cup''), the AI must deduce the functional purpose of an object and frame the question as a human need (e.g., ``I need something to hold water''). Moreover, the target object must serve as the absolute unique solution to that specific need within the room.
    \item The prompt mandates strict boundaries to prevent ``technically correct but ambiguous'' VQA pairs. It forces the Agent to actively distinguish between subtle functional roles, such as the difference between a surface and a tool (e.g., a whiteboard vs.\ a marker) or a container and its contents (e.g., a cup vs.\ water).
    \item To ensure question diversity, we instructed the Agent to formulate 2 to 4 distinct questions for the same target object based on its various functional uses.
\end{enumerate}

To further ensure the quality and diversity of the generated questions, we employed two of the most powerful closed-source models (Gemini-3-Pro and GPT-5) as our backbones to independently conduct batch generation. Ultimately, their respective outputs were merged, followed by screening and refinement by human experts.

\begin{renderedpromptbox}[System Prompt: Affordance Question Generation]
    \label{prompt:qa_generation}
    \subsubsection{Role}\label{role}

You are an expert AI agent specializing in Visual Question Answering
(VQA) dataset generation.

\subsubsection{Task}\label{task}

You will be provided with \textbf{7 images} for each scene: 1.
\textbf{Image 1:} A 360-degree equirectangular panoramic view of the
room. 2. \textbf{Images 2-7:} Six cubemap projection faces generated
from the panoramic view (in order: Front, Right, Back, Left, Top,
Bottom), these images can assist you to understand the spatial context
and object details.

Your task is to analyze these images collectively to identify distinct
objects and generate ``Affordance VQA'' pairs. You must prioritize
generating unambiguous questions where the target object is the only
logical answer.

\subsubsection{Definition of Affordance}\label{definition-of-affordance}

An ``affordance'' defines the possible actions an agent can perform with
an object. * \textbf{Goal:} Create a query regarding a need/intent where
the \textbf{unique solution} is the specific object.

\subsubsection{Step-by-Step
Instructions}\label{step-by-step-instructions}

\begin{enumerate}
\def\labelenumi{\arabic{enumi}.}
\tightlist
\item
  \textbf{Object Detection \& Adaptive Description:} Identify objects
  and valid interactable parts using all available views.

  \begin{itemize}
  \tightlist
  \item
    \textbf{Rule:} \textbf{Adaptive Detail.}

    \begin{itemize}
    \tightlist
    \item
      \textbf{Simple Scenes/Unique Objects:} If an object is unique and
      unambiguous in the scene (e.g., only one sofa), use a
      \textbf{concise name} (e.g., ``sofa'').
    \item
      \textbf{Complex Scenes/Ambiguous Objects:} If there are multiple
      similar objects or the scene is crowded, you \textbf{MUST} provide
      a \textbf{detailed description} including \textbf{Location} (e.g.,
      ``on the left desk''), \textbf{Appearance} (e.g., ``red
      ceramic''), or \textbf{State} (e.g., ``open/closed'') to
      distinguish it.
    \end{itemize}
  \item
    \emph{Goal:} Keep it short when possible, but specific when
    necessary.
  \end{itemize}
\item
  \textbf{Disambiguation Logic:} Ensure the question points \emph{only}
  to the target object.

  \begin{itemize}
  \tightlist
  \item
    \emph{Distinguish Surface vs.~Tool:} ``Write \textbf{on}''
    (Whiteboard) vs.~``Write \textbf{with}'' (Pen).
  \item
    \emph{Distinguish Container vs.~Content:} ``Pour \textbf{into}''
    (Cup) vs.~``Drink'' (Water).
  \item
    \emph{Distinguish Identical Objects:} Use the detailed description
    from Step 1 if multiple instances exist (e.g., ``the open laptop''
    vs ``the closed laptop'').
  \end{itemize}
\item
  \textbf{Question Phrasing (CRITICAL):}

  \begin{itemize}
  \tightlist
  \item
    \textbf{DO NOT} start every question with ``I need'' or ``I want''.
  \item
    You \textbf{MUST} rotate between the following \textbf{4 phrasing
    styles}:

    \begin{itemize}
    \tightlist
    \item
      \textbf{Style A (Search/Location):} ``Where can I find a
      {[}property{]} place to {[}action{]}?''
    \item
      \textbf{Style B (Problem/Context):} ``{[}Situation description{]},
      what can I use to {[}solve it{]}?''
    \item
      \textbf{Style C (Operational/How-to):} ``How can I {[}action{]}
      using an object in this room?''
    \item
      \textbf{Style D (Functional Selection):} ``Which object allows me
      to {[}specific capability{]}?''
    \end{itemize}
  \end{itemize}
\item
  \textbf{Formatting:} Output strictly in \textbf{JSON Lines} format.
\end{enumerate}

\subsubsection{Output Format Schema}\label{output-format-schema}

Each line must follow this exact JSON structure: \{``object'':
``\textless Adaptive Object Description (Concise or
Detailed)\textgreater{}'', ``affordance\_question'': ``''\}

\subsubsection{Examples (Demonstrating Variety \& Adaptive
Detail)}\label{examples-demonstrating-variety-adaptive-detail}

\textbf{Input Image Context:} A living room with a beige sofa, a
wall-mounted TV, drawn curtains, a wooden coffee table, and TWO remote
controls (one on the sofa, one on the table).

\textbf{Output:} \{``object'': ``sofa'', ``affordance\_question'': ``I
am feeling exhausted and looking for a soft place to lie down.''\}
\{``object'': ``television'', ``affordance\_question'': ``Which device
should I look at to watch the evening news?''\} \{``object'':
``curtains'', ``affordance\_question'': ``The sun is too bright in my
eyes; what can I use to block the light?''\} \{``object'':
``coffee\_table'', ``affordance\_question'': ``Where is a stable surface
specifically designed to hold my beverages?''\} \{``object'': ``remote
control on the sofa armrest'', ``affordance\_question'': ``How can I
change the channel without standing up?''\} \{``object'': ``carpet'',
``affordance\_question'': ``I want to sit on the floor but need
something warmer than the tiles.''\}

\subsubsection{Constraints}\label{constraints}

\begin{enumerate}
\def\labelenumi{\arabic{enumi}.}
\tightlist
\item
  \textbf{Variety is Key:} Ensure the questions sound natural and
  distinct from one another. Avoid repetitive sentence starters.
\item
  \textbf{Avoid Ambiguity:} The question should not arguably apply to
  other common objects.
\item
  \textbf{Adaptive Detail:} Use detailed descriptions ONLY when
  necessary to distinguish objects. Otherwise, keep the object name
  concise.
\item
  \textbf{Format:} One JSON object per line. No markdown blocks.
\item
  \textbf{Language:} English only.
\end{enumerate}
 
\end{renderedpromptbox}

\begin{figure}[!h]
    \centering
    \includegraphics[width=\linewidth]{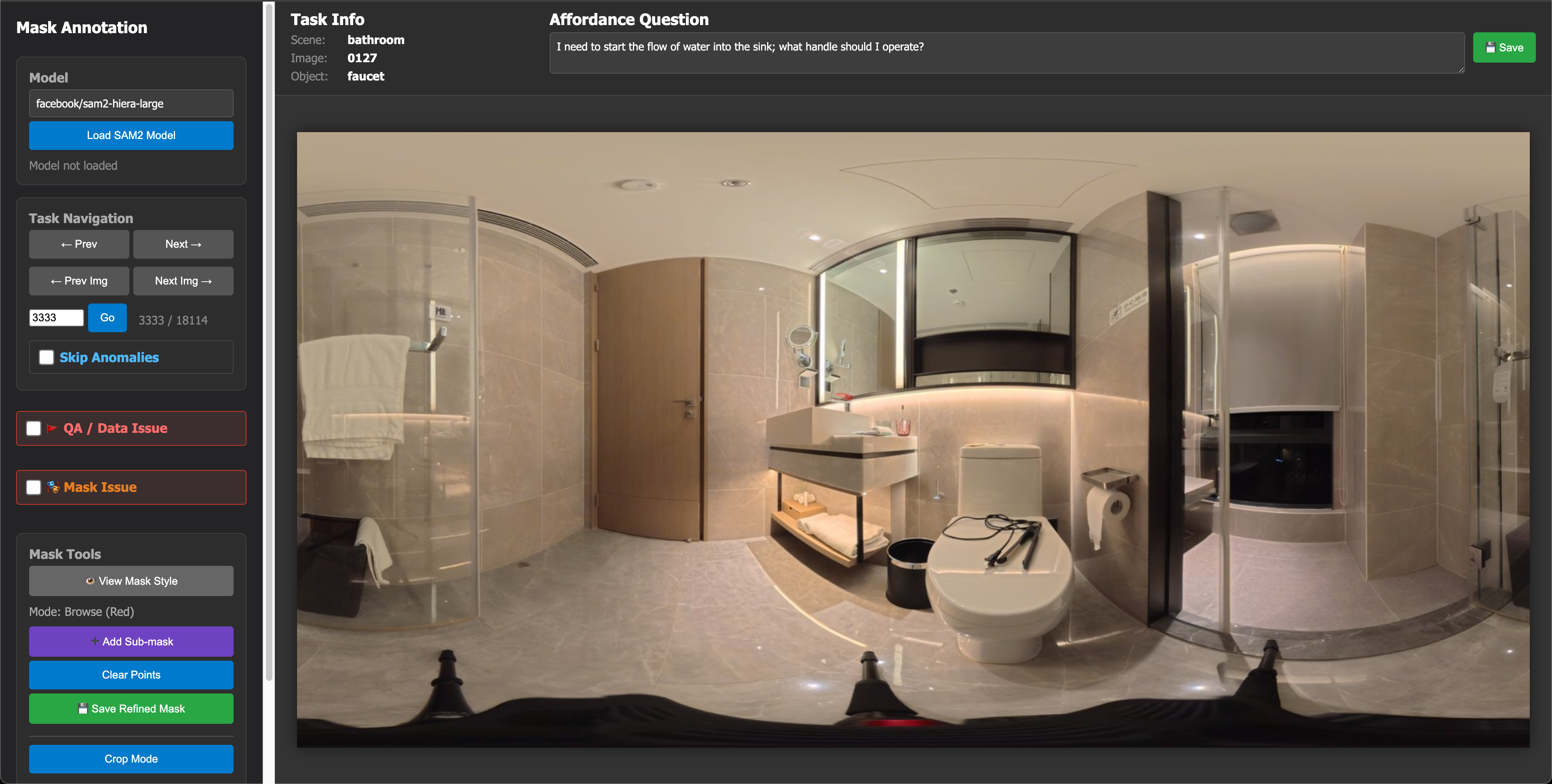}
    \caption{\centering The WebUI for Affordance VQA Annotation.}
    \label{fig:UI}
\end{figure}

\paragraph{\textbf{Phase 2: Mask Segmentation.}} 
Once the affordance question-answer pairs were established, the subsequent step involved generating the corresponding segmentation masks based on the answers. This process was manually executed by a dedicated annotation team of five members. To streamline the workflow, we developed a customized WebUI specifically for this dataset, as illustrated in Fig.~\ref{fig:UI}. The front-end of the interface displays relevant information to the annotator, such as the affordance question and answer, while the back-end integrates the SAM2 model, enabling users to provide prompts for accurate, interactive segmentation. The entire annotation effort spanned two months, ultimately yielding over 15,000 affordance VQA instances paired with precise segmentation masks.

\paragraph{\textbf{Phase 3: Final Verification.}} 
The final step involves a comprehensive review of the generated question-answer pairs and segmentation masks to ensure high quality and eliminate ambiguity. Specifically, the annotation team conducted a rigorous cross-checking process across the entire dataset. Following two complete rounds of manual review, we filtered out over 1,000 potentially ambiguous instances from the initial pool of 15,000+ samples. This rigorous verification process resulted in a final, high-quality dataset consisting of 13,943 affordance VQA instances paired with segmentation masks.

\section{More Visualizations of \dataset}
\begin{figure}[h]
    \centering
    \includegraphics[width=0.95\linewidth]{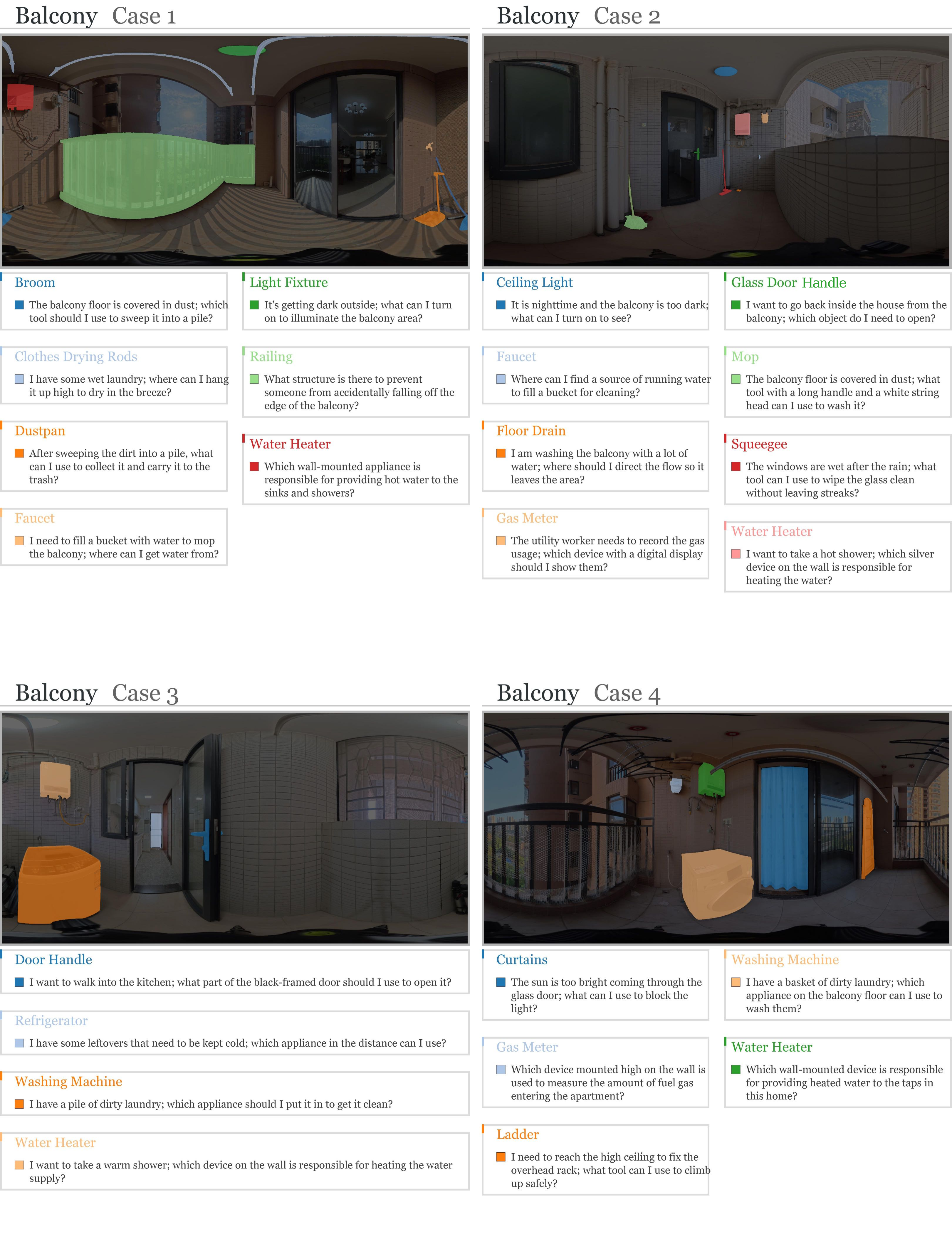}

    \caption{\centering Visualizations of \dataset (Balcony).}
    \label{fig:bal}
\end{figure}

\begin{figure}[t]
    \centering
    \includegraphics[width=0.95\linewidth]{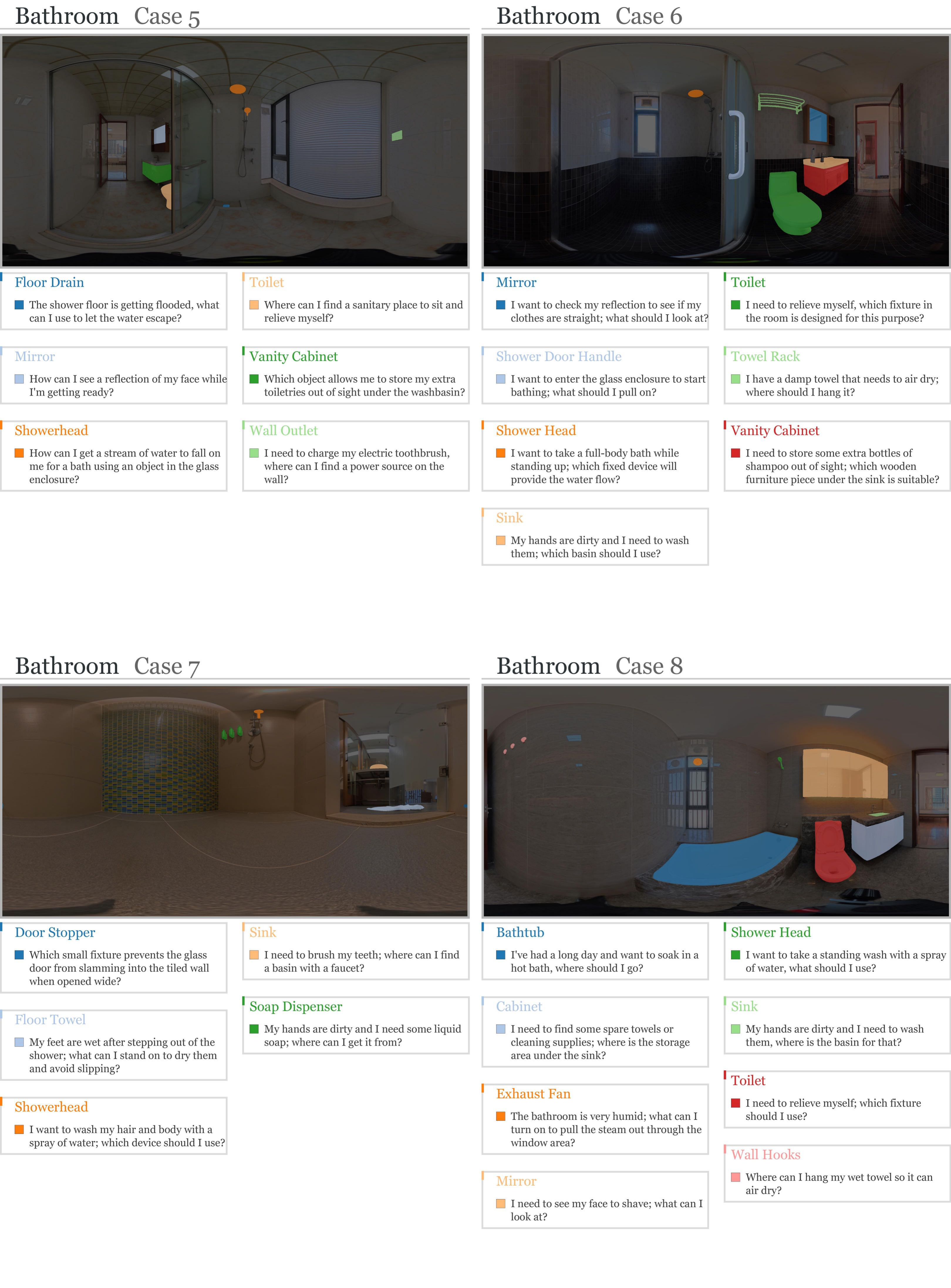}
    \caption{\centering Visualizations of \dataset (Bathroom).}
    \label{fig:bat}
\end{figure}

\begin{figure}[t]
    \centering
    \includegraphics[width=0.95\linewidth]{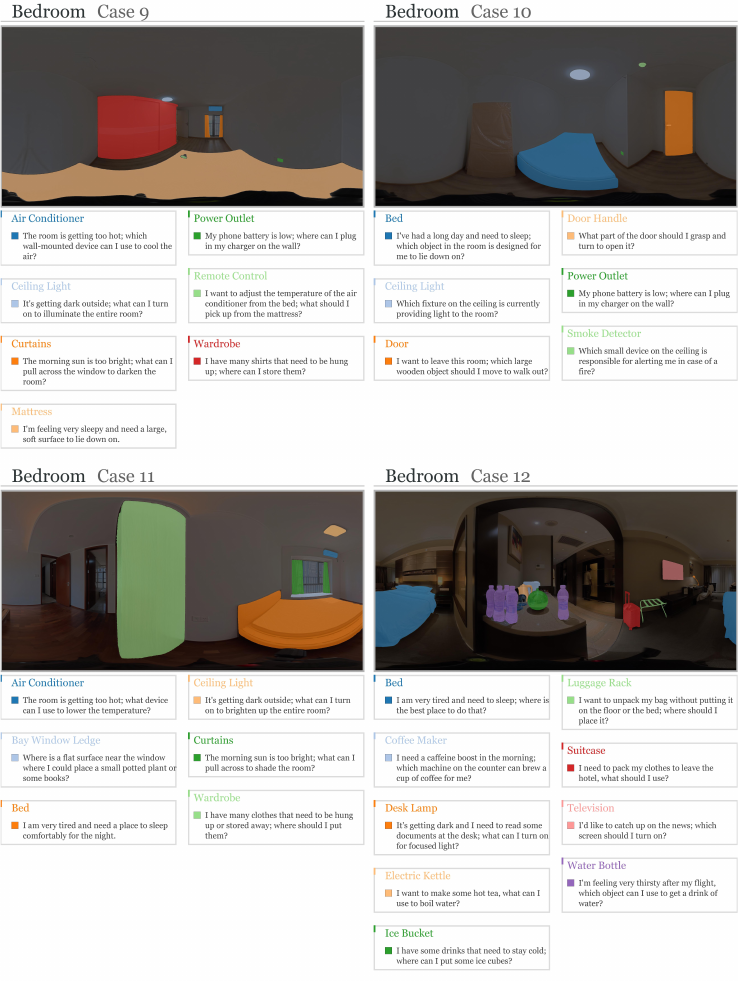}
    \caption{\centering Visualizations of \dataset (Bedroom).}
    \label{fig:bed}
\end{figure}

\begin{figure}[t]
    \centering
    \includegraphics[width=0.95\linewidth]{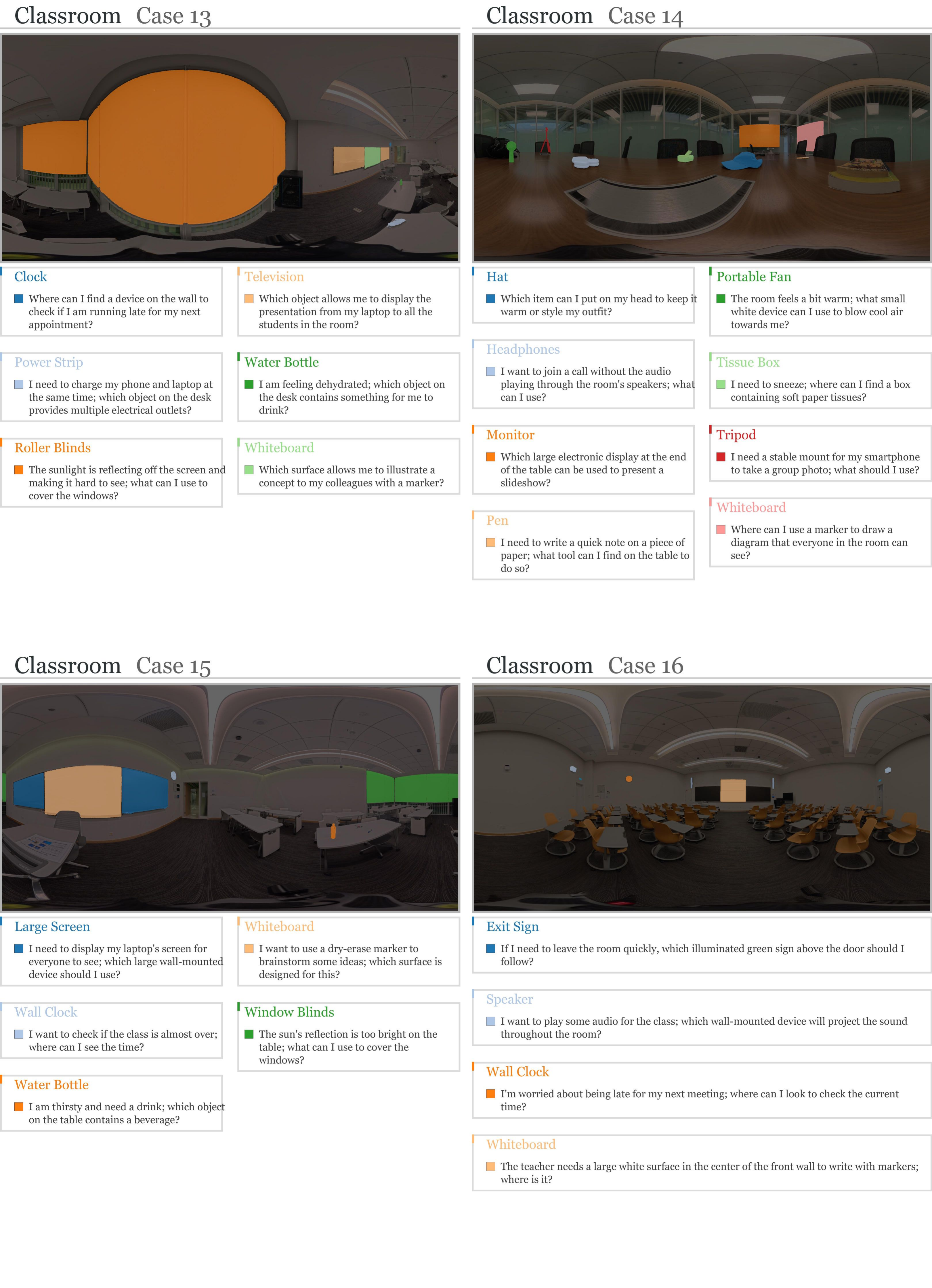}
    \caption{\centering Visualizations of \dataset (Classroom).}
    \label{fig:cla}
\end{figure}

\begin{figure}[t]
    \centering
    \includegraphics[width=0.95\linewidth]{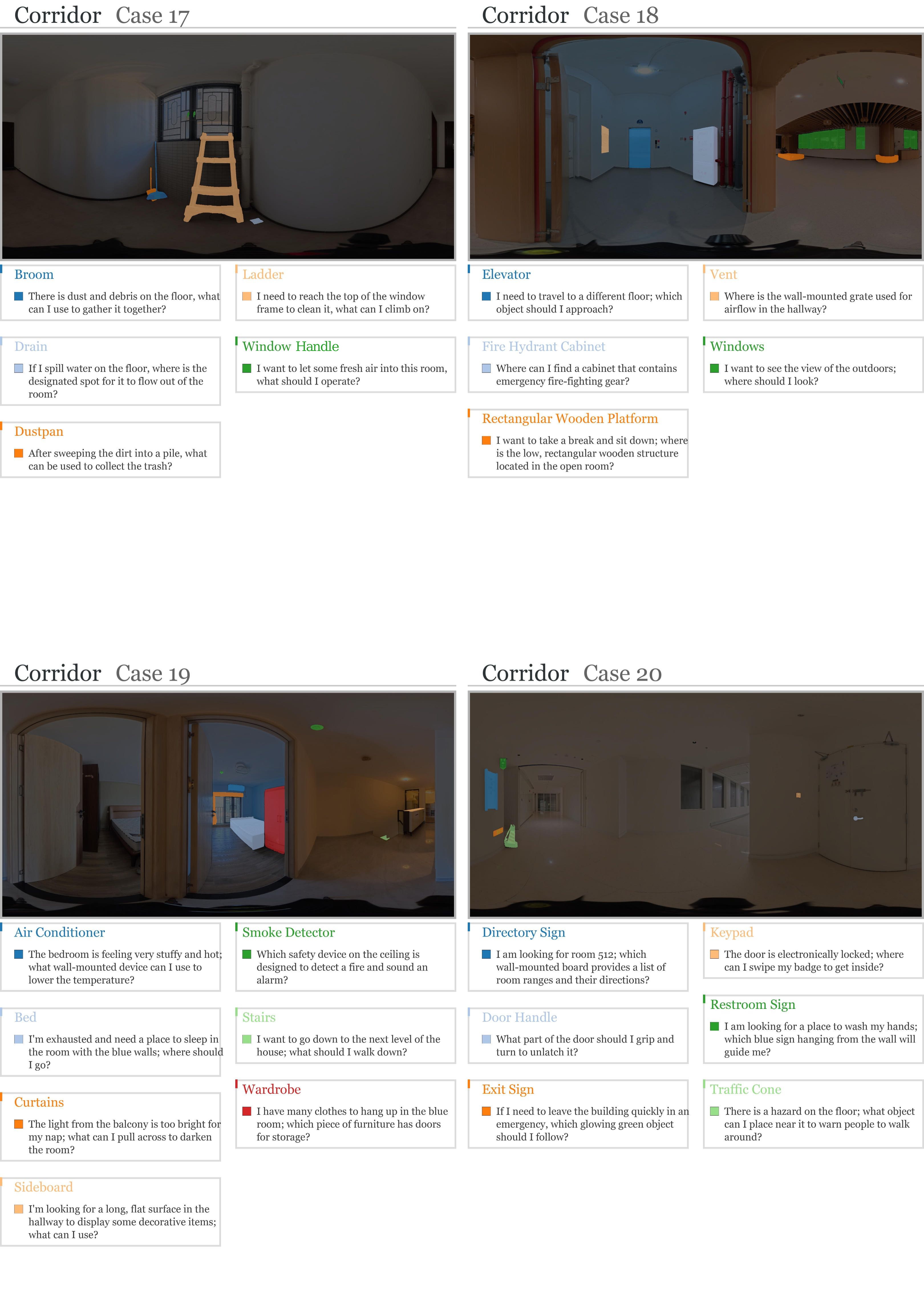}
    \caption{\centering Visualizations of \dataset (Corridor).}
    \label{fig:cor}
\end{figure}

\begin{figure}[t]
    \centering
    \includegraphics[width=0.95\linewidth]{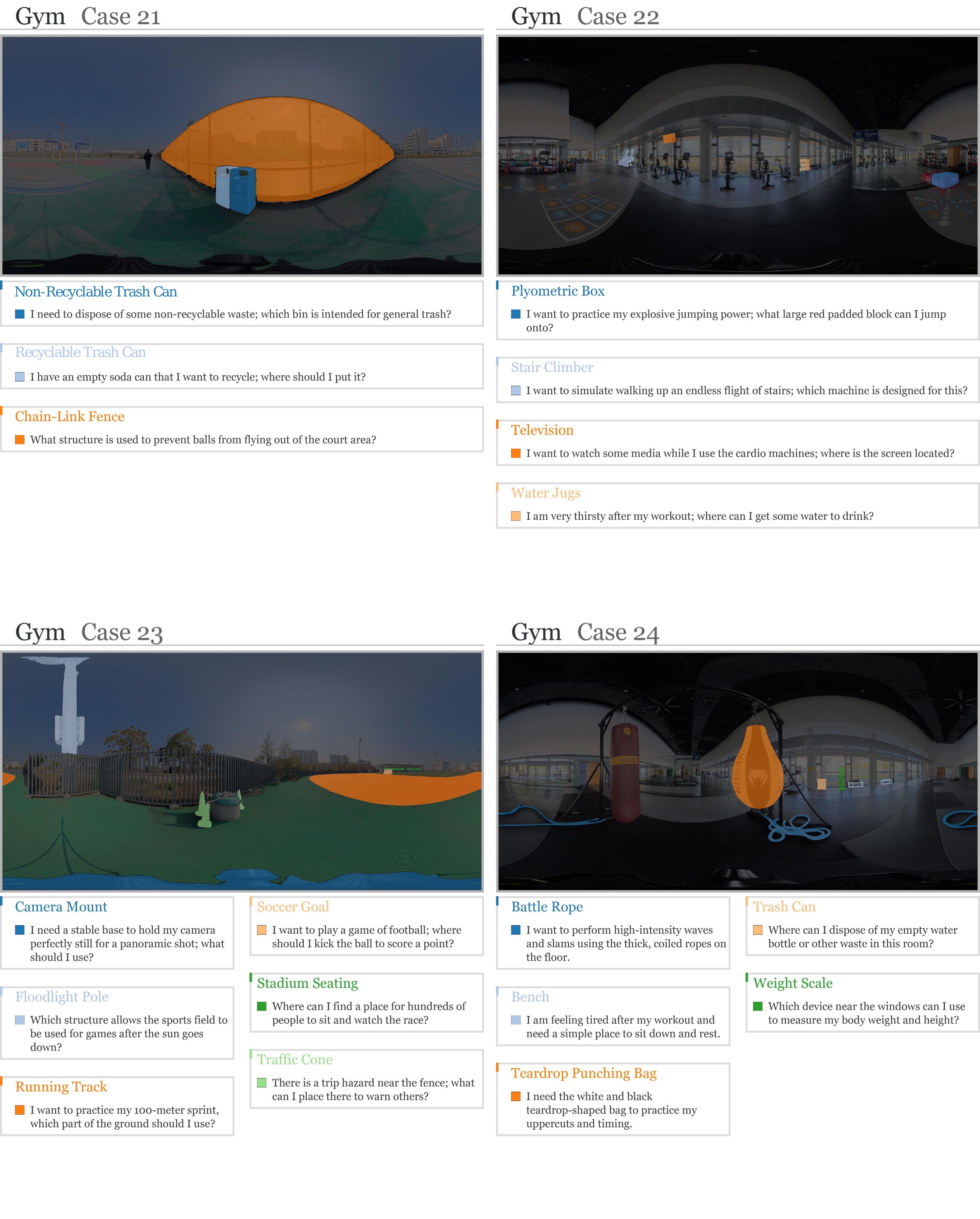}
    \caption{\centering Visualizations of \dataset (Gym).}
    \label{fig:gym}
\end{figure}

\begin{figure}[t]
    \centering
    \includegraphics[width=0.95\linewidth]{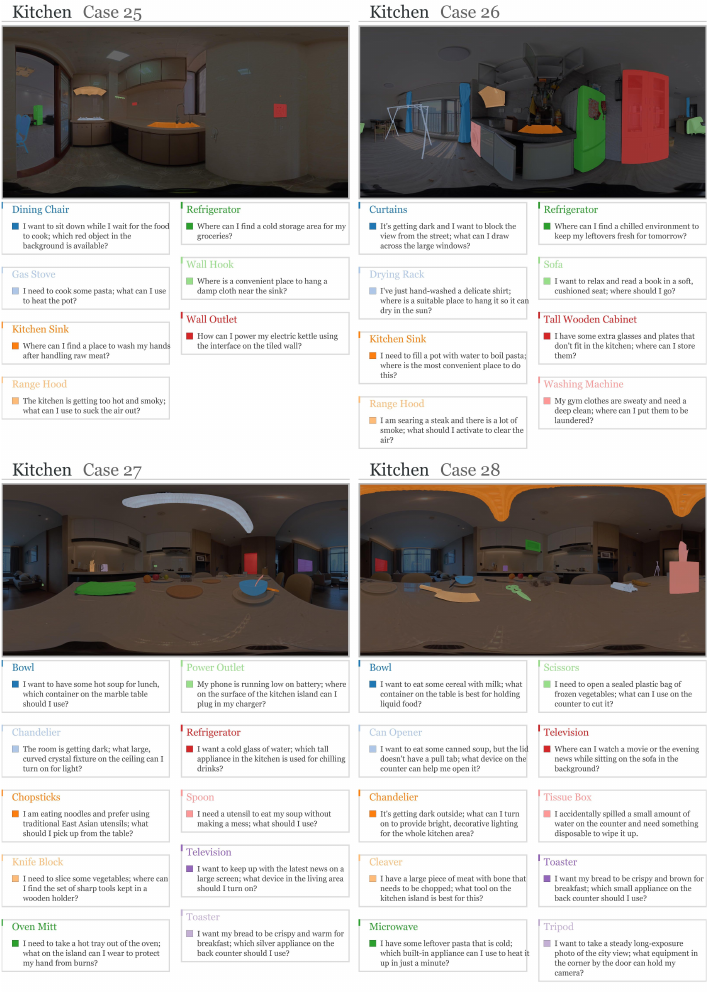}
    \caption{\centering Visualizations of \dataset (Kitchen).}
    \label{fig:kit}
\end{figure}

\begin{figure}[t]
    \centering
    \includegraphics[width=0.95\linewidth]{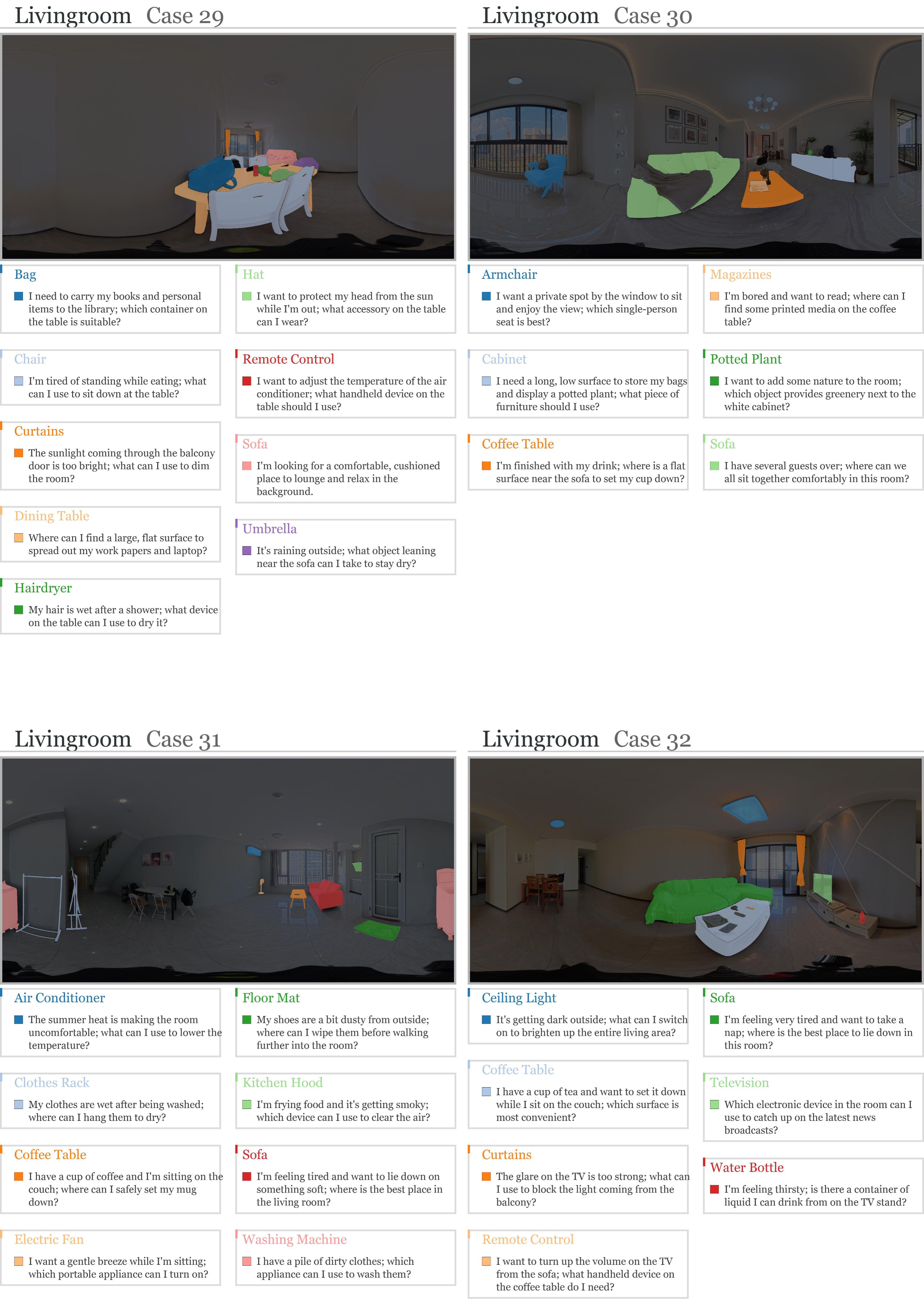}
    \caption{\centering Visualizations of \dataset (Livingroom).}
    \label{fig:liv}
\end{figure}

\begin{figure}[t]
    \centering
    \includegraphics[width=0.95\linewidth]{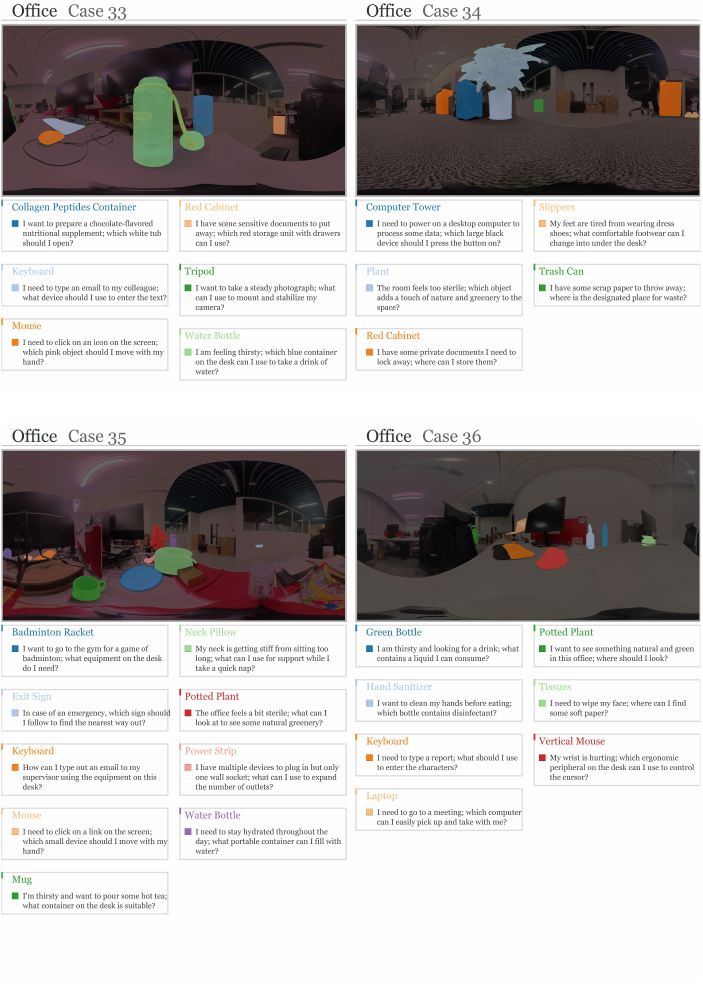}
    \caption{\centering Visualizations of \dataset (Office).}
    \label{fig:off}
\end{figure}

\begin{figure}[t]
    \centering
    \includegraphics[width=0.95\linewidth]{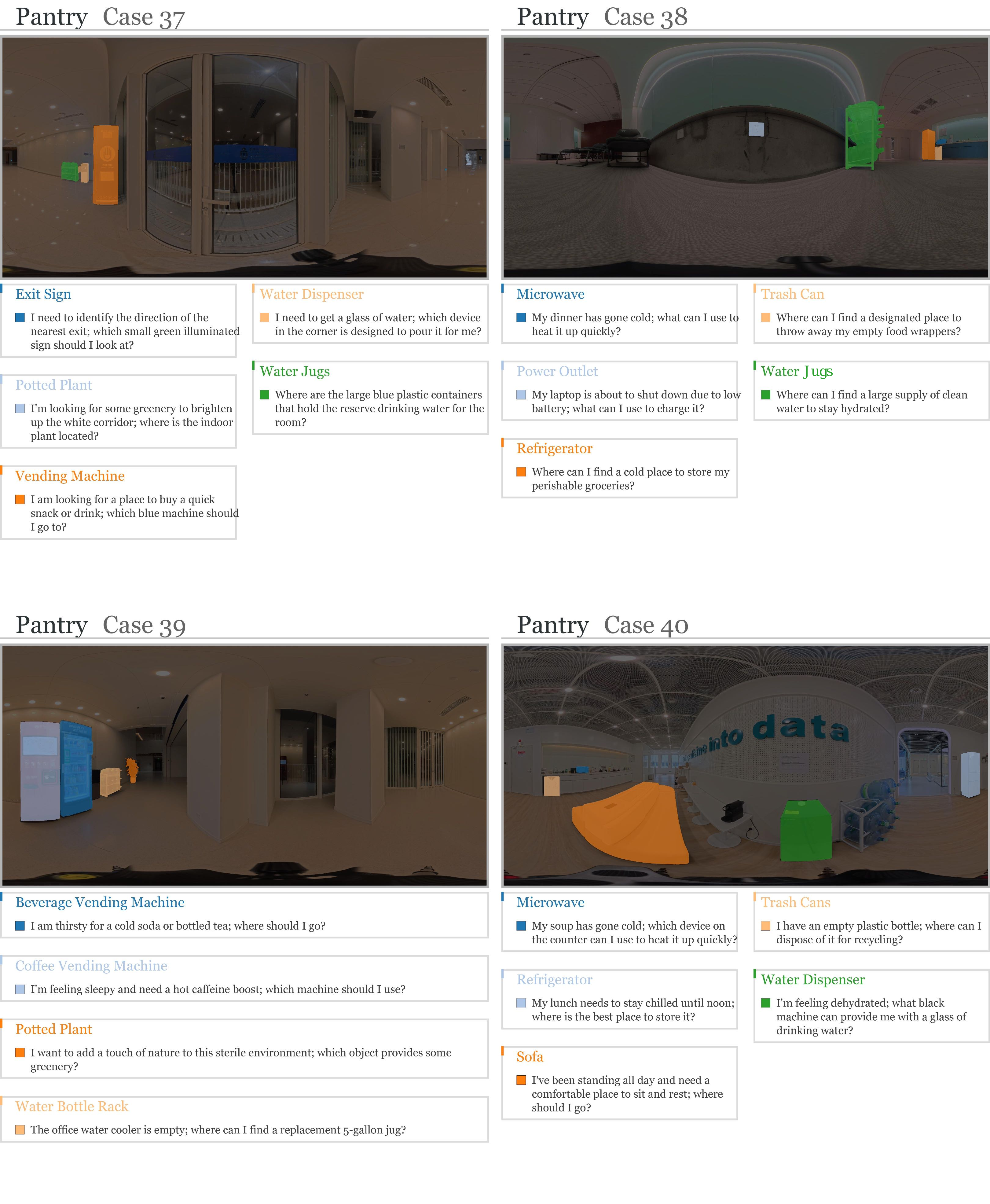}
    \caption{\centering Visualizations of \dataset (Pantry).}
    \label{fig:pan}
\end{figure}

\begin{figure}[t]
    \centering
    \includegraphics[width=0.95\linewidth]{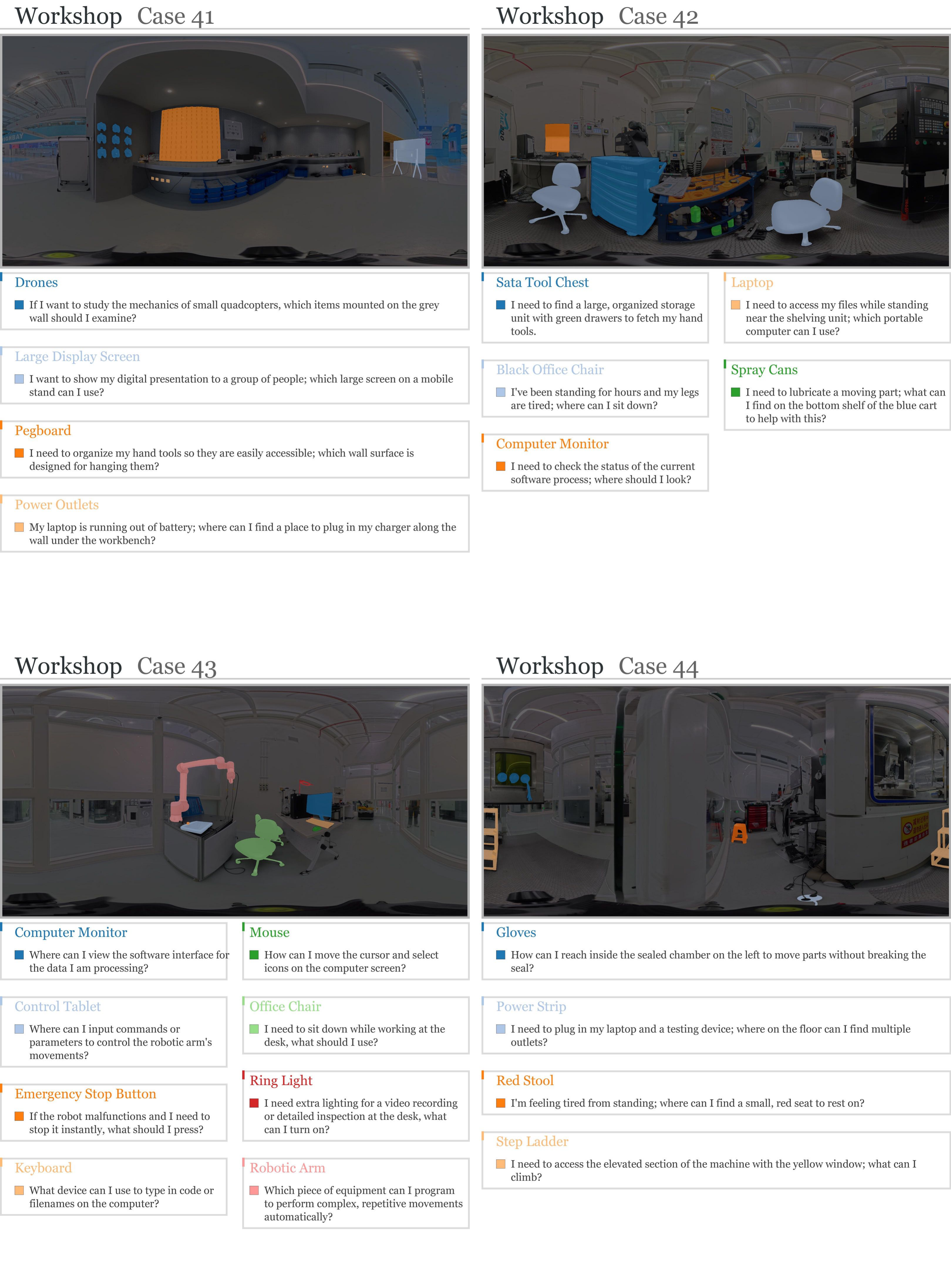}
    \caption{\centering Visualizations of \dataset (Workshop).}
    \label{fig:wor}
\end{figure}

\begin{figure}[t]
    \centering
    \includegraphics[width=0.95\linewidth]{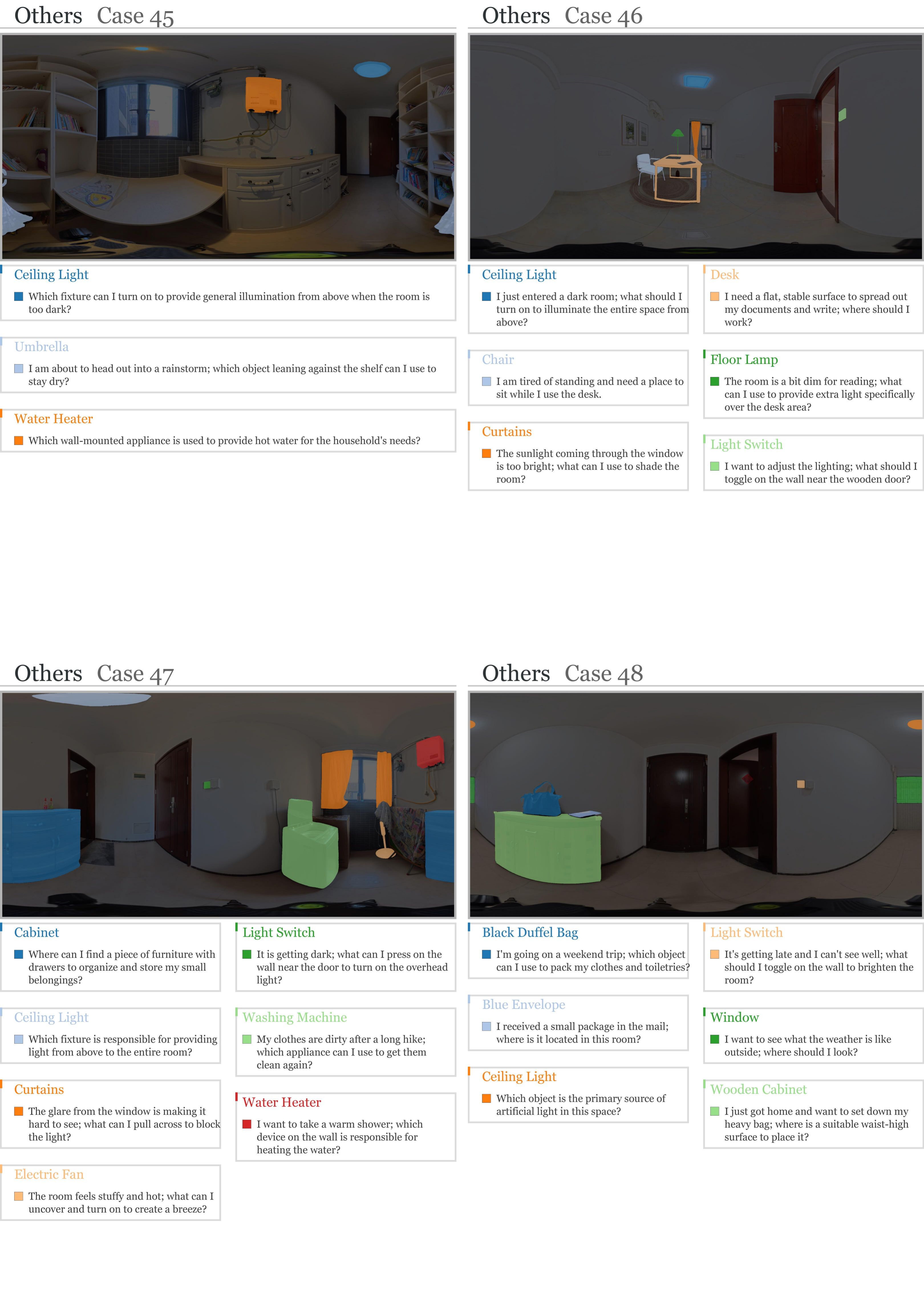}
    \caption{\centering Visualizations of \dataset (Others).}
    \label{fig:oth}
\end{figure}

We provide more visualizations of the \dataset~in this section. For each scene type, we randomly select 4 cases. Each case features a variable number of objects with their corresponding ground-truth affordance maps and questions. All of them are visualized in distinct colors within the same panoramic image. Please refer to Fig.~\ref{fig:bal}~(Balcony), Fig.~\ref{fig:bat}~(Bathroom), Fig.~\ref{fig:bed}~(Bedroom), Fig.~\ref{fig:cla}~(Classroom), Fig.~\ref{fig:cor}~(Corridor), Fig.~\ref{fig:gym}~(Gym), Fig.~\ref{fig:kit}~(Kitchen), Fig.~\ref{fig:liv}~(Livingroom), Fig.~\ref{fig:off}~(Office), Fig.~\ref{fig:pan}~(Pantry), Fig.~\ref{fig:wor}~(Workshop), and Fig.~\ref{fig:oth}~(Others) for detailed visualizations. We have also provided a dataset preview in our \href{https://zixinzhang02.github.io/Panoramic-Affordance-Prediction/}{Webpage}.

\section{More Qualitative Comparisons on \dataset}
\begin{figure}
    \centering
    \includegraphics[width=0.88\linewidth]{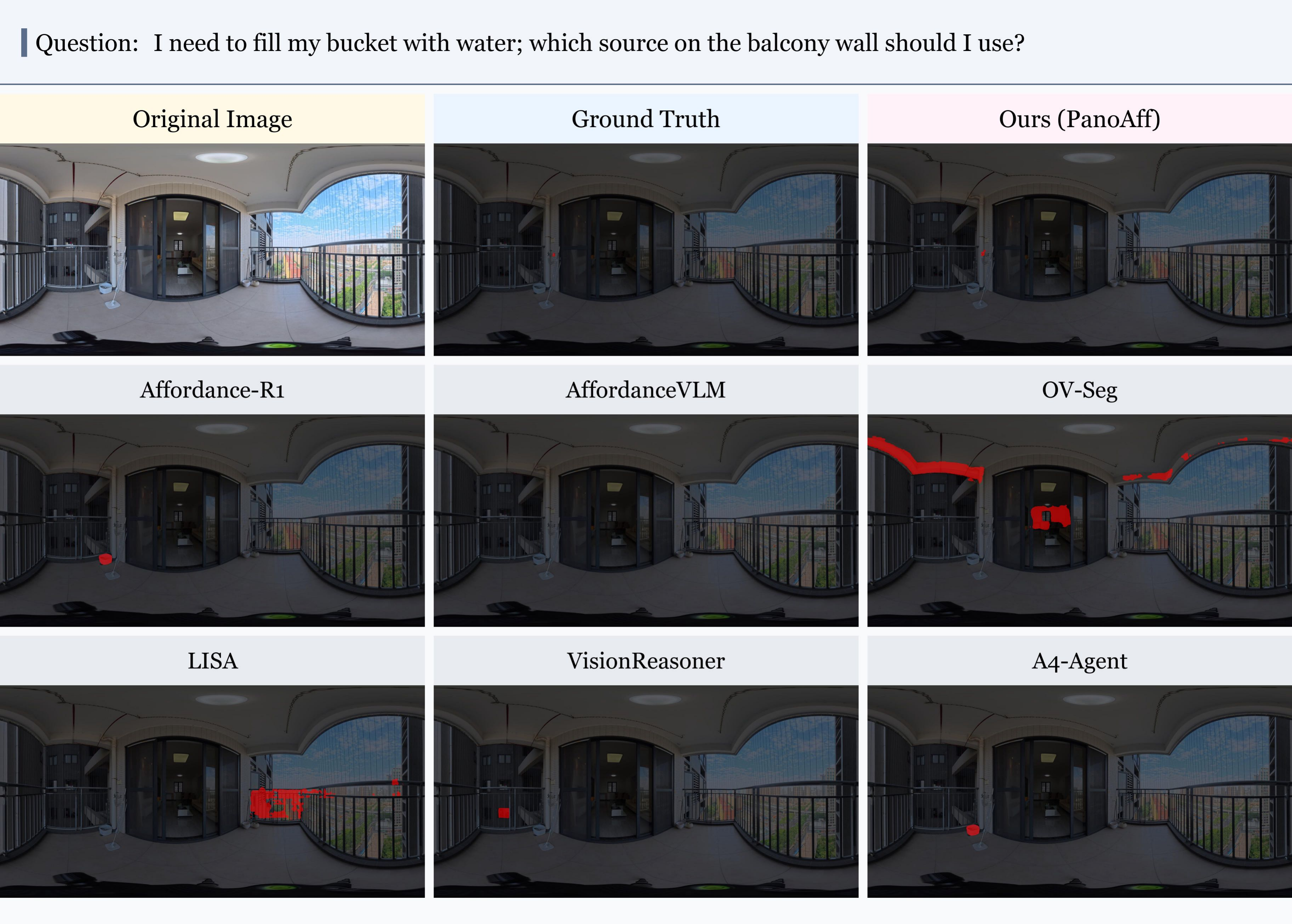}
    \caption{\centering Qualitative comparison on PAP-12K~(Balcony).}
    \label{fig:compare_balcony}
\end{figure}

\begin{figure}
    \centering
    \includegraphics[width=0.88\linewidth]{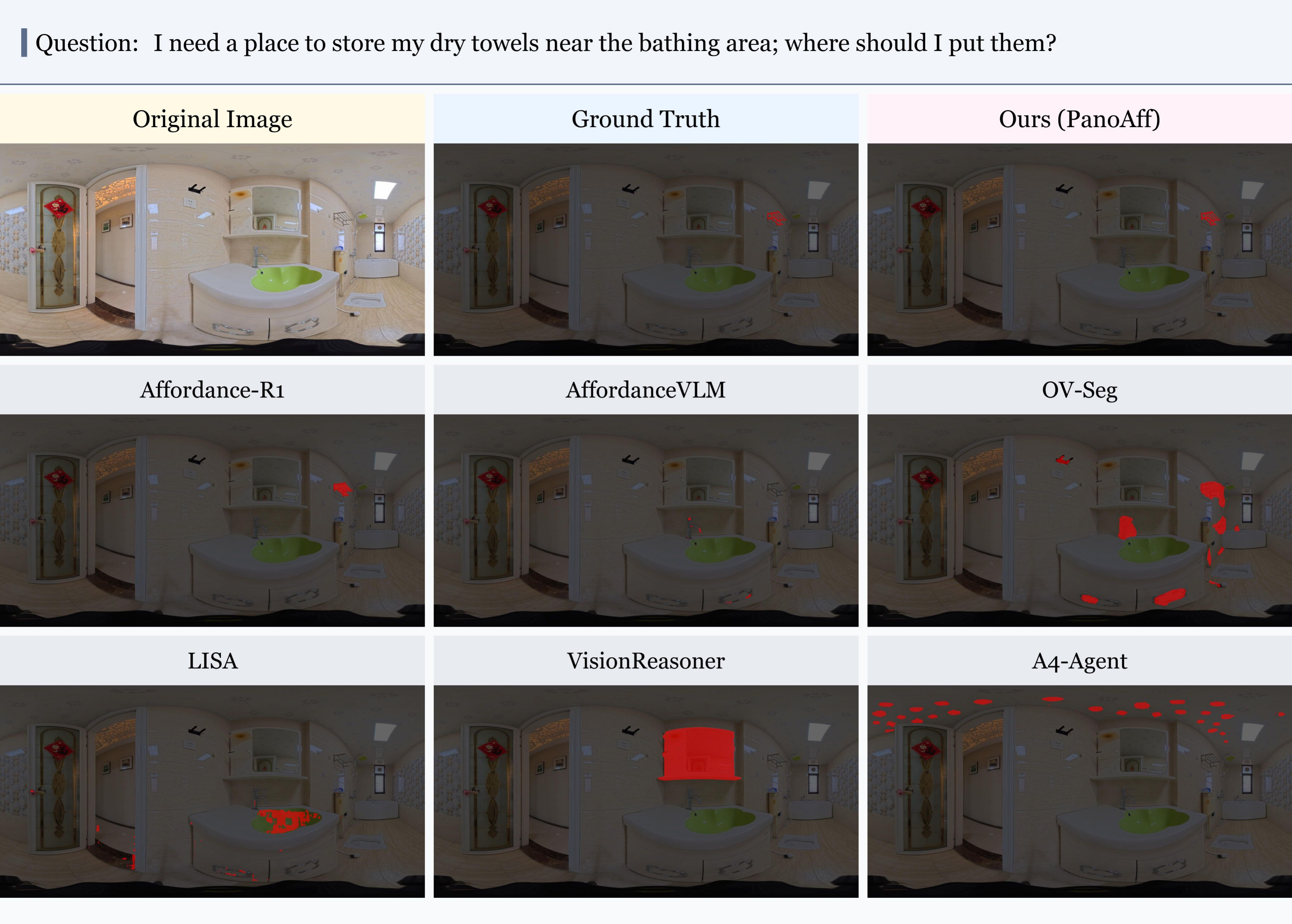}
    \caption{\centering Qualitative comparison on PAP-12K~(Bathroom).}
    \label{fig:compare_bathroom}
\end{figure}

\begin{figure}
    \centering
    \includegraphics[width=0.88\linewidth]{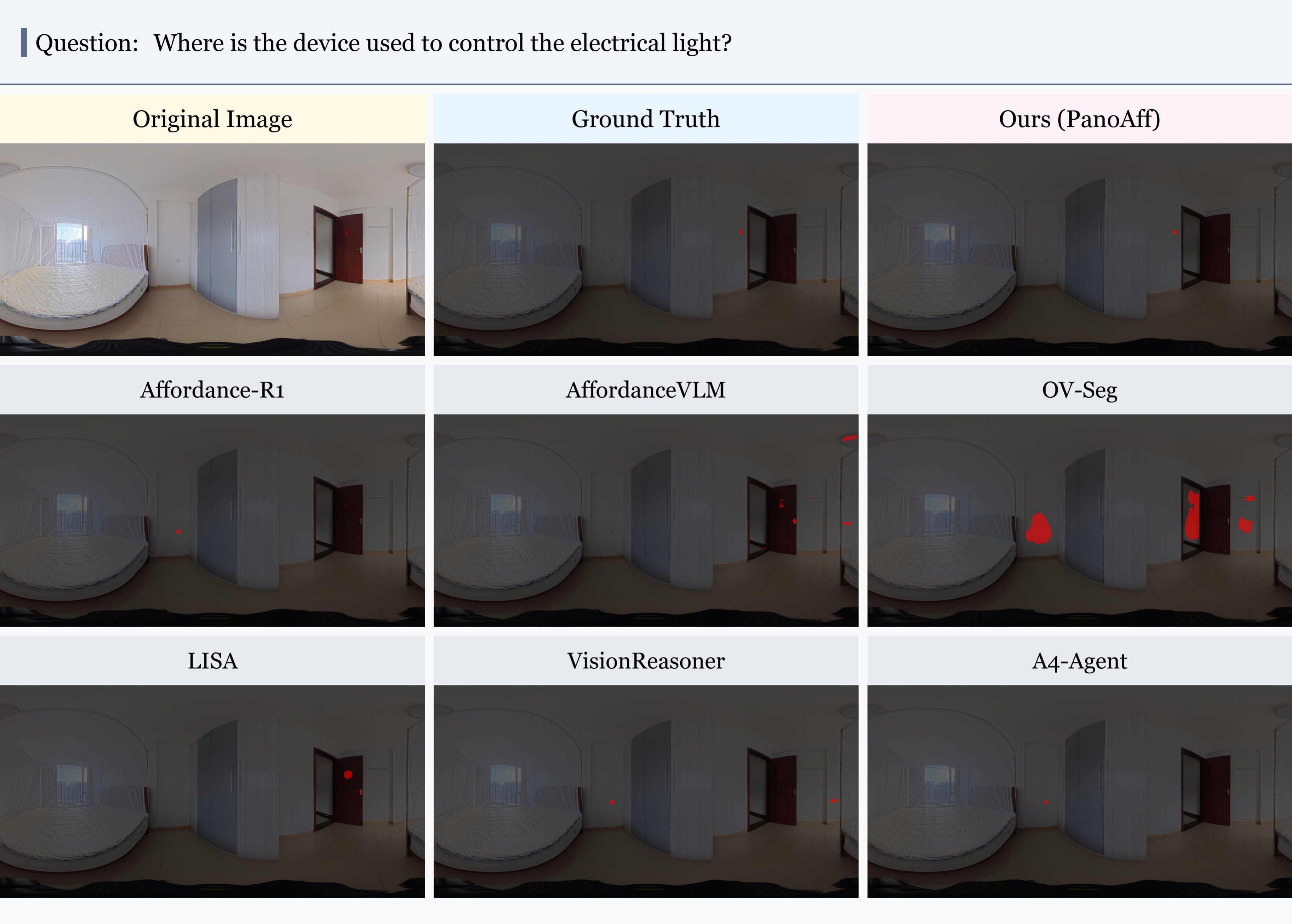}
    \caption{\centering Qualitative comparison on PAP-12K~(Bedroom).}
    \label{fig:compare_bedroom}
\end{figure}

\begin{figure}
    \centering
    \includegraphics[width=0.88\linewidth]{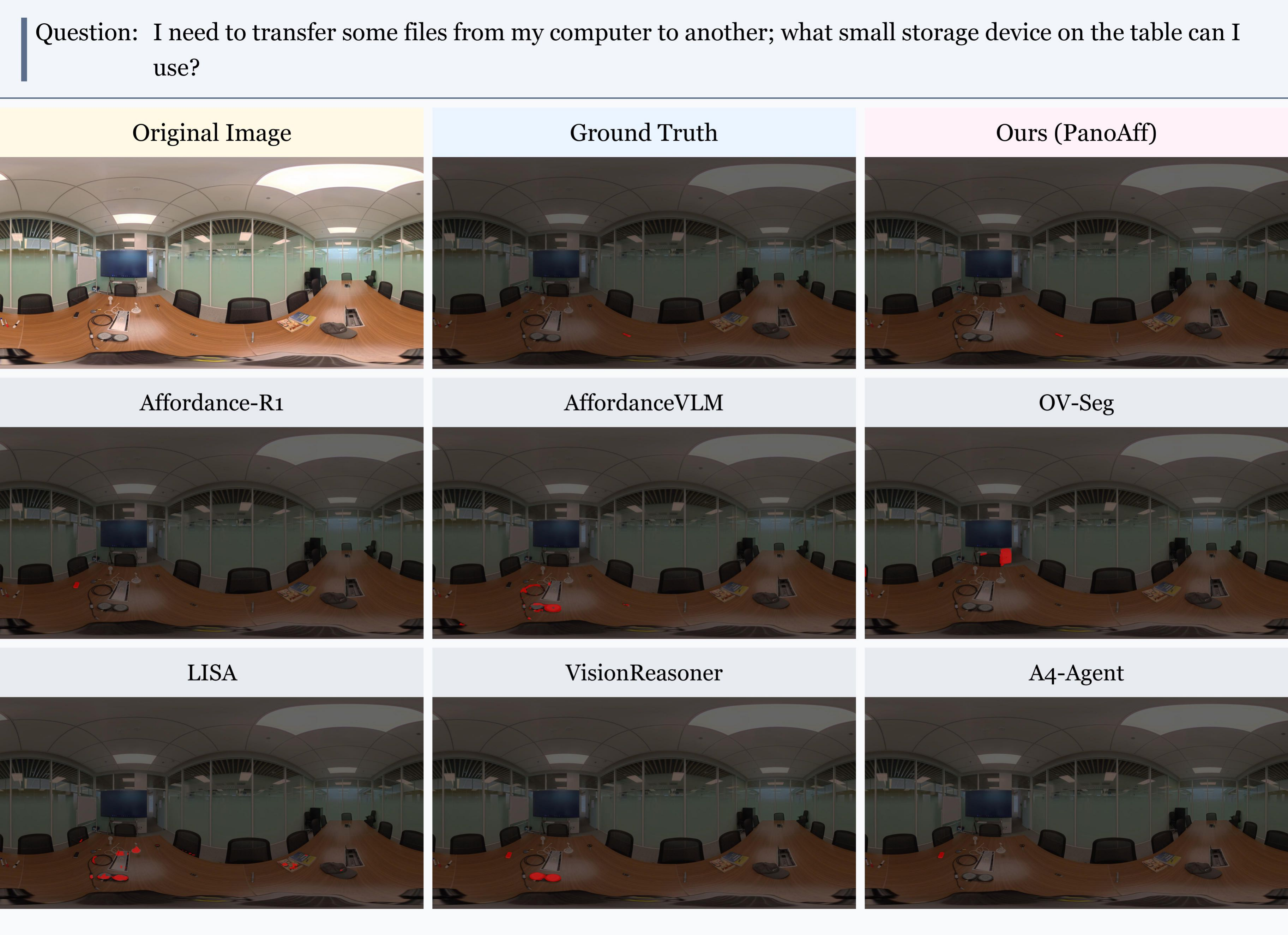}
    \caption{\centering Qualitative comparison on PAP-12K~(Classroom).}
    \label{fig:compare_classroom}
\end{figure}

\begin{figure}
    \centering
    \includegraphics[width=0.88\linewidth]{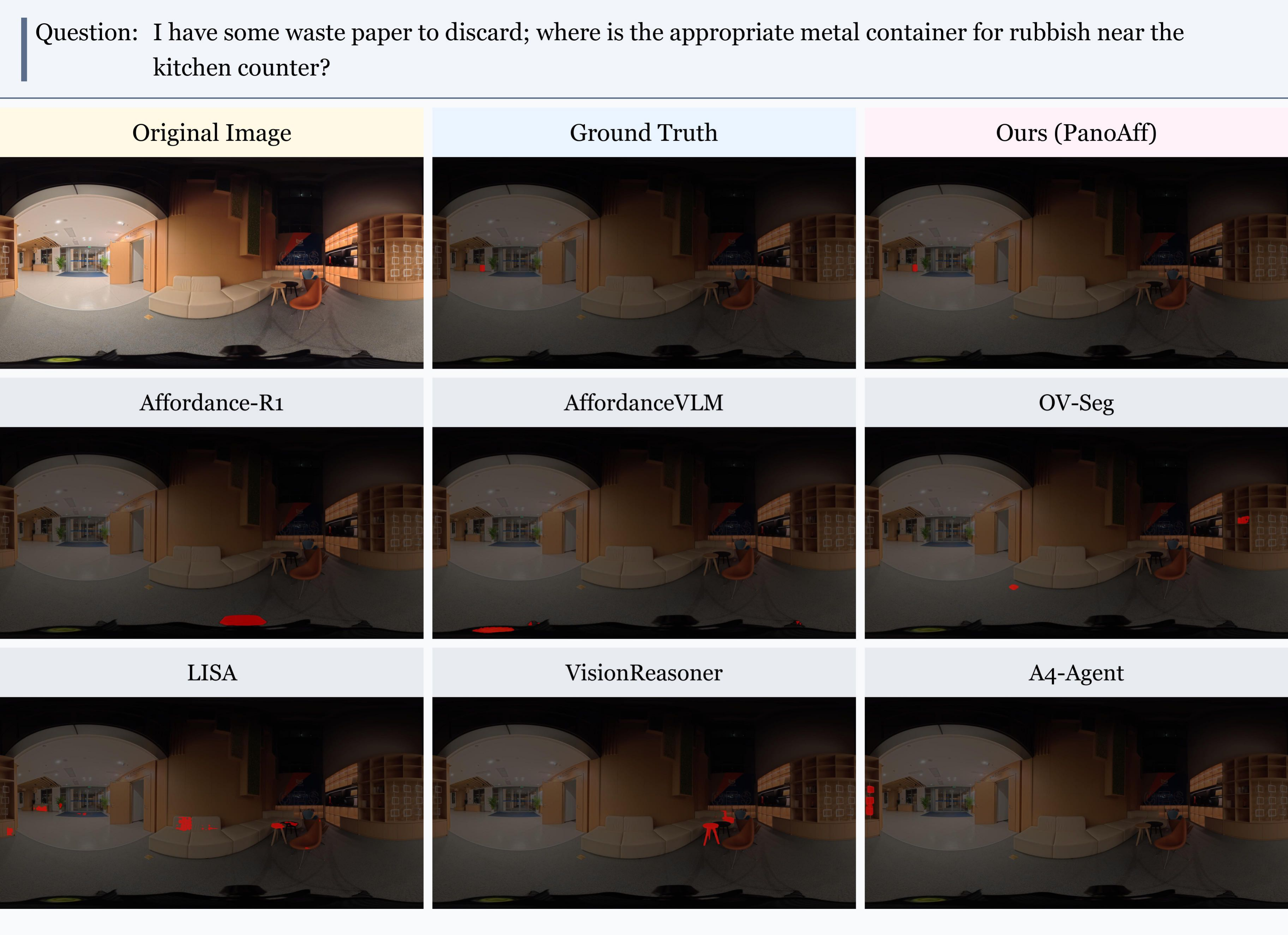}
    \caption{\centering Qualitative comparison on PAP-12K~(Corridor).}
    \label{fig:compare_corridor}
\end{figure}

\begin{figure}
    \centering
    \includegraphics[width=0.88\linewidth]{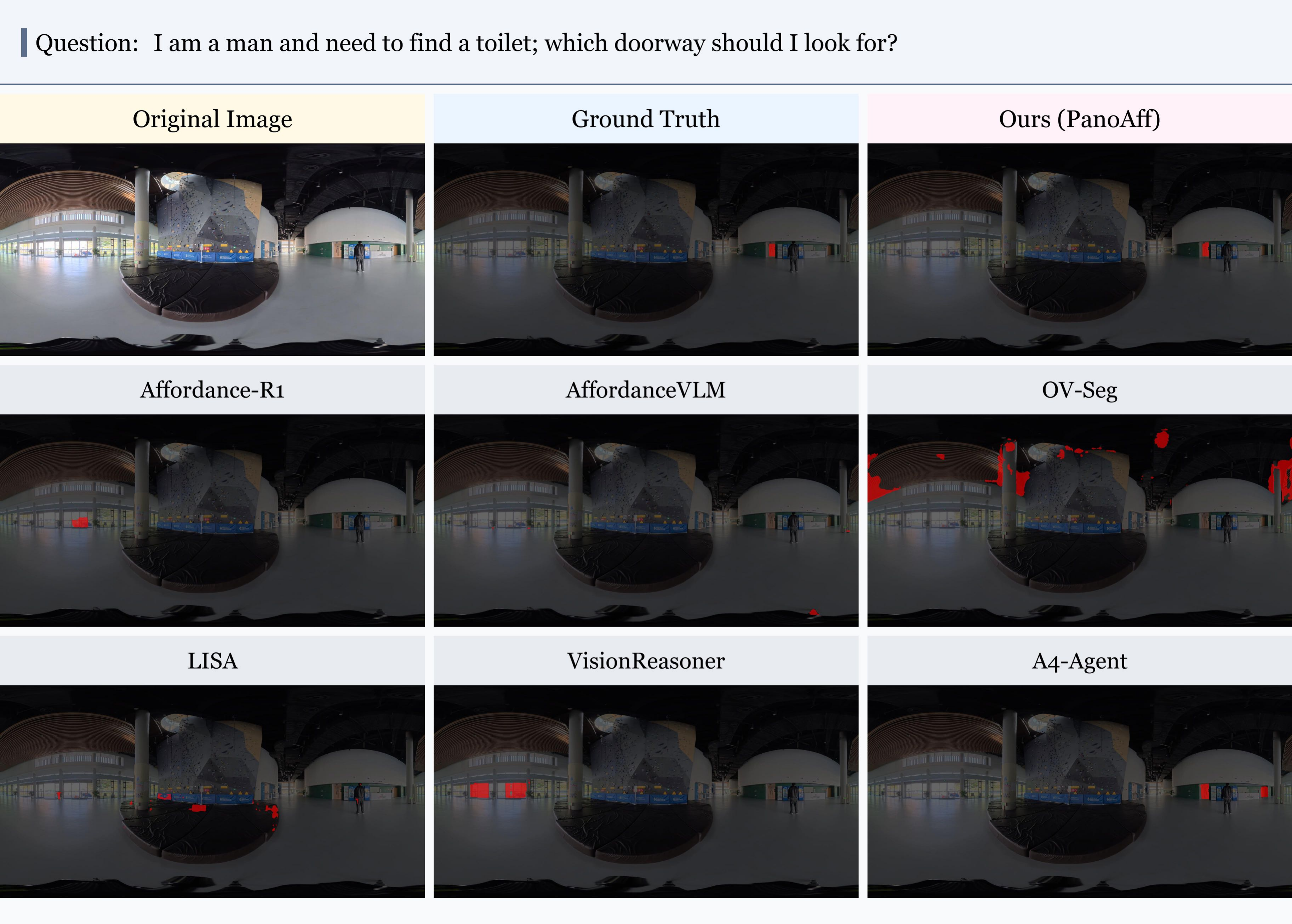}
    \caption{\centering Qualitative comparison on PAP-12K~(Gym).}
    \label{fig:compare_gym}
\end{figure}

\begin{figure}
    \centering
    \includegraphics[width=0.88\linewidth]{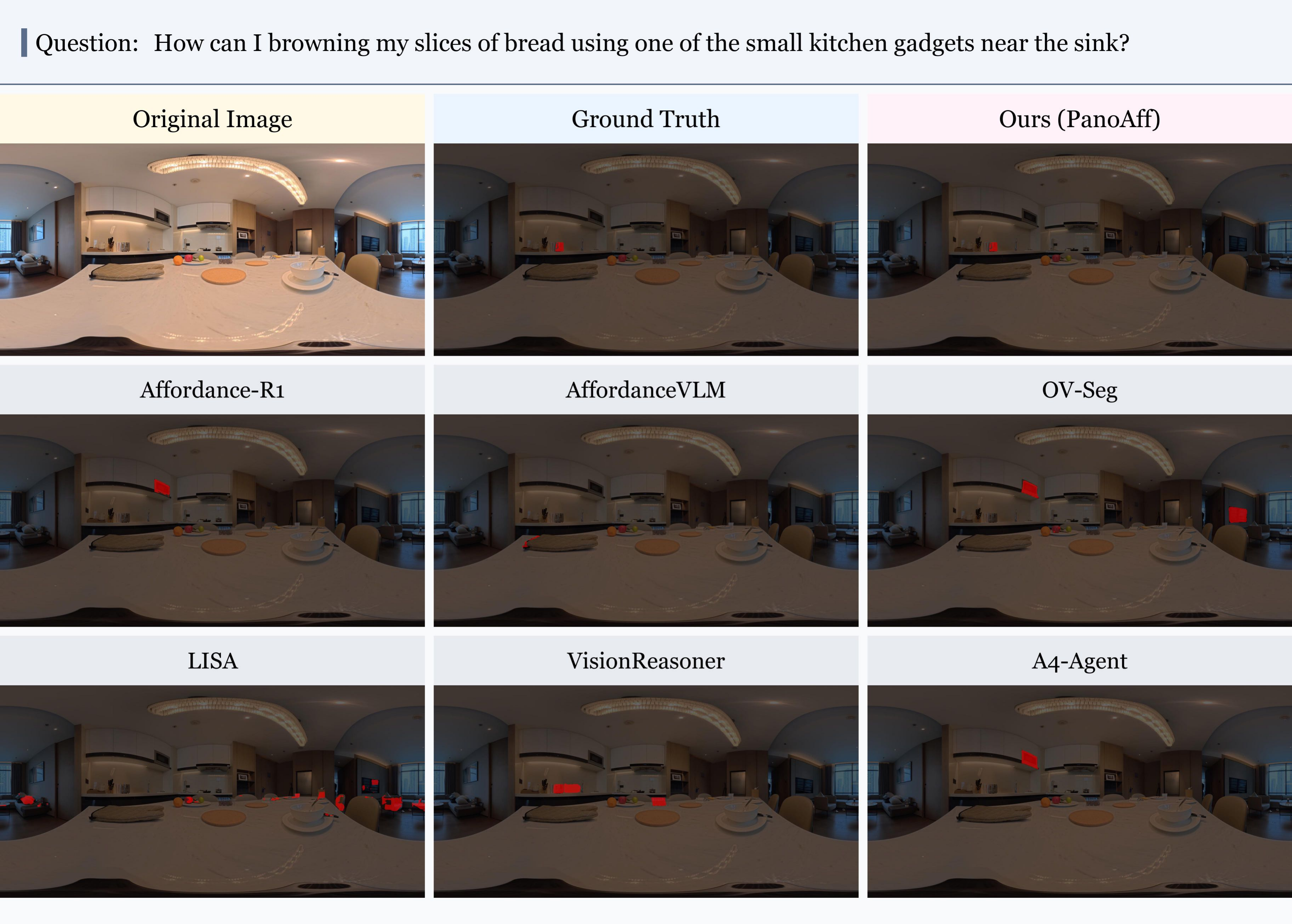}
    \caption{\centering Qualitative comparison on PAP-12K~(Kitchen).}
    \label{fig:compare_kitchen}
\end{figure}

\begin{figure}
    \centering
    \includegraphics[width=0.88\linewidth]{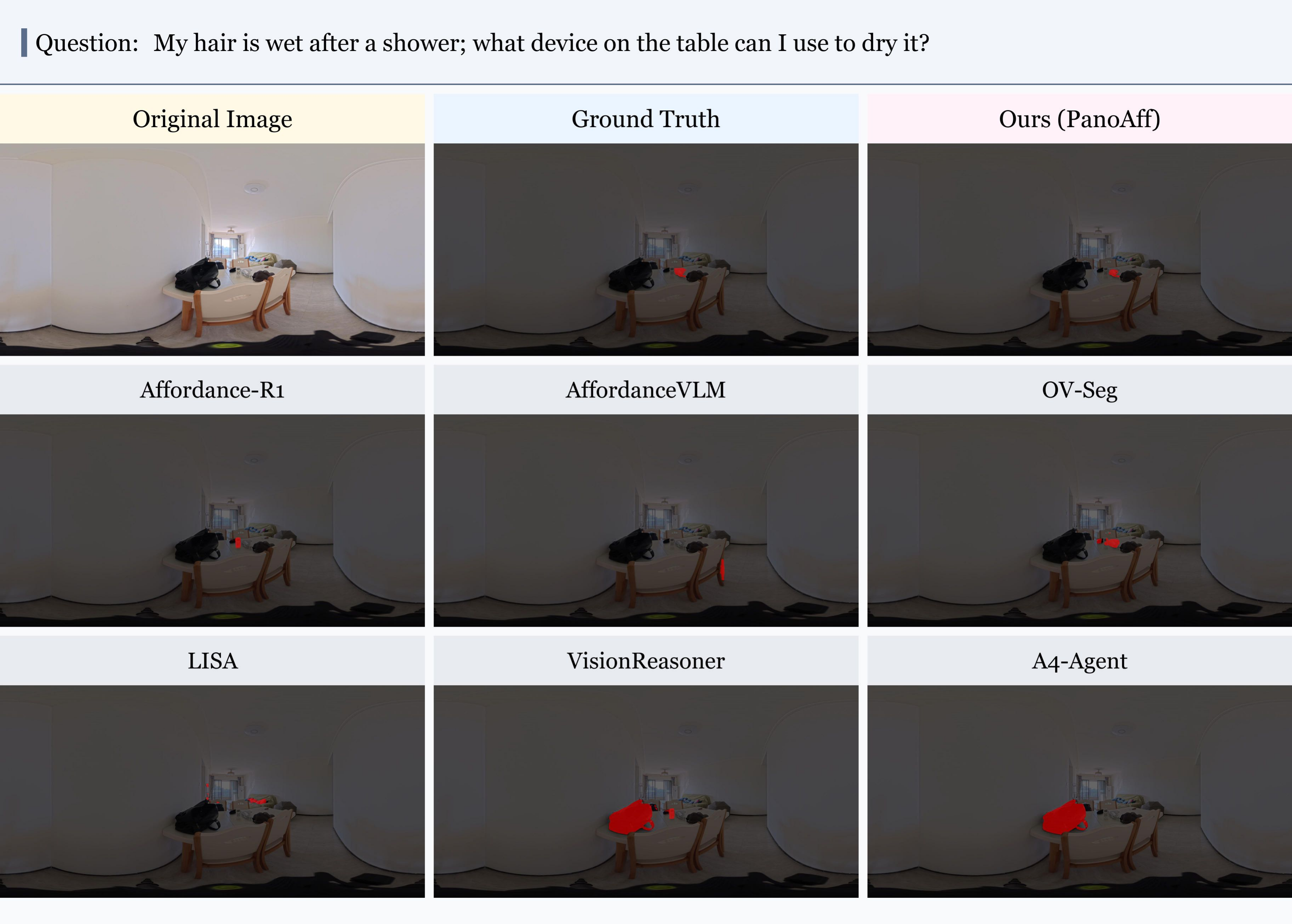}
    \caption{\centering Qualitative comparison on PAP-12K~(Livingroom).}
    \label{fig:compare_livingroom}
\end{figure}

\begin{figure}
    \centering
    \includegraphics[width=0.88\linewidth]{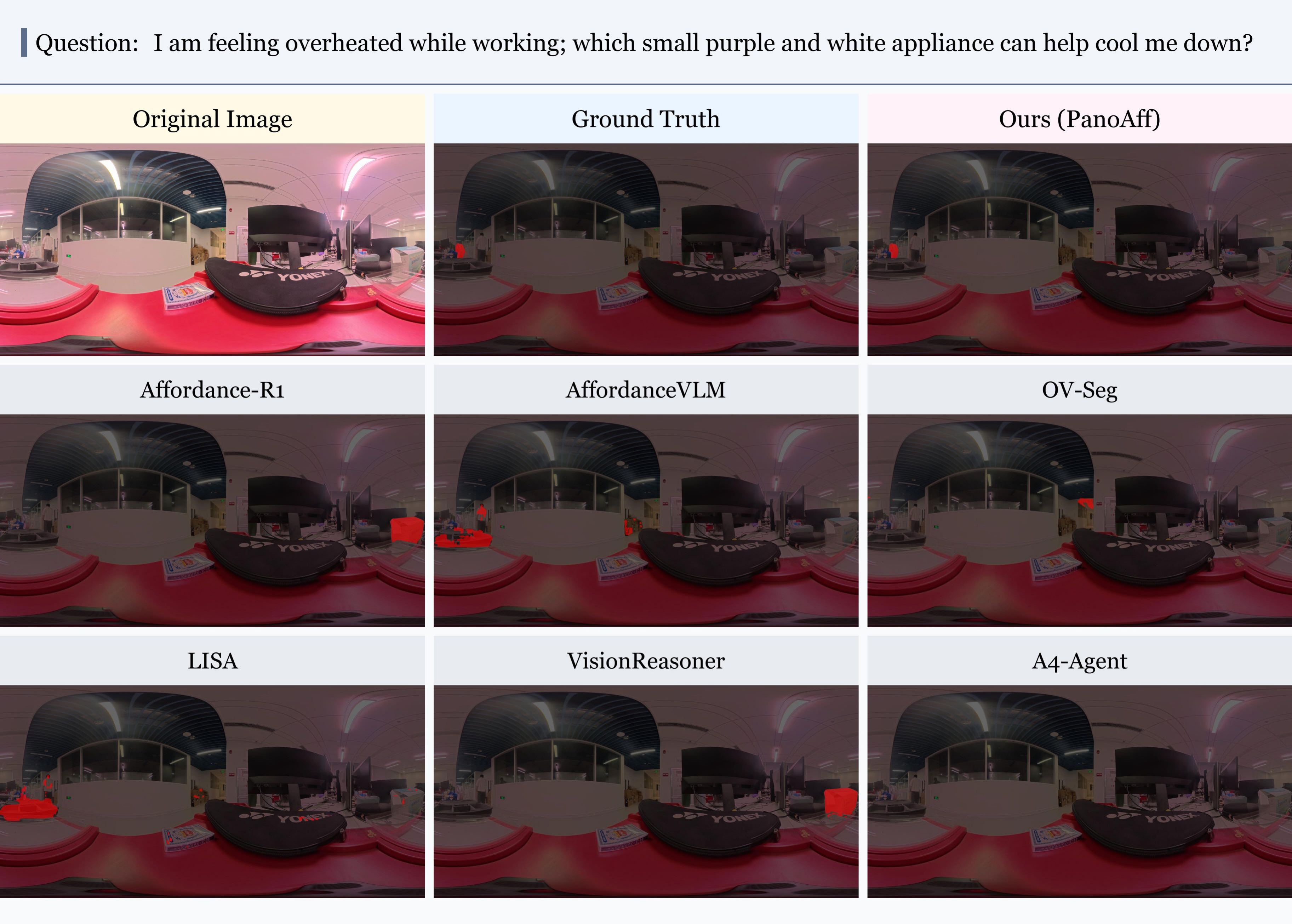}
    \caption{\centering Qualitative comparison on PAP-12K~(Office).}
    \label{fig:compare_office}
\end{figure}

\begin{figure}
    \centering
    \includegraphics[width=0.88\linewidth]{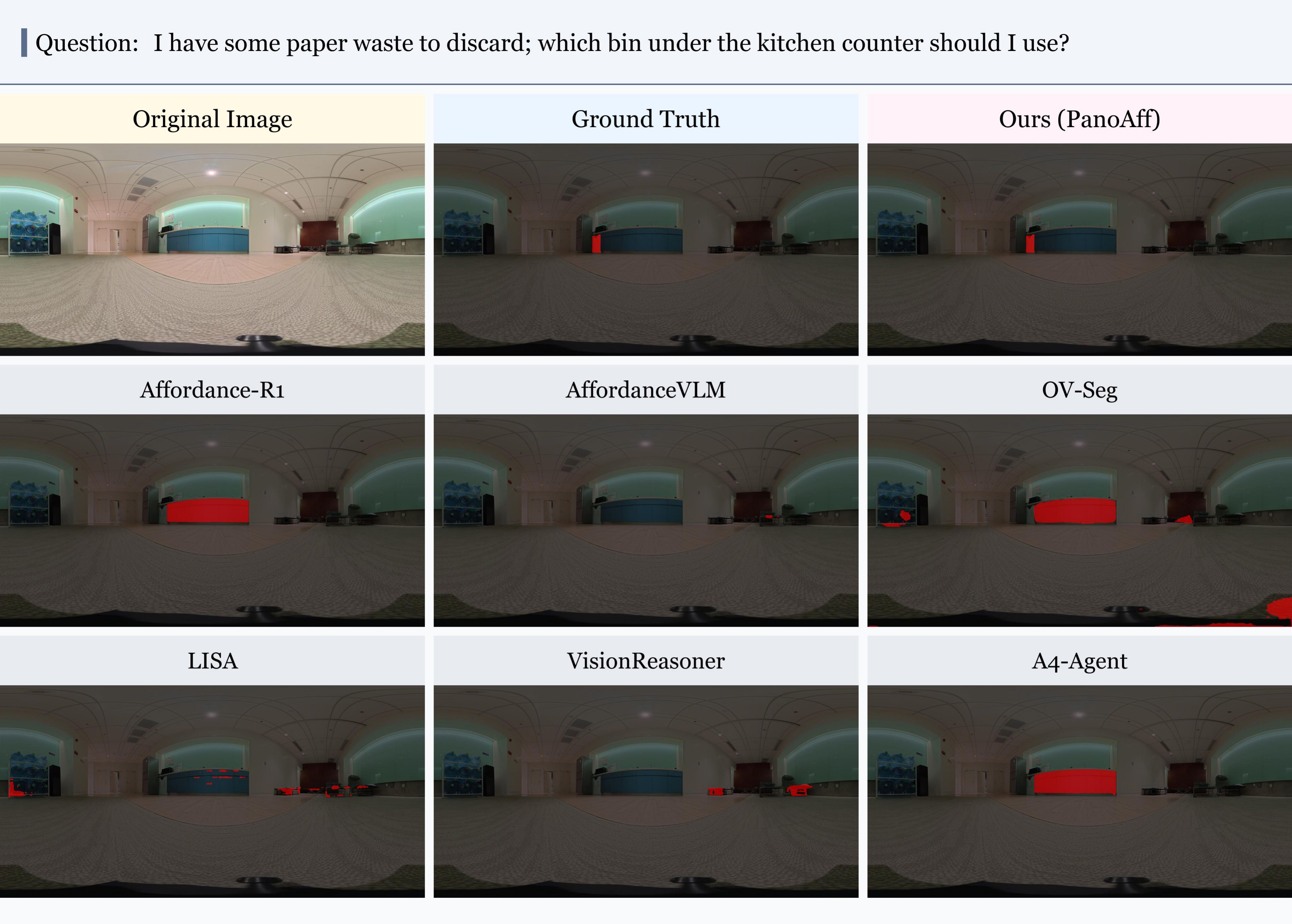}
    \caption{\centering Qualitative comparison on PAP-12K~(Pantry).}
    \label{fig:compare_pantry}
\end{figure}

\begin{figure}
    \centering
    \includegraphics[width=0.88\linewidth]{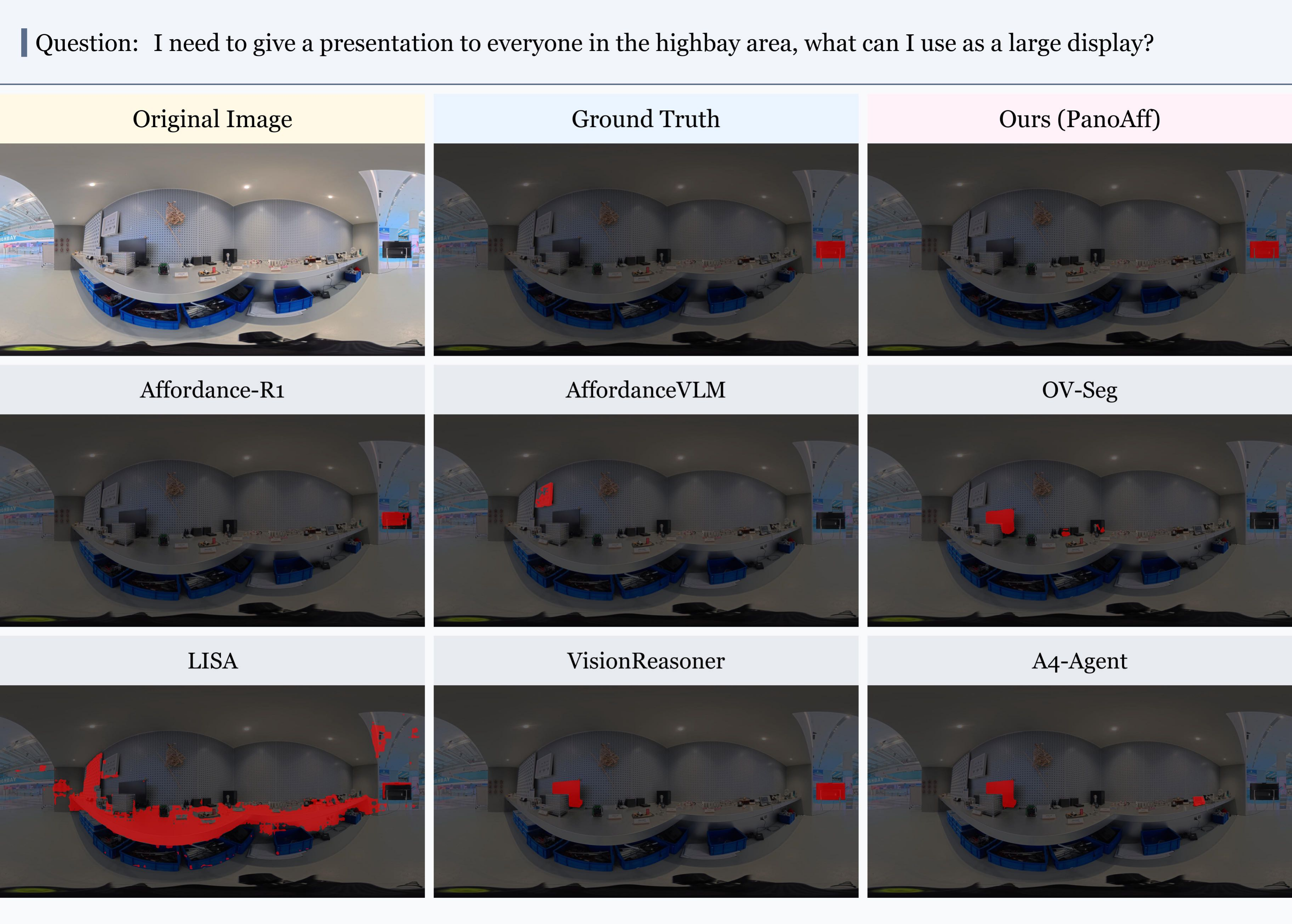}
    \caption{\centering Qualitative comparison on PAP-12K~(Workshop).}
    \label{fig:compare_workshop}
\end{figure}

\begin{figure}
    \centering
    \includegraphics[width=0.88\linewidth]{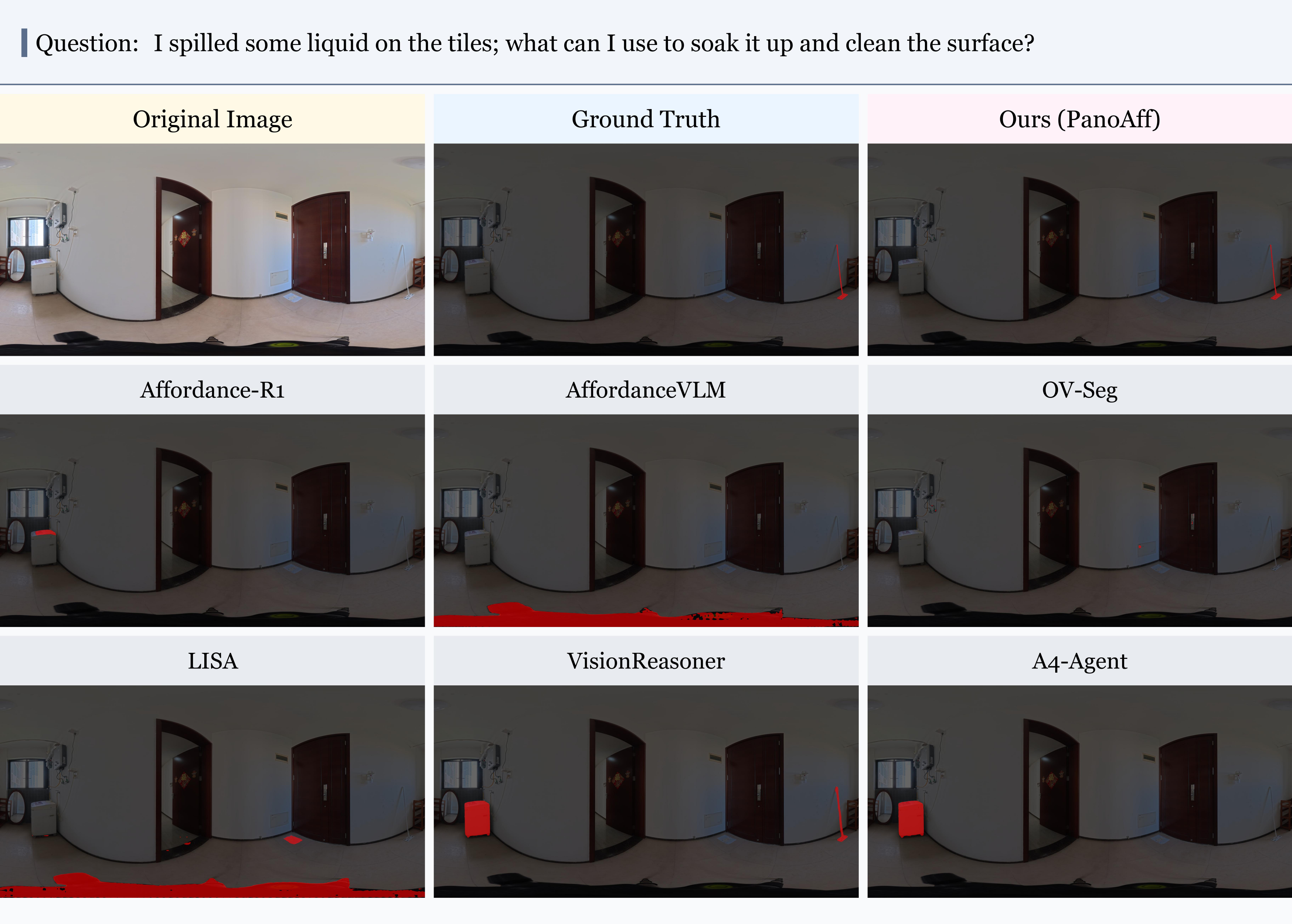}
    \caption{\centering Qualitative comparison on PAP-12K~(Others).}
    \label{fig:compare_others}
\end{figure}

Here, we present further qualitative comparisons between our proposed PAP and the baseline methods evaluated on \dataset. Detailed visualizations can be found in Figs.~\ref{fig:compare_balcony}--\ref{fig:compare_others}. As illustrated, our approach consistently demonstrates superior performance over the baseline methods across all scene categories.

\section{Ethics Statement}
The data collection process for this research involved capturing panoramic images in both private and public environments. To ensure strict adherence to ethical standards and privacy protection, the following protocols were implemented:
\textbf{1) Data Collection Consent:} For specific private or semi-public locations, explicit permission was obtained from the respective property owners or managers prior to capturing the images. Images in public areas were captured without disrupting normal public activities.
\textbf{2) Privacy Protection:} During the image acquisition phase, we proactively avoided capturing highly private or sensitive areas. Furthermore, we deliberately ensured that no frontal human faces or any identifiable personal features were captured during the process.


\clearpage

\clearpage

\end{document}